\documentclass[twoside,11pt,letter]{article}

\usepackage{jmlr2e}

\usepackage[ansinew]{inputenc}
\usepackage{amsfonts}
\usepackage{amssymb}
\usepackage{amsbsy}
\usepackage[usenames]{color}
\usepackage{amsmath}
\usepackage{multirow}
\usepackage{times}
\usepackage{float}
\usepackage[all]{xy}
\usepackage{lscape}
\usepackage{rotating}
\usepackage{hyperref}
\usepackage{pdfcolmk}
\usepackage{pslatex}

\newcommand{\green}[1]{\textcolor[rgb]{0,0,0}{#1}}
\newcommand{\blue}[1]{\textcolor[rgb]{0,0,0}{#1}}

\jmlrheading{Journal of Machine Learning Research 11}{2010}{873-903}{}{2/10}{Valero Laparra, Juan Guti\'errez, Gustavo Camps-Valls, and Jes\'us Malo}

\ShortHeadings{Image Denoising with Kernels based on Natural Image Relations}{Laparra et al.} \firstpageno{1}

\begin{document}

\title{Image Denoising with Kernels Based on Natural Image Relations}

\author{\name Valero Laparra \email valero.laparra@uv.es \\
       \addr Image Processing Lab, Universitat de Val\`encia \\
       46100 Burjassot, Val\`encia, Spain.
       \AND
       \name Juan Guti\'errez \email juan.gutierrez@uv.es \\
       \addr Dept. Inform\`atica, Universitat de Val\`encia \\
       46100 Burjassot, Val\`encia, Spain.
       \AND
       \name Gustavo Camps-Valls \email gustavo.camps@uv.es \\
       \addr Image Processing Lab, Universitat de Val\`encia \\
       46100 Burjassot, Val\`encia, Spain.
       \AND
       \name Jes\'us Malo \email jesus.malo@uv.es \\
       \addr Image Processing Lab, Universitat de Val\`encia \\
       46100 Burjassot, Val\`encia, Spain.}

\editor{Donald Geman}

\maketitle

\begin{abstract}
\green{A successful class} of image denoising methods is based on Bayesian approaches working in
wavelet representations. The performance of these \green{methods} improves \green{when relations among the local frequency coefficients are explicitly included}. However, in these techniques, analytical
estimates can be obtained {\em only} for particular combinations of \green{analytical models}
of signal and noise, thus precluding its straightforward extension to deal with other arbitrary noise sources.

In this paper,
we propose an alternative non-explicit way to take into account the relations
among natural image wavelet coefficients for denoising: we use support vector regression
(SVR) in the wavelet domain to enforce these relations in the estimated signal.
Since relations
among the coefficients are specific to the signal, the
regularization property of SVR is exploited to remove the noise, which does not
share this feature. The specific signal relations
are encoded in an anisotropic kernel obtained from mutual information measures computed on a
representative image database. In the proposed scheme,
training considers minimizing the Kullback-Leibler divergence (KLD) between the estimated
and actual probability functions (or histograms) of signal and noise in order
to enforce similarity up to the higher (computationally estimable) order.
Due to its non-parametric nature, the method can eventually cope
with different noise sources without the need of an explicit re-formulation, as it is
strictly necessary under parametric Bayesian formalisms.

Results under several noise levels and noise sources show that:
(1) the proposed method outperforms conventional wavelet
methods that assume coefficient independence, (2) it is similar to state-of-the-art methods
that do explicitly include these relations when the noise source is Gaussian, \blue{and (3)
it gives better numerical and visual performance when more complex, realistic
noise sources are considered.} Therefore, the
proposed machine learning approach can be seen as a more
flexible (model-free) alternative to the explicit description
of wavelet coefficient relations \green{for image denoising}.
\end{abstract}


\section{Introduction}\label{statement}
Denoising requires representing the distorted signal in a domain
where signal and noise display different enough behavior. In such a
representation, noise is removed by imposing the known properties of the
signal to the distorted samples. In image denoising, classical
regularization techniques are used to impose smoothness in the spatial domain since noise is typically
white~\citep{Banham97}. Smoothness in the spatial domain
means predictability of the signal from the neighborhood, and thus
\blue{classical} approaches exploit the low-pass behavior of
the power spectrum to rely on band-limitation or autoregressive
models of the signal~\citep{Andrews77,Banham97,Bertero88}. Several image denoising methods working in the spatial domain have been presented
in the literature, either based on splines~\citep{Takeda07}, patch-based
approximations~\citep{Kervrann07}, local auto-regressive models~\citep{Gutierrez06},
or support vector regression~\citep{Chow01,Ginneken06} to perform
smooth (regularized) approximations of the noisy signal.
\green{Recently, successful methods use adaptive
local basis representations \citep{Dabov07}. Approaches to the problem
using local basis is qualitatively related to wavelet descriptions.
In fact, wavelet representations have been recognized as quite appropriate
domains for image denoising\footnote{In the 2007 IEEE International Symposium on
Information Theory (ISIT2007), the tutorial ``Recent Trends
in Denoising'' ({\tt http://www.stanford.edu/$\sim$slansel/tutorial/summary.htm})
pointed out that state-of-the-art methodologies are usually defined in the wavelet domain.}.}

Wavelet representations are convenient in image denoising because
natural image samples have a very specific statistical behavior in this domain.
On the one hand, smoothness is represented by a strong energy compaction
in coarse scales. On the other hand, the combination of smooth regions with
local, high contrast features, such as edges, gives rise to sparse activation
of wavelet sensors. \blue{This leads to very particular, heavy-tailed, marginal
probability density functions (PDFs) of the wavelet coefficients \citep{Burt83,Field87,Simoncelli97,Hyvarinen99}.}
These basic features were incorporated in the classical wavelet-based image denoising
techniques~\citep{Donoho95,simoncelli99b,Figueiredo01}.
Classical techniques such as hard and soft thresholding~\citep{Donoho95}
have been derived by using Bayesian approaches in \green{non-redundant wavelets}, looking for
either \emph{Maximum a Posteriori} (MAP) or \emph{Bayesian Least Squares} (BLS) estimators,
in combination with simple marginal models and assuming
statistical independence among coefficients~\citep{simoncelli99b,Figueiredo01}.

It is well-known, however, that marginal models in the wavelet
domain are not enough for a proper signal characterization:
\blue{relevant relations among coefficients still remain after wavelet transforms~\citep{simoncelli99b}.
For instance,
edges lead to strong coupling between the energy of neighboring wavelet coefficients of natural images}.
These relations among wavelet coefficients have proven to be a key issue in
applications such as image coding~\citep{Malo06a,Camps07}, texture analysis and synthesis~\citep{Portilla00} or image quality metrics~\citep{Laparra10}. The use of these relations is in the roots of the most recent and successful image
denoising approaches as well~\citep{Portilla03,simoncelli06,Goossens09}. In this case, more complex image models explicitly including the relations among coefficients
have to be plugged and fitted into the Bayesian framework to obtain the image estimates.

\newpage

Unfortunately, all these model-based Bayesian techniques have three common problems:
\begin{enumerate}
\item They critically depend on the accuracy of the image model, whose definition is not trivial;
\item MAP or LS estimations can only be derived analytically for particular, typically Gaussian,
noise sources. For different noise sources, the whole technique has to be
reformulated which may not be analytically tractable;
\item The estimation of the parameters of the image model from the noisy observation is difficult in general.
\end{enumerate}
Conversely, non-parametric approaches can include the above qualitative properties
in an indirect way without the restriction of being analytically attached
to particular image or noise models. These approaches are based on {\em learning} the underlying model directly from
the image samples.

In this work we apply support vector regression (SVR) in a redundant (overcomplete) wavelet domain to the noisy image. The proposed method has the following advantages in front of the Bayesian framework:
\begin{enumerate}
\item It does not use a particular parametric image model to be fitted, \blue{e.g., no analytical PDF is required}.
\item Its solution may be found for arbitrary noise sources even without knowing the
functional form of the noise PDF since it can work with just noise histograms. Therefore,
the procedure does not need to be reformulated for different noise sources.
\item It is capable to take into account the relations among wavelet coefficients of
natural images through the use of a suitable kernel. In this way,
the method preserves the relevant relations among the coefficients of the true signal
and better removes the degradation.
\end{enumerate}
The proposed method does not assume
independence among the signal coefficients in the wavelet domain, as opposed to \citep{simoncelli99b,Figueiredo01}, nor an explicit model of signal relations, as done in \citep{Portilla03}.
Therefore, the proposed machine learning approach can be seen as a more flexible (model-free)
alternative to the explicit description of wavelet coefficient relations for image denoising.
\blue{ Even though the selection of a particular SVR may be seen as a signal parametrization,
the model is still non-parametric in the sense that no functional form of the signal (or noise)
characteristics (e.g. the PDF) is assumed.}

Non-explicit use of dependencies in local frequency domains for denoising was
also introduced in~\citep{Gutierrez06}. In that case, relations were embedded into a perceptual
model used for non-parametric spectrum estimation, and offered better results
than local parametric autoregressive models not including these relations.
Here we pursue the same goal (a model-free technique including local frequency relations), but with a completely
different fra\-me\-work (SVR instead of perceptual information).
The idea of using SVR regularization in the wavelet domain for image denoising has been recently
introduced in~\citep{Chow01,Cheng04,Ginneken06}. \blue{However, in these works,
(1) the qualitative effect of the different parameters of the SVR was not analyzed,
(2) these parameters were set without plausible justification of their values, and more importantly,
(3) the relevance of the relations among the wavelet coefficients of the signal
was not an issue, so the ability of SVR to take these relations into account in the kernel
was neither assessed nor compared to other methods that do consider them.
In fact, a trivial isotropic Gaussian kernel was used in all cases.}
\green{ On the contrary, in this paper we address the key following issues:}
\begin{itemize}
 \item \green{\textbf{Natural images features in redundant wavelet domains.}
        Interesting insight about the problem can be obtained by analyzing the mutual information between the coefficients of wavelet representations \citep{Buccigrossi99,Liu01}. However, in redundant domains, it is strictly necessary to discern what are the relations specific to
        the signal and those due to the transform.}

\item \green{\textbf{General constraints of the SVR parameters in image denoising.}
        Generic recommendations about the SVR parameters have been adapted to propose specific subband-dependent profiles for the insensitivity and the penalization parameters, and to propose a mutual information based kernel.}

\item \green{\textbf{Effect of the SVR parameters.} We show the qualitative effect of varying the values of the parameters under the constrained parameter space.}

\item \green{\textbf{Procedure to optimize the SVR parameters.} We propose an automatic procedure to select the SVR parameters based on the Kullback-Leibler divergence, under certain assumptions on signal and noise.}
\end{itemize}
\green{Even though this methodological framework is proposed in the context of achromatic image denoising, it can be readily extended to other denoising problems in which wavelet coefficients exhibit particular relations, such as in color or multispectral images, speech signals, etc.}

The remainder of the paper is outlined as follows. In Section 2, we point out relevant signal features in redundant
wavelet domains through mutual information measurements. These key properties will be used by the proposed algorithm presented in
Section 3. In Section 4, the effect of SVR parameters and the validity of the proposed criterion for its selection is addressed experimentally.
Section 5 shows the performance of the proposed method compared to standard denoising methods in the wavelet domain.
Several experiments dealing with different amount and nature of noise illustrate the capabilities of our proposal.
Finally, Section 6 draws some conclusions and outlines the further work.

\section{Features of natural images in the steerable wavelet domain}
\label{features}

The starting hypothesis for image denoising is that signal and noise display different
characteristics and thus it is possible to separate them in a certain domain. Natural images
show non-trivial relationships among wavelet transform coefficients.
In the following, we review the reported statistical properties of natural images in orthogonal wavelet domains,
and then \blue{analyze them in the redundant steerable
wavelet domain selected in our implementation.} \green{Specifically we will use mutual information (MI) to assess the statistical relations among wavelet coefficients of natural images as in~\citep{Buccigrossi99,Liu01}.}

\subsection{Intraband versus interband signal relations in orthogonal wavelets}

 Dependencies among {\em orthogonal wavelet} coefficients were measured using mutual information in~\citep{Liu01}. The
dependencies were studied at interband and intraband levels, and the results suggested
that the mutual information between intraband neighbors is typically {\em larger} than the
interband relations for several models and types of interaction. In \citep{Buccigrossi99},
the authors analyzed the linear predictability of a coefficient's magnitude from a
conditioning coefficient set, either its parent, neighbors (left and upper),
cousins (coefficients at the same location but in different orientation subbands),
or aunts (cousins of the parent). After an exhaustive mutual information analysis,
the parent provided less information content than the neighbors. These evidences suggest that the dependencies among
spatial neighboring coefficients (intraband) in orthogonal wavelet descriptions are stronger
than the interband dependencies.

\subsection{Natural images relations in steerable wavelets}\label{steerablerelations}

Redundant (non-orthogonal) wavelet representations may be more suited to
image denoising applications since redundant representation of the image features \green{may make the signal inherent relations clearer}.
\blue{
Specifically, some redundant representations are designed to be translation or rotation invariant \citep{Freeman91,coifman95,Kingsbury06}. This behavior is convenient to ensure that a particular feature in different spatial regions (or with different orientations)
gives rise to the same neighboring relations.
Some translation invariant wavelets \citep{simoncelli95} have also a smoother rotation behavior than non-redundant transforms.
This justifies applying the same processing all over a particular subband and dealing with the different
orientations in similar ways. Besides, this prevents aliasing artifacts appearing in critically-sampled wavelets.
In this work we choose a redundant steerable pyramid representation \citep{simoncelli95} to take
advantage of these properties.
}

Despite the reported results on the relations of signal coefficients in orthogonal transforms,
a number of questions have to be answered in the \green{case of redundant representations}, and in particular, in the
steerable wavelet domain:
\begin{enumerate}
\item \blue{How relevant are the relations among coefficients of natural images in this domain?}
\item How relatively important are interorientation, interscale and \blue{intraband} signal relations?
\item How is the spatial arrangement of these signal relations?
\end{enumerate}
The first question is particularly important since, even though the steerable transform
may intensify the relations among signal coefficients, its redundant
nature may also introduce relations
\blue{which could be due to the transform but not to the signal}.
The second question allows us to focus on the most significant relations. Answering the third question is crucial to design suitable kernels for image denoising.

In the following, we get some insight on these concerns by performing two experiments
on a representative database of $920$ achromatic images of size
$256\times 256$ extracted from the McGill Calibrated Colour Image Database\footnote{See {\tt http://tabby.vision.mcgill.ca/}.}.

\subsubsection{\green{Signal relations are specific to the signal}}

\begin{figure}[t!]
   \begin{center}
      \begin{tabular}{ccc}
        \hspace{-0.6cm}(a) & \hspace{-0.6cm}(b) & \hspace{-0.6cm}(c) \\
         \hspace{-0.5cm}\includegraphics[width=0.33\textwidth]{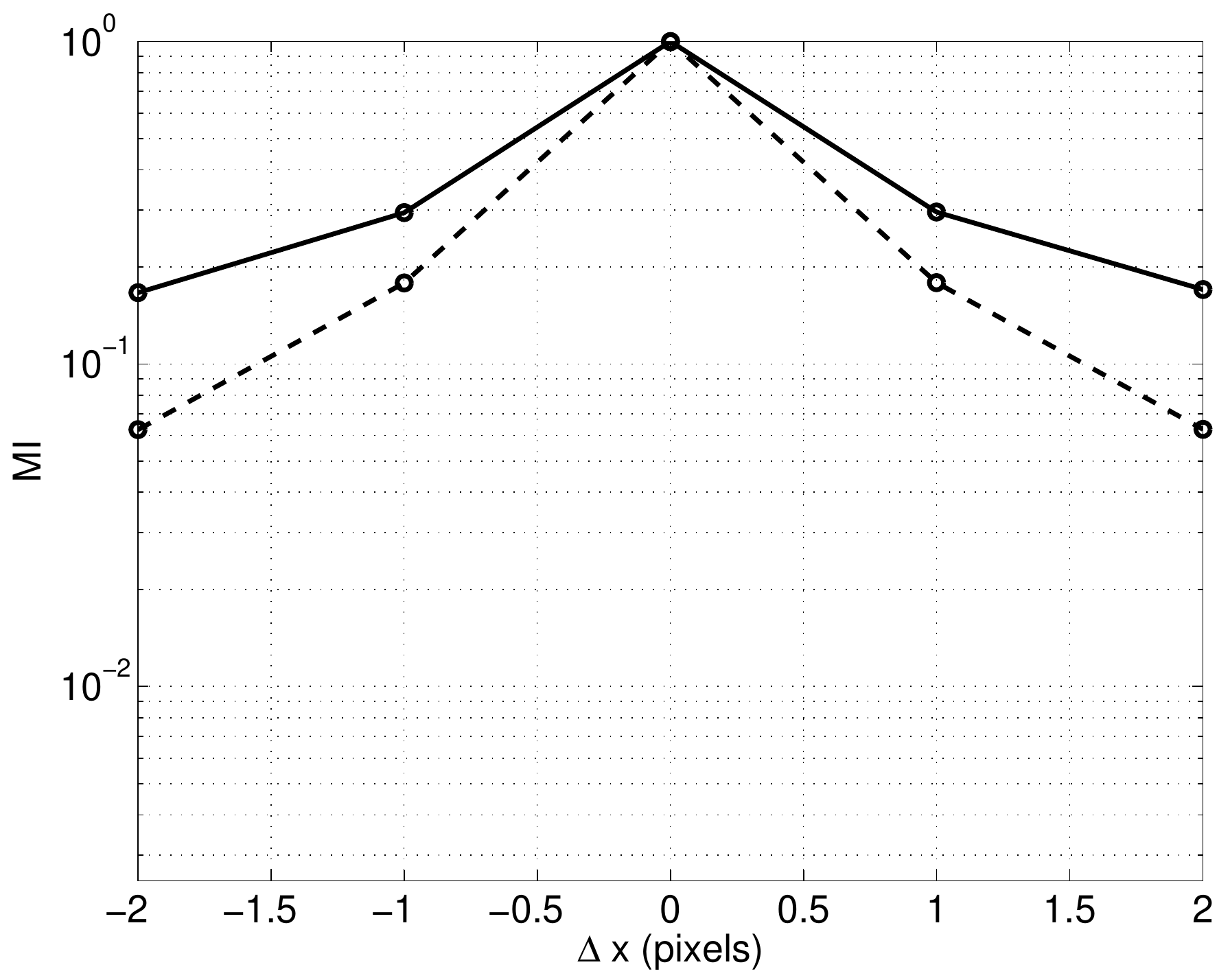} &
         \hspace{-0.2cm}\includegraphics[width=0.33\textwidth]{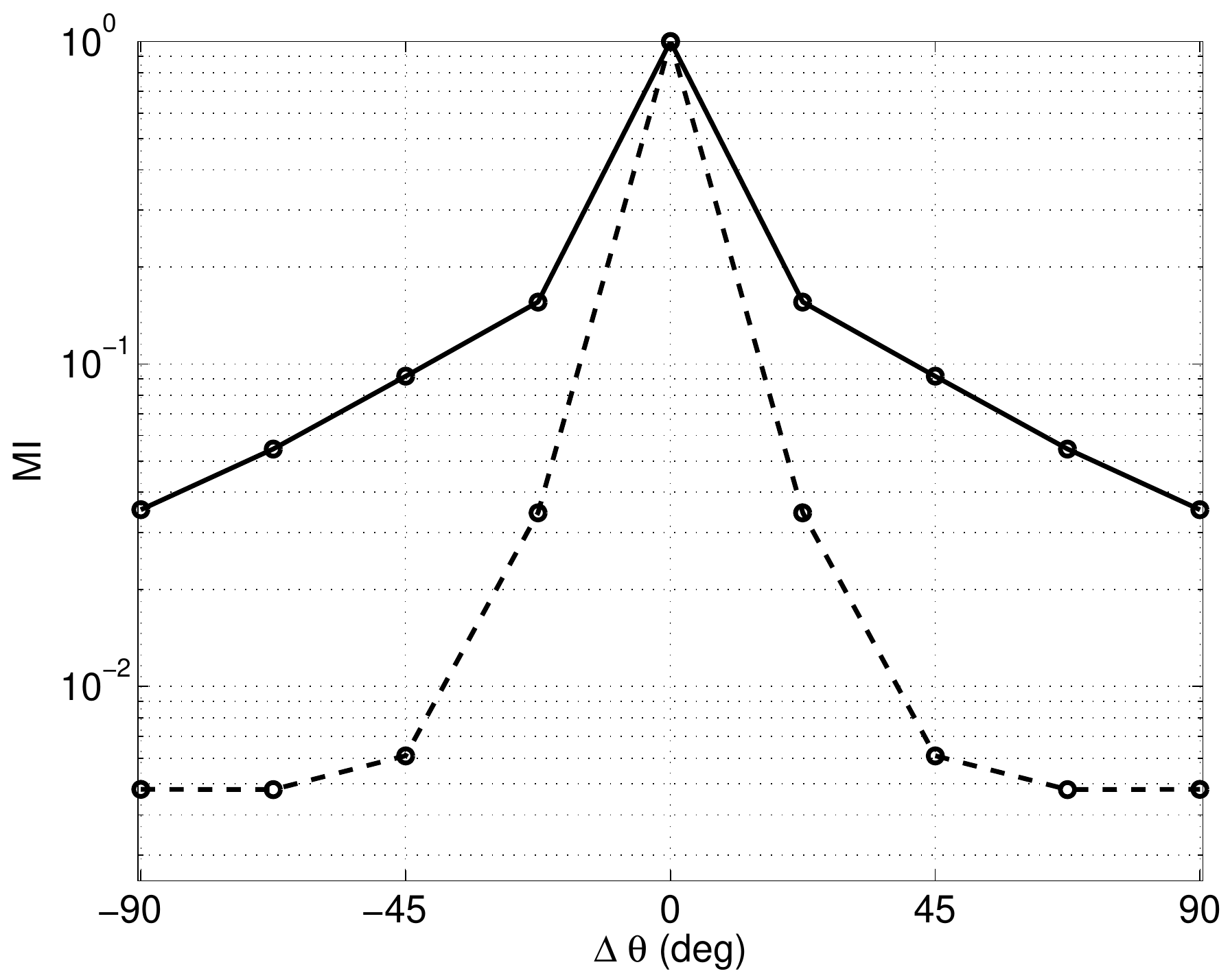} &
         \hspace{-0.2cm}\includegraphics[width=0.33\textwidth]{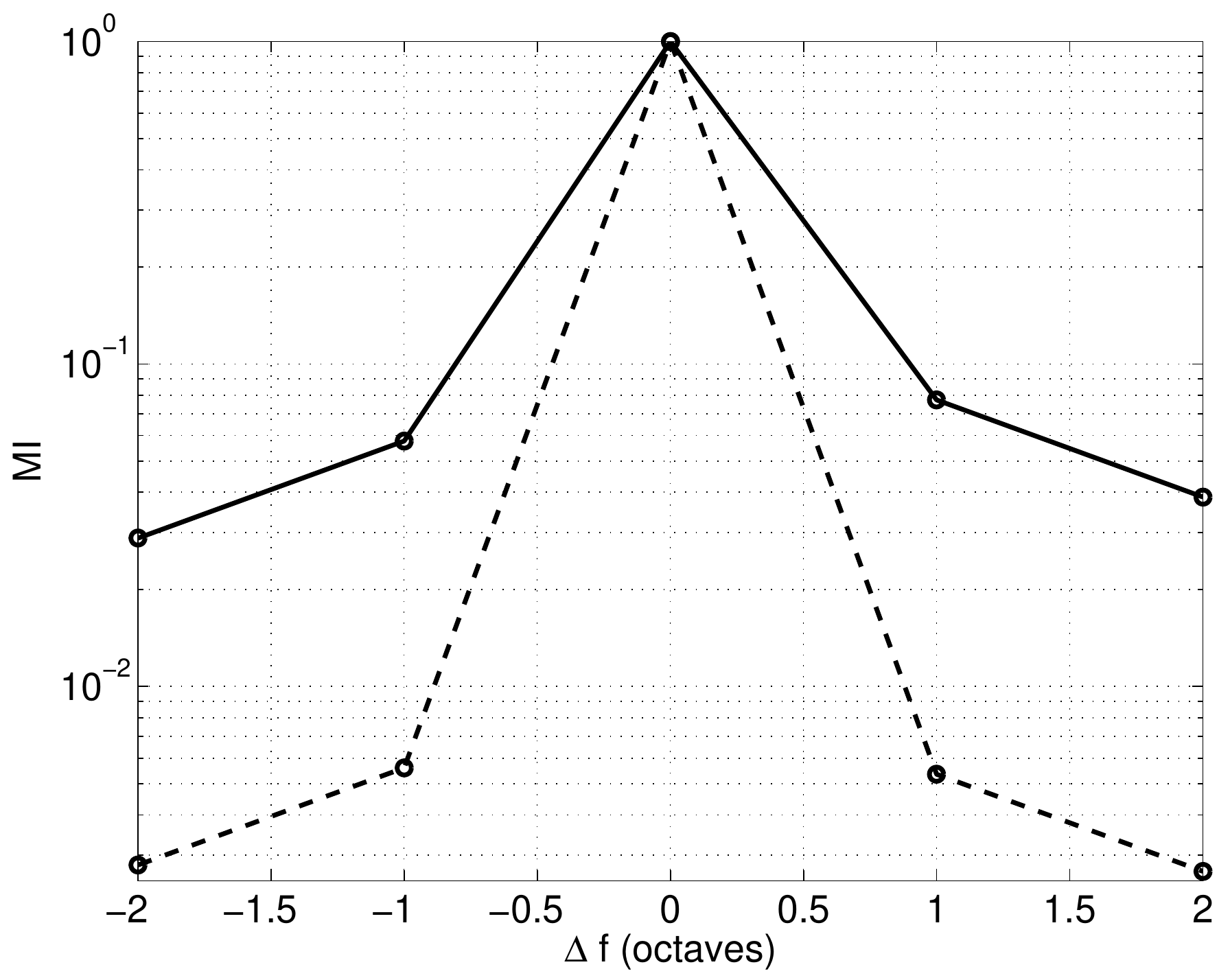}
      \end{tabular}
         \caption{\green{Comparison between redundancy of natural image coefficients
         in the steerable wavelet representation (solid), and the redundancy due to
         this representation (dashed). Redundancy is measured
         in terms of relative mutual information in logarithmic scale
         among (a) spatial (b) orientation  and (c) scale neighbors. }}
         \label{inter_intra_ori}
   \end{center}
\end{figure}

In our first test, \green{following \citep{Liu01},} we computed the mutual information among steerable wavelet coefficients of the dataset for different spatial, orientation, and scale distances. \blue{We used a steerable pyramid with $8$ orientations and $4$ scales.}
The mutual information was estimated from the uniformly binned empirical data ($256$ bins) by computing the histogram of all available sample pairs ($721280$ samples) for the three considered neighborhoods.
\green{In addition, as stated above, in redundant domains it is necessary to know whether these relations come from the images or they are due to the transform.}
\green{Note that, considering i.i.d. signals, any relation among the coefficients after a linear transform will be due to the transform no matter their PDF in the original domain. Therefore, in order to assess the amount of relations due to the transform, we compared the MI among natural images coefficients, and the MI among the coefficients of an i.i.d. signal (Fig. \ref{inter_intra_ori}).
The relations displayed by i.i.d. signals in the transformed domain may be seen as a lower bound for the mutual information of signal coefficients.}
\green{From Fig. \ref{inter_intra_ori}}, it can be noticed that, in every case, \blue{relations found in natural images are bigger than those introduced by the transform}.

\subsubsection{\green{Intraband signal relations dominate over interscale or orientation}}

Besides, the results show that intraband relations in the signal are also
more important than interorientation or interscale relations.
\green{Note that mutual information measures are defined to depend on logarithms of probability so
that comparisons have to be done by subtraction, not by division.}
Beyond consistency with previously reported results for orthogonal wavelet transforms \citep{Buccigrossi99,Liu01}, \blue{it has been observed that the relations are specific to the signal and not} \green{just} \blue{induced by the transform}.

\subsubsection{\green{Intraband relations are strongly oriented}}

\begin{figure}[t!]
   \begin{center}
      \begin{tabular}{cccccccc}
         \includegraphics[width=1.5cm]{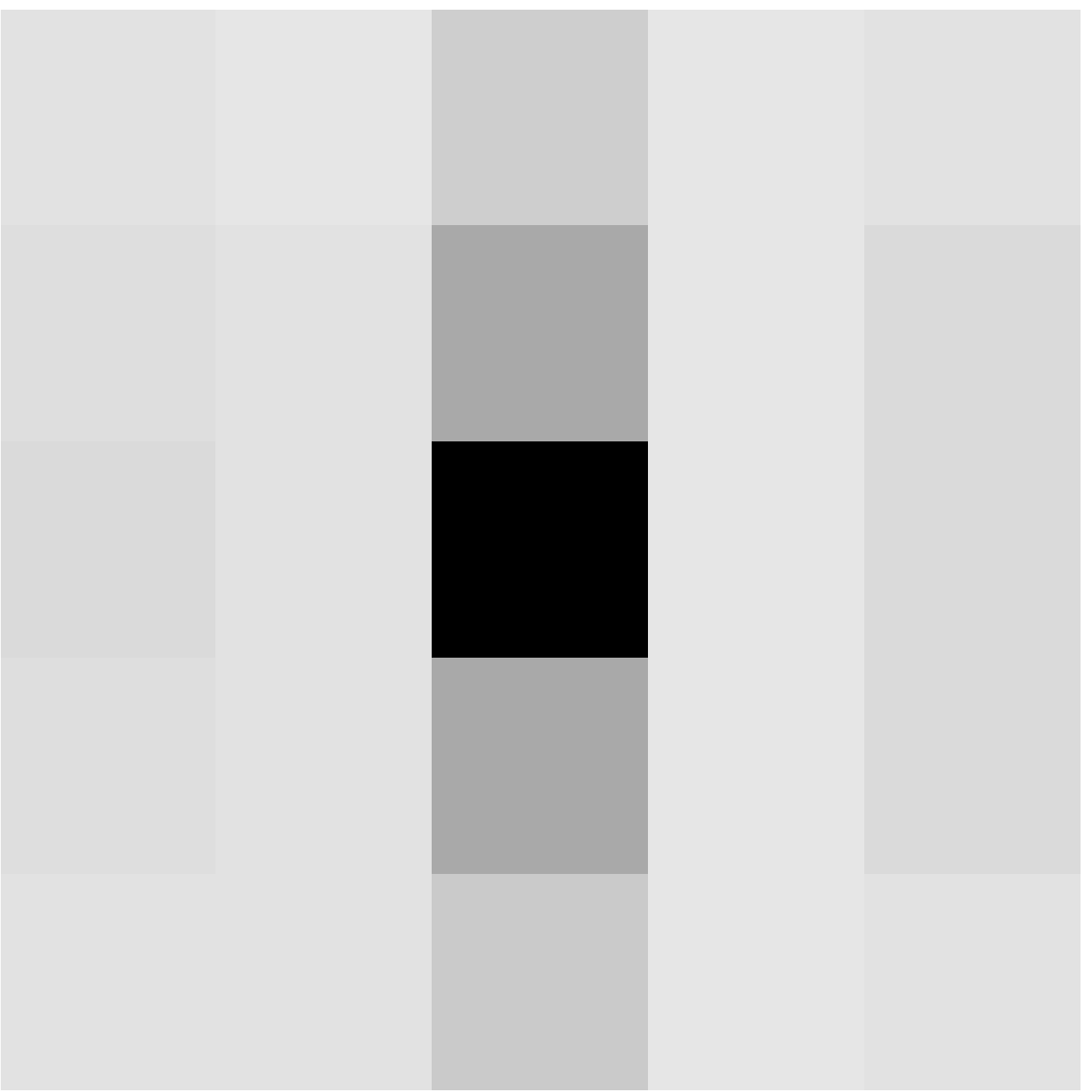} &
         \includegraphics[width=1.5cm]{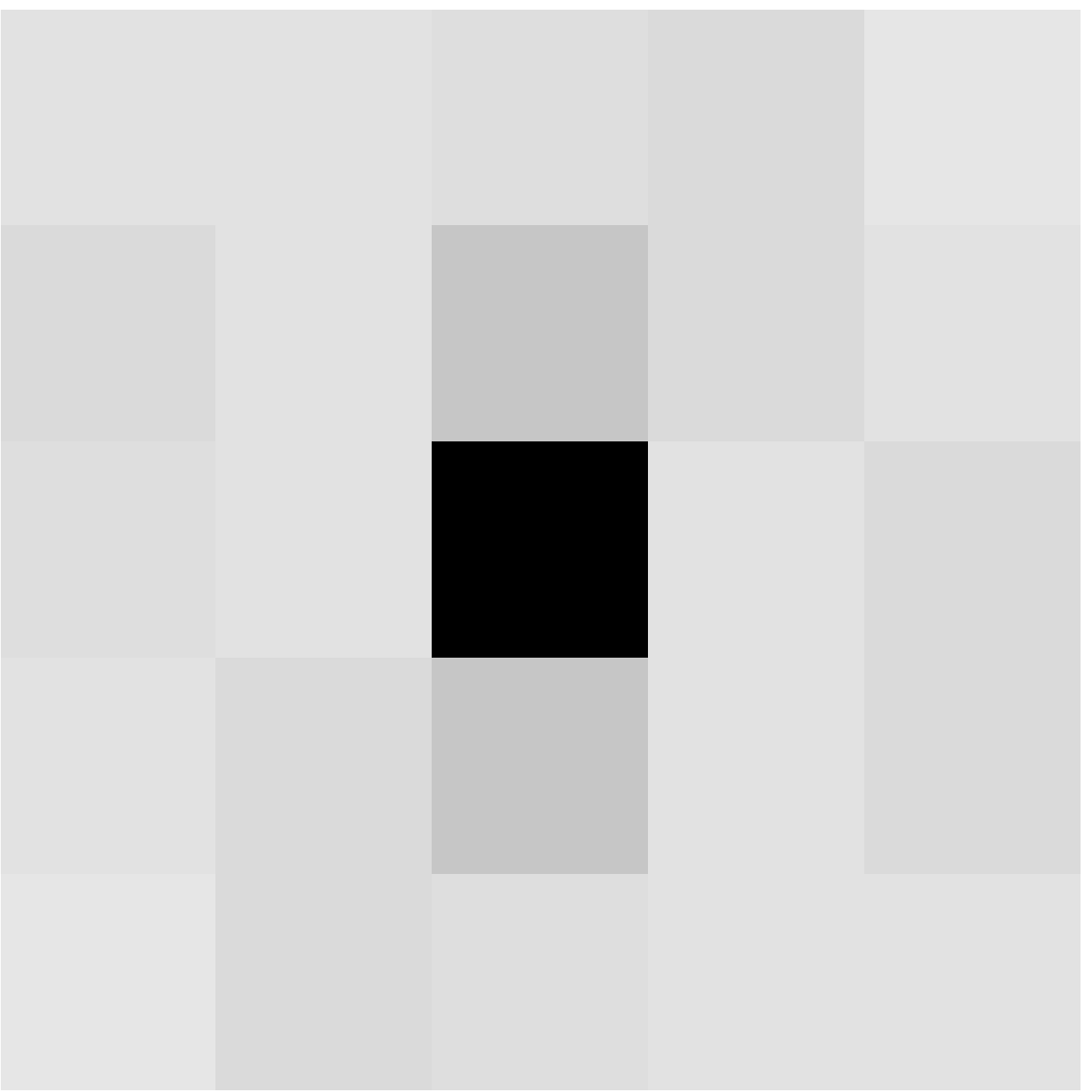} &
         \includegraphics[width=1.5cm]{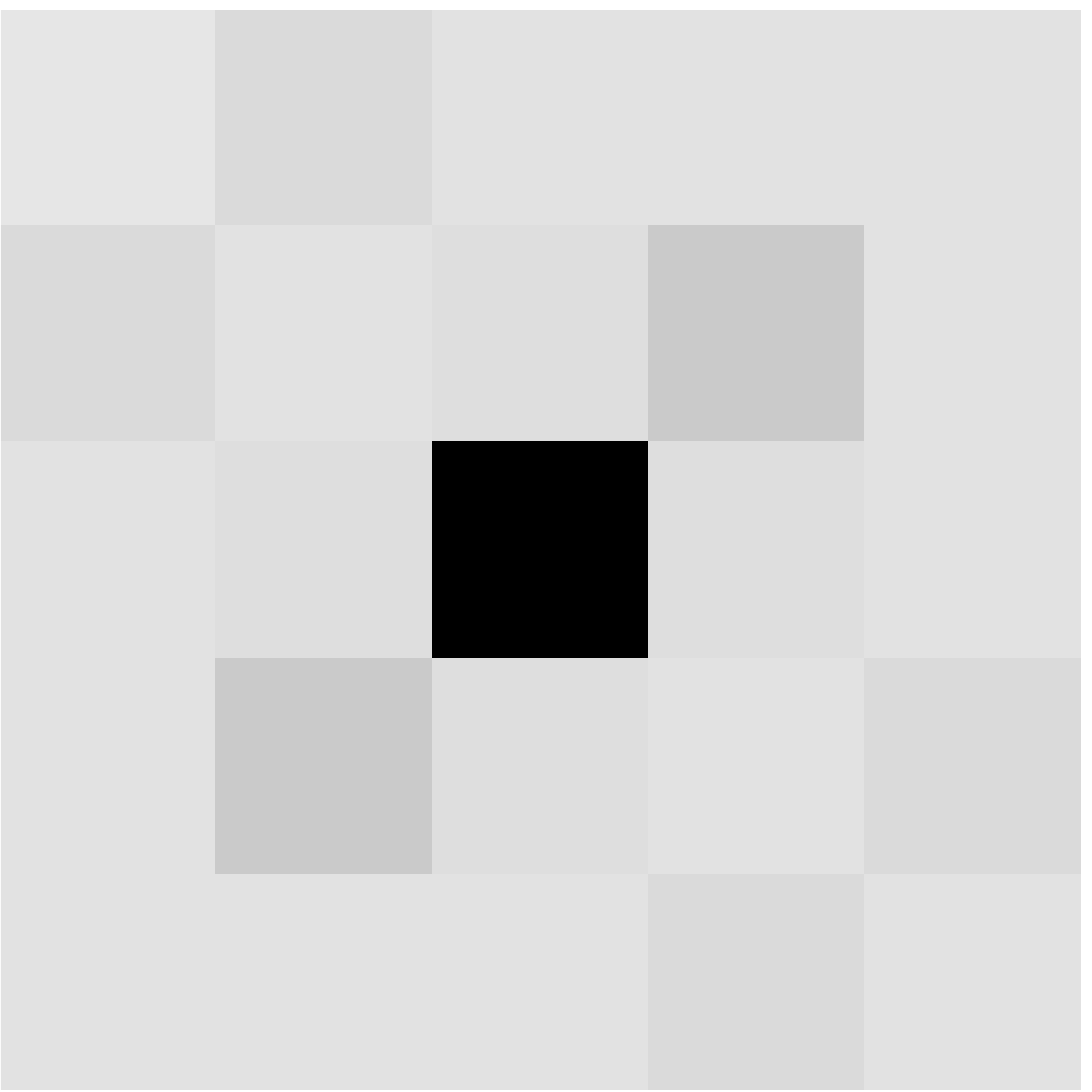} &
         \includegraphics[width=1.5cm]{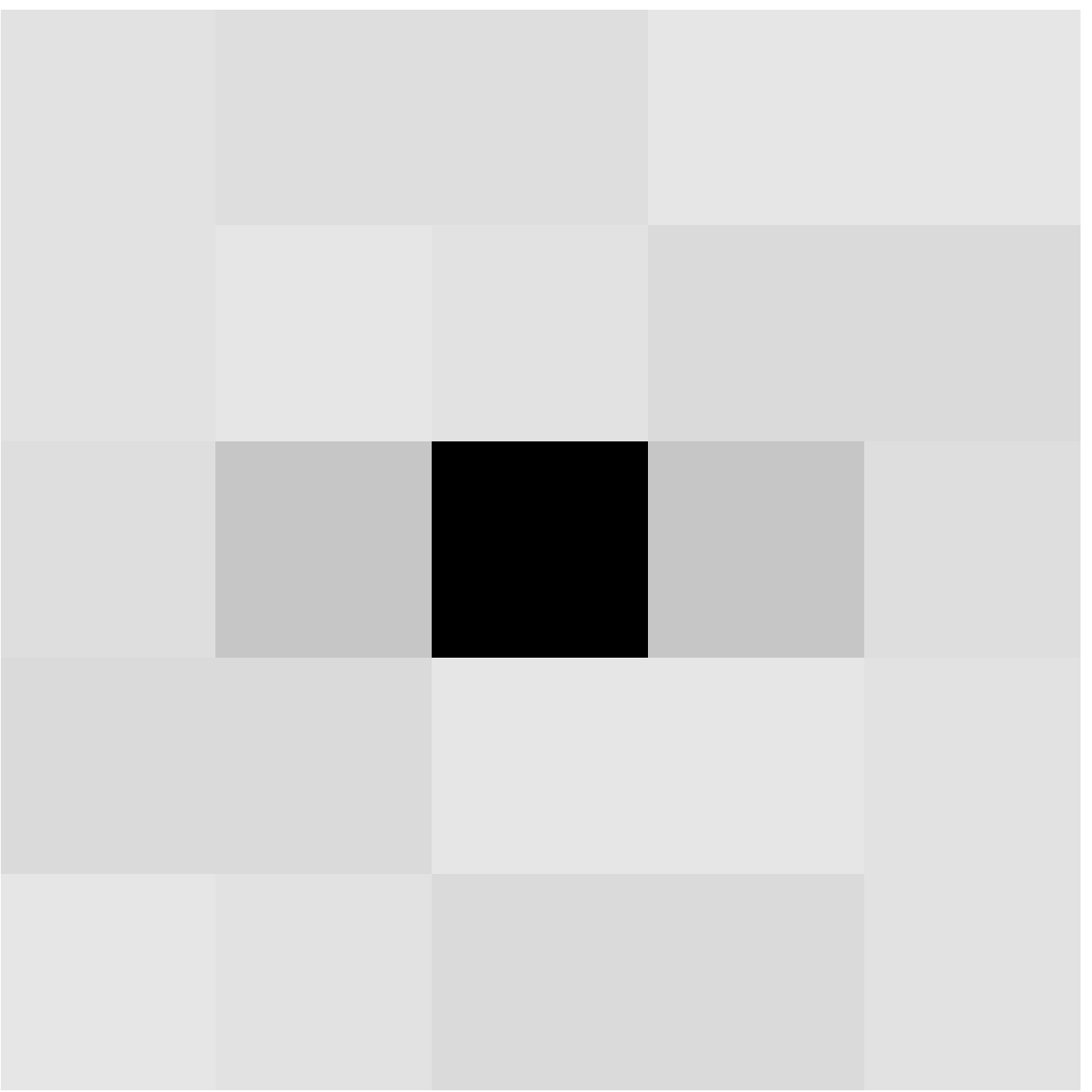} &
         \includegraphics[width=1.5cm]{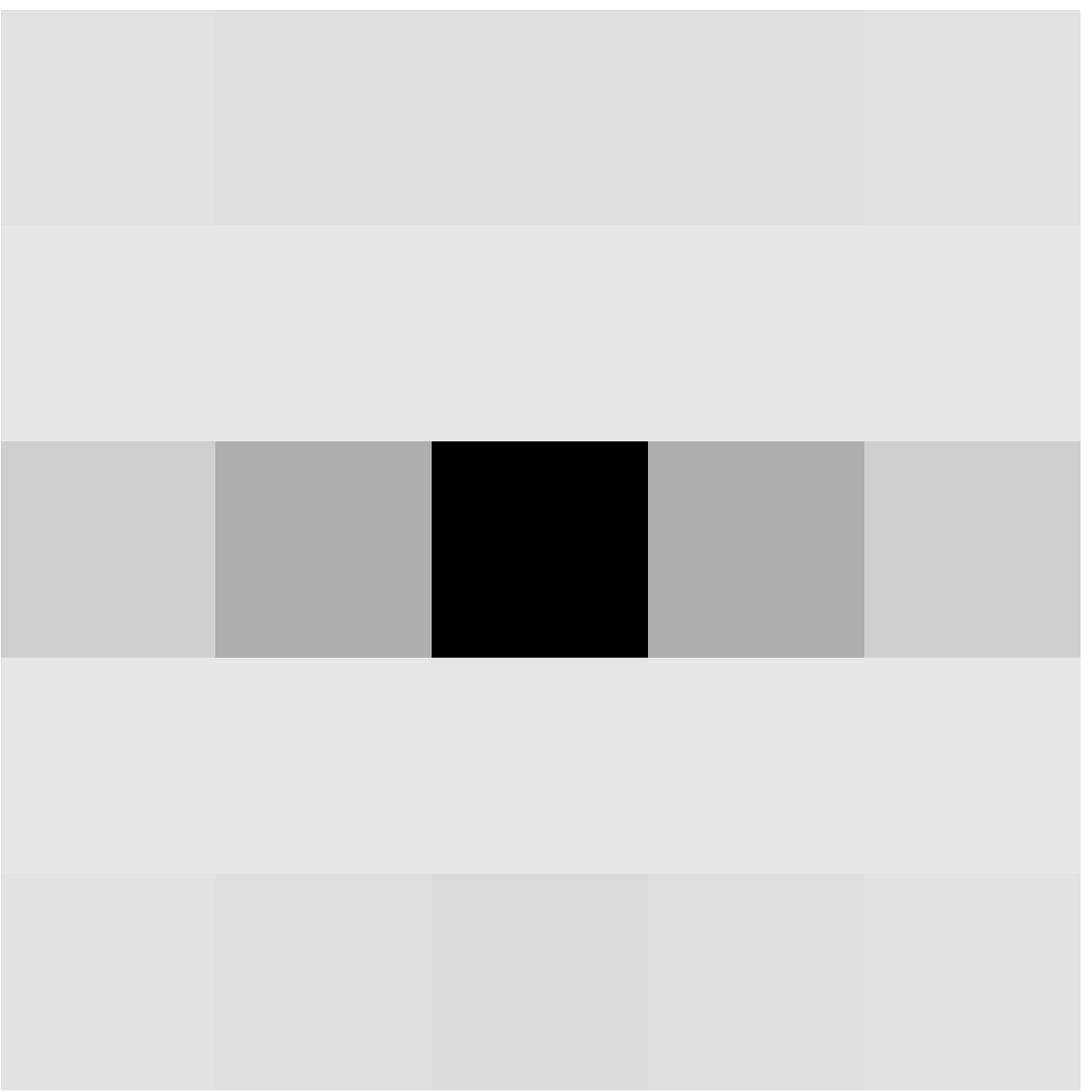} &
         \includegraphics[width=1.5cm]{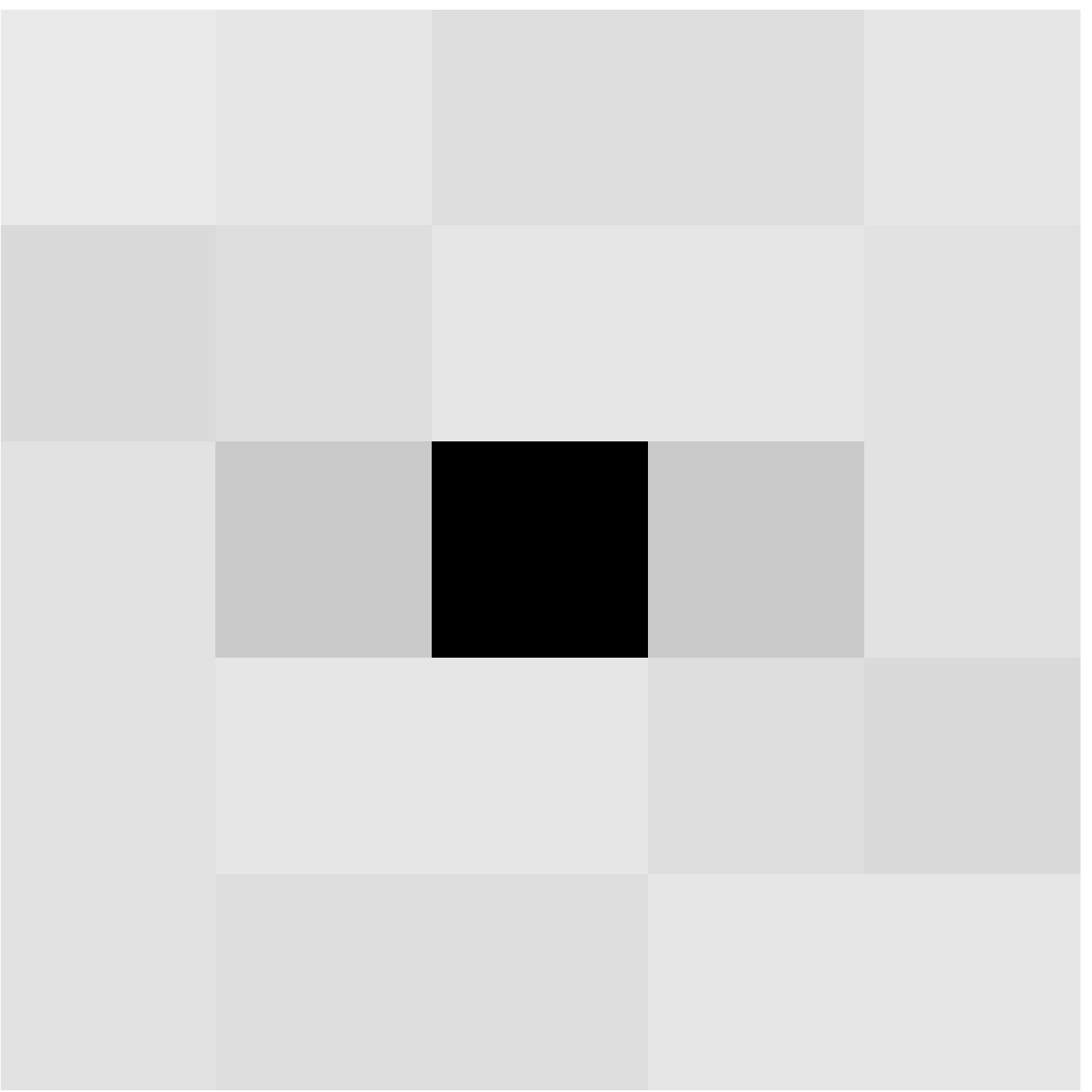} &
         \includegraphics[width=1.5cm]{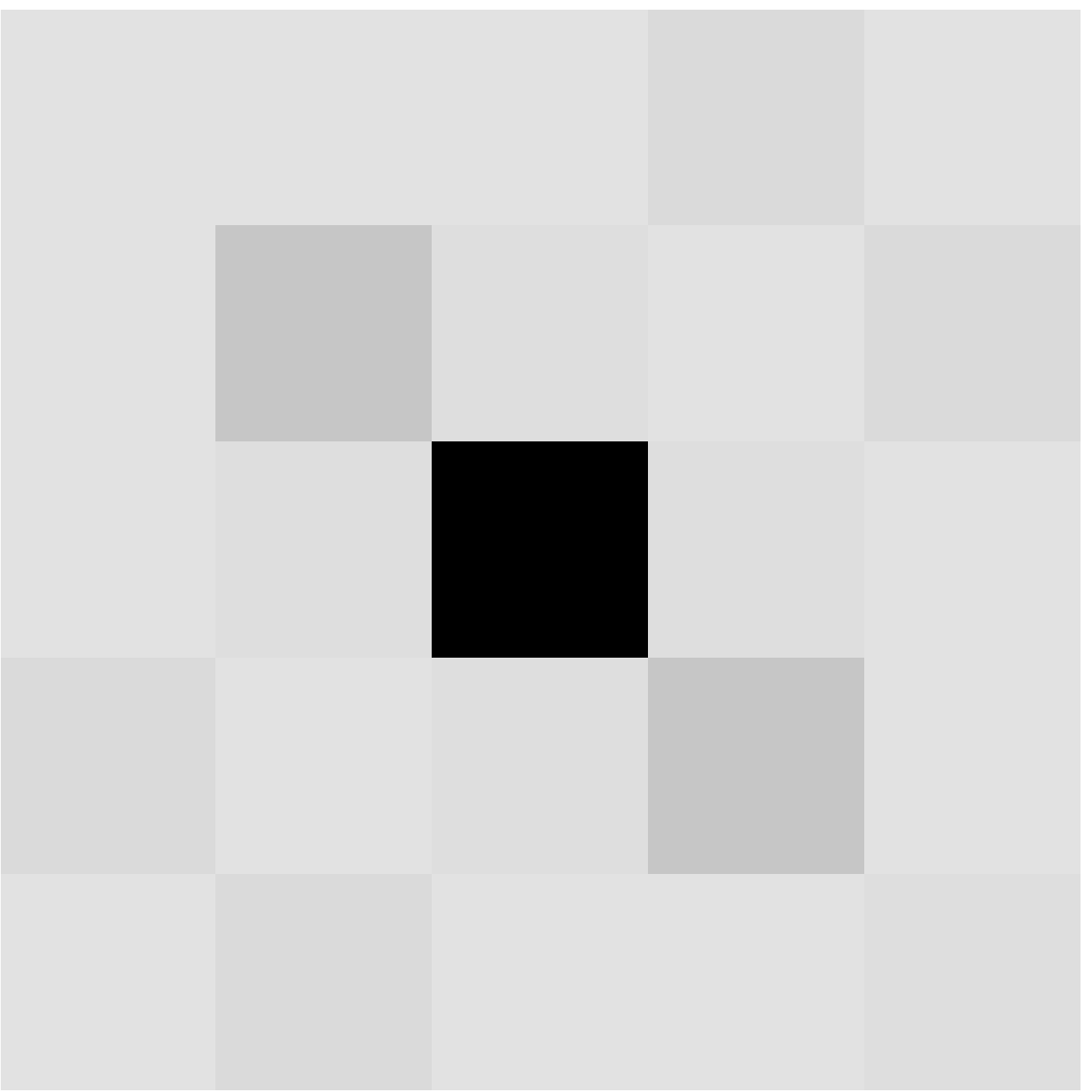} &
         \includegraphics[width=1.5cm]{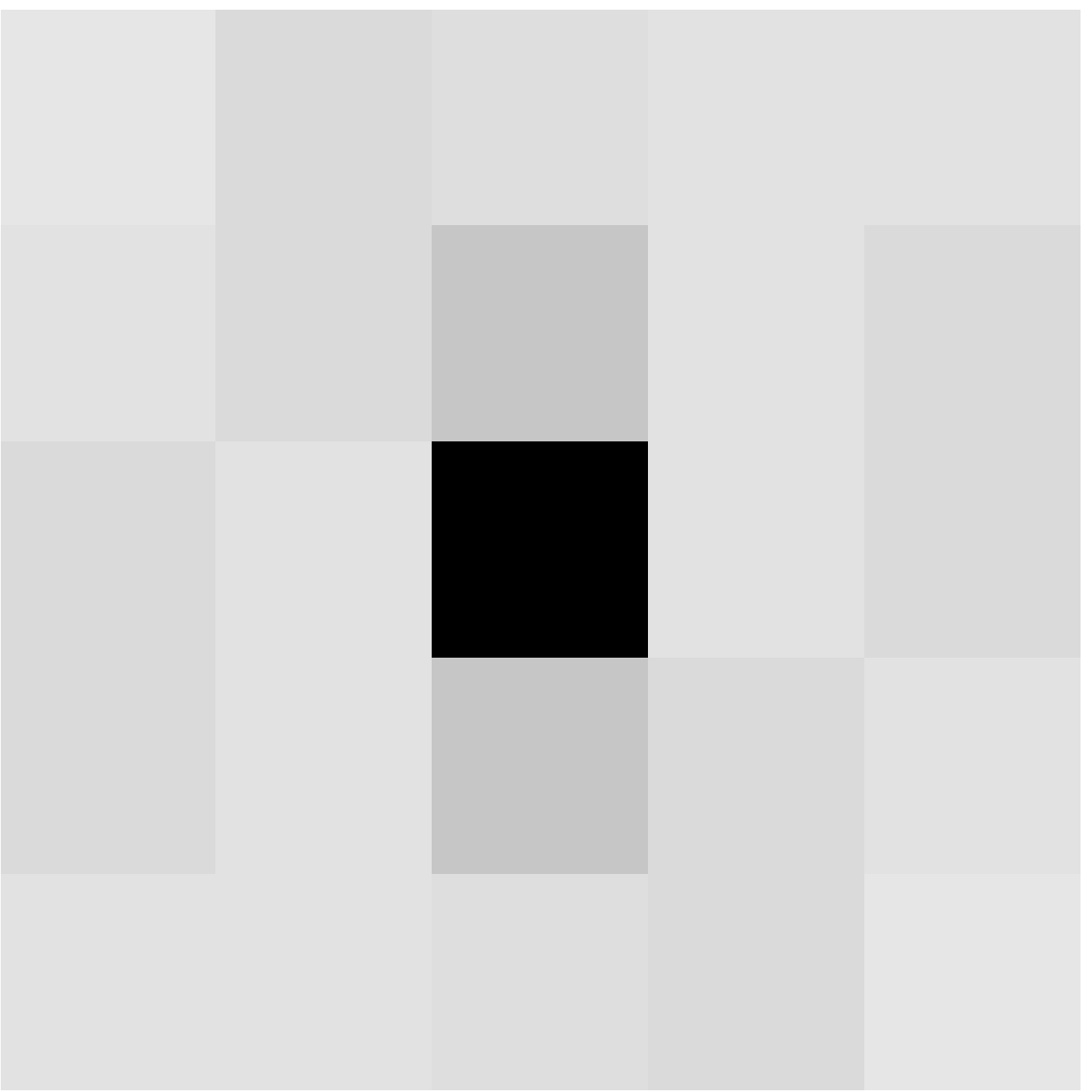} \\

         \includegraphics[width=1.5cm]{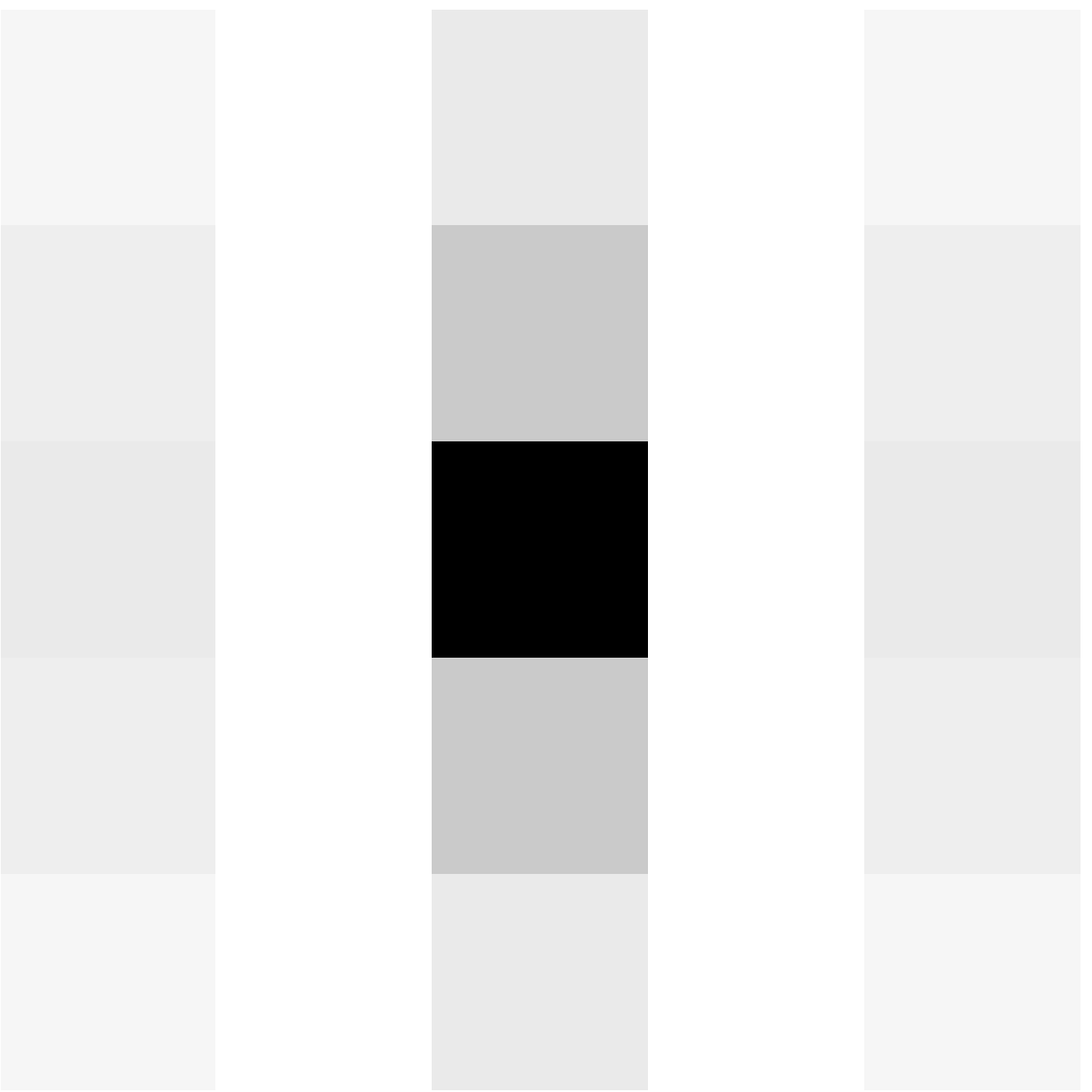} &
         \includegraphics[width=1.5cm]{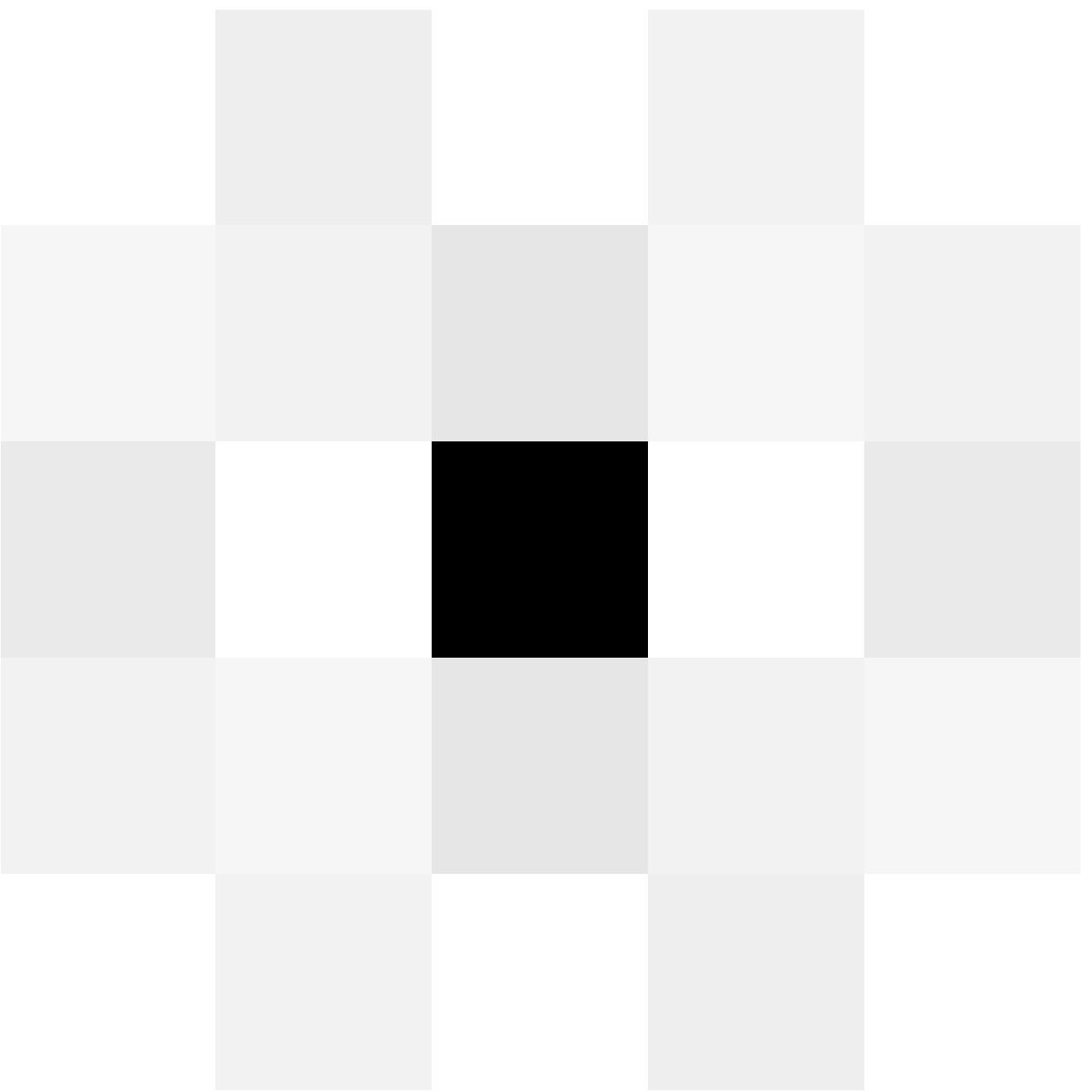} &
         \includegraphics[width=1.5cm]{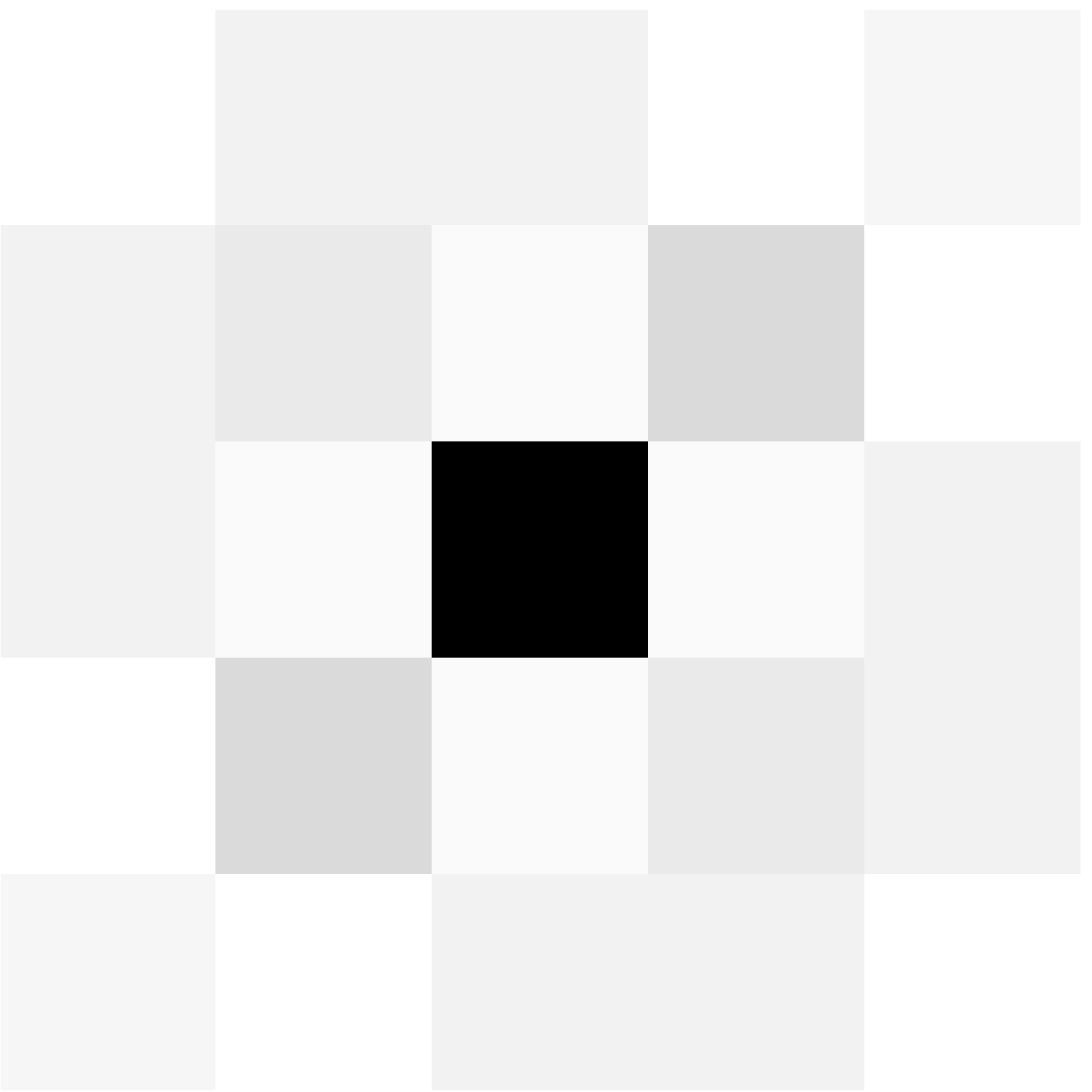} &
         \includegraphics[width=1.5cm]{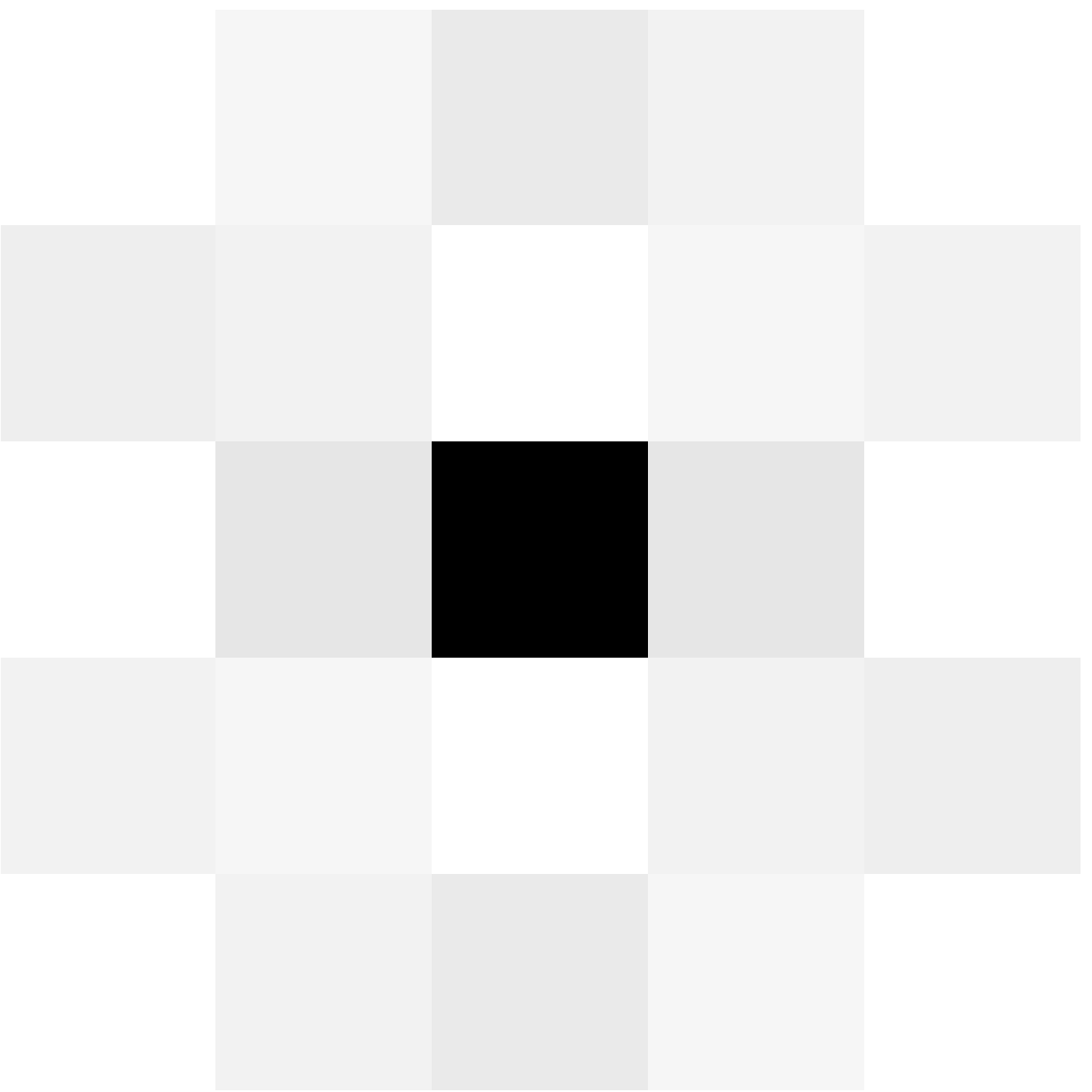} &
         \includegraphics[width=1.5cm]{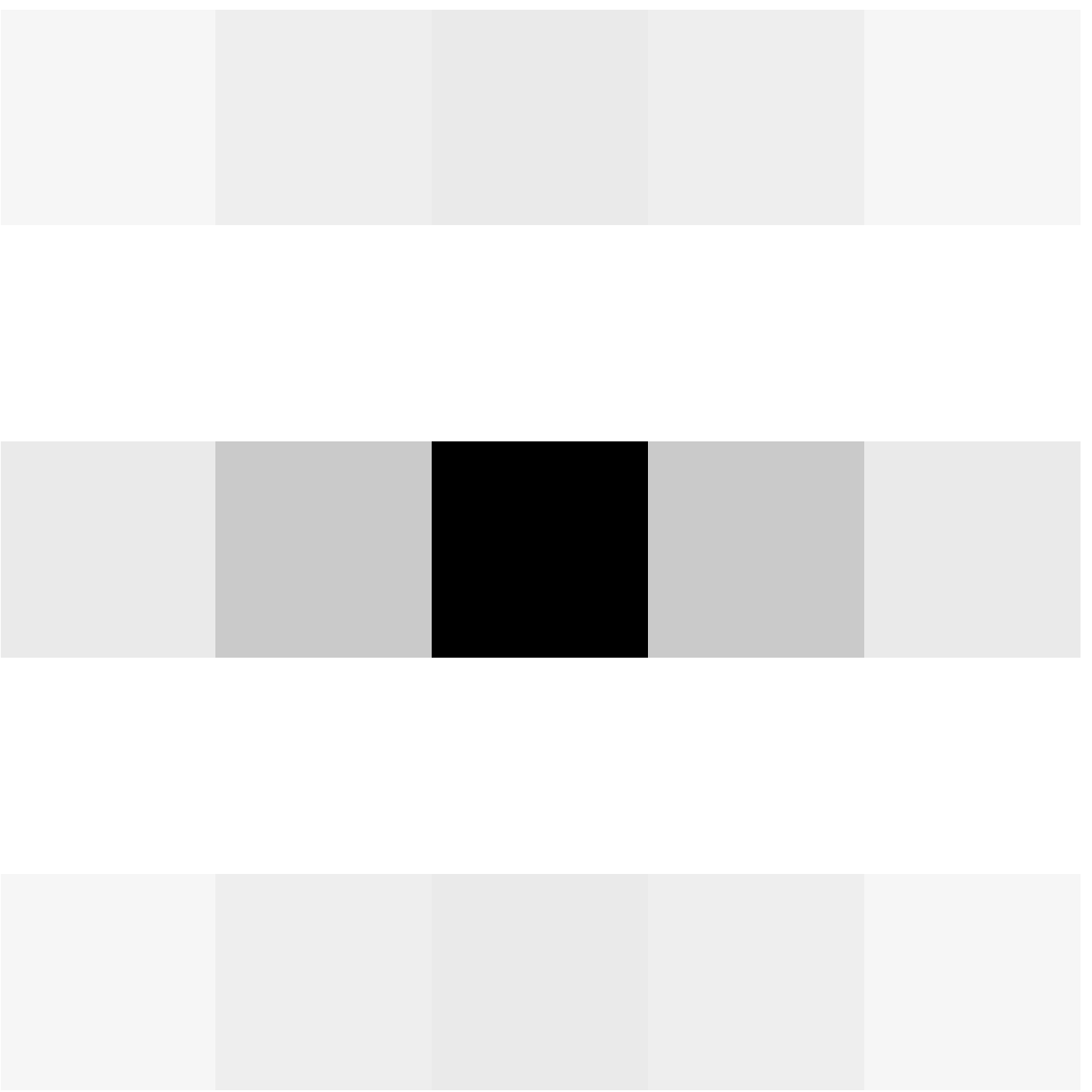} &
         \includegraphics[width=1.5cm]{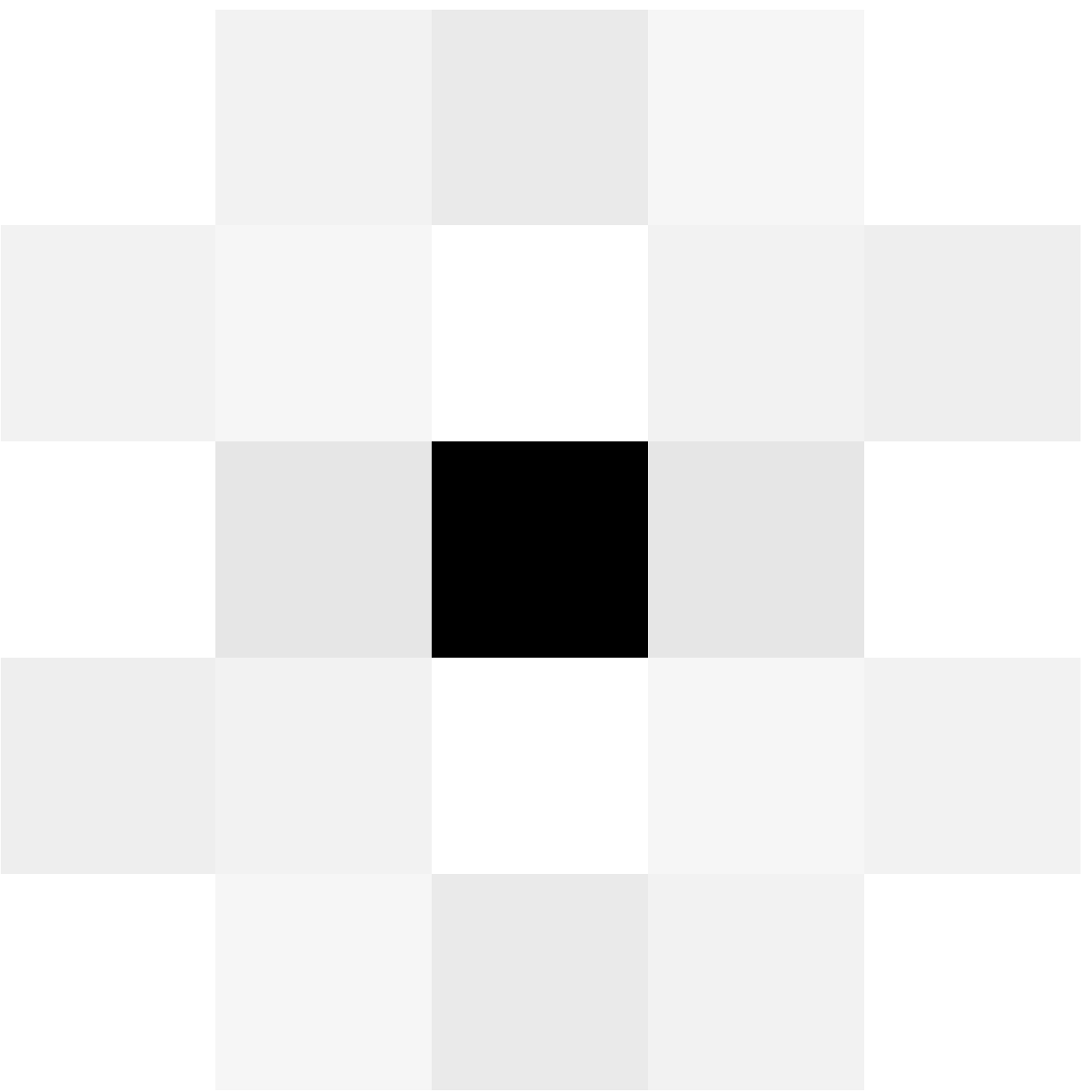} &
         \includegraphics[width=1.5cm]{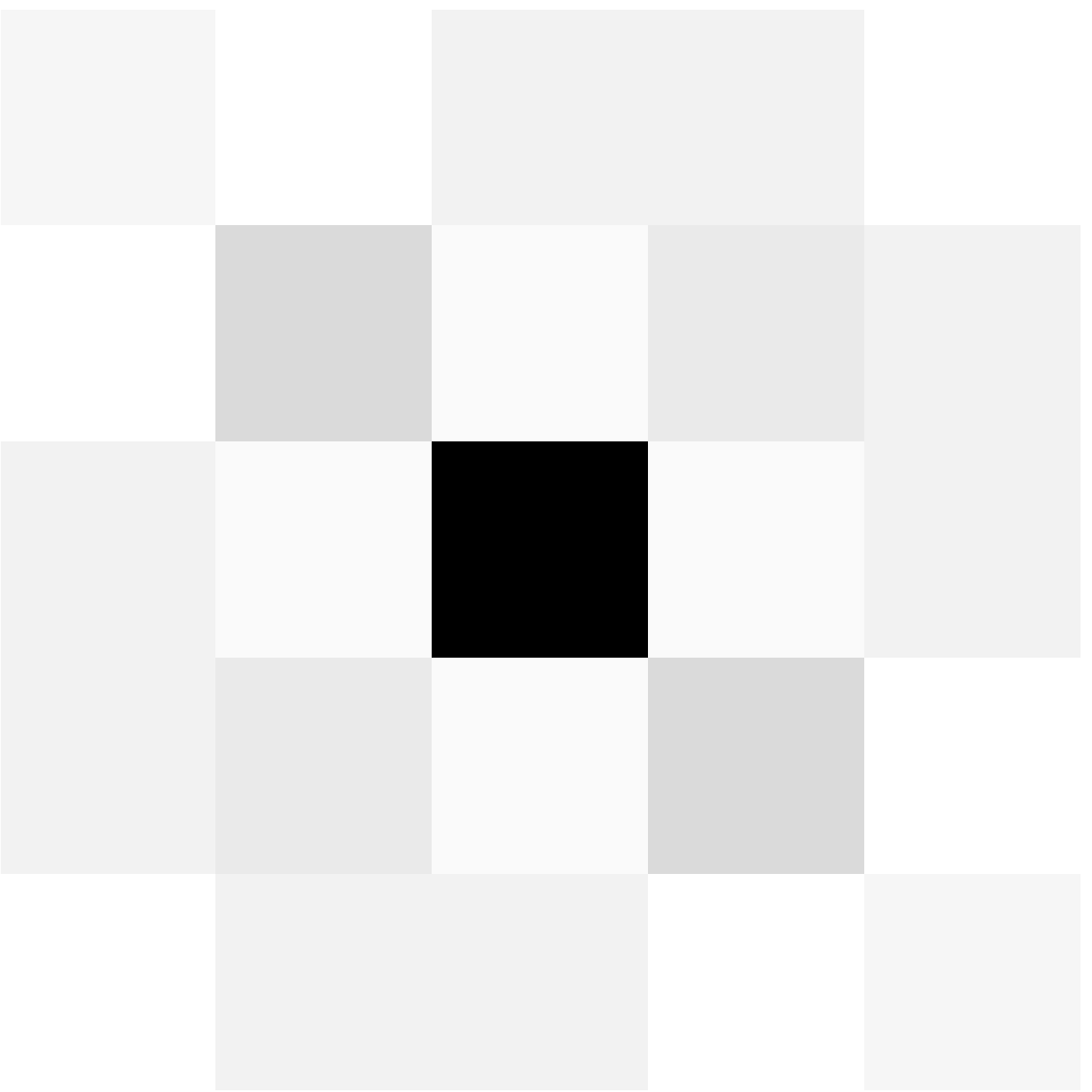} &
         \includegraphics[width=1.5cm]{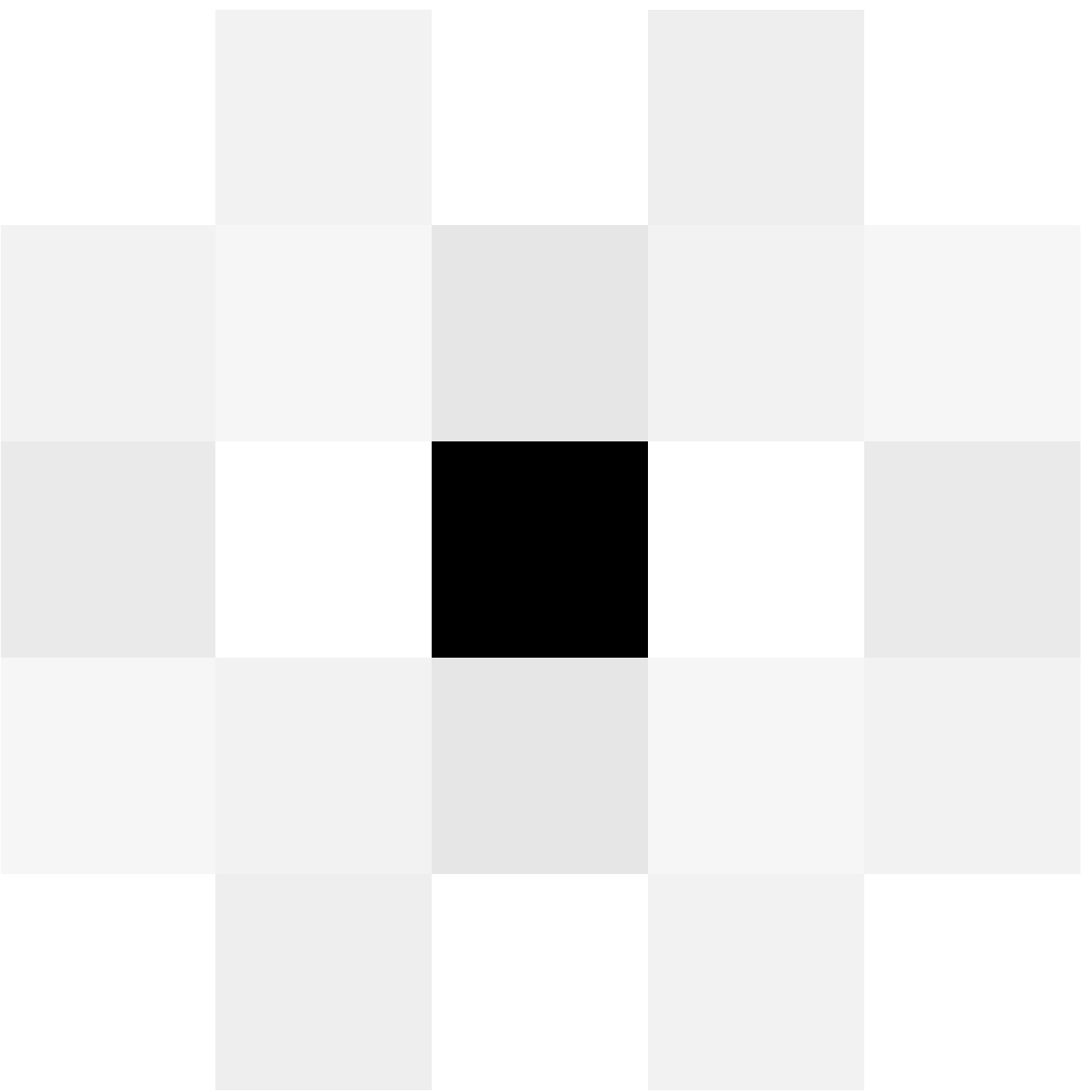} \\
$\alpha = 0$ &
$\alpha =\frac{\pi}{8}$&
$\alpha =\frac{2\pi}{8}$&
$\alpha =\frac{3\pi}{8}$&
$\alpha =\frac{4\pi}{8}$&
$\alpha =\frac{5\pi}{8}$&
$\alpha =\frac{6\pi}{8}$&
$\alpha =\frac{7\pi}{8}$
      \end{tabular}
      \caption{Mutual information among \blue{the central coefficient and} its spatial neighbors in the same subband (intraband) in the steerable wavelet domain. \blue{Darker gray values indicate higher mutual information}. Top row
       shows the results for the different orientations of the finest scale of the natural image database, and bottom row shows the equivalent results for Gaussian noise.}\label{intra}
   \end{center}
\end{figure}

\blue{In our second test, we studied the spatial arrangement of the relations among intraband coefficients since they display the most relevant relations.
 To this end, we computed the mutual information in a 2D $5\times5$ neighborhood for the different orientations and scales.}
Figure \ref{intra}[top] shows the above mentioned results for the set of natural images (finest scale).
\blue{We also provide the relations introduced by the transform (i.i.d. signal, Fig. \ref{intra} [bottom]).}
Similar results were obtained for the other (coarser) scales. Again, \blue{the relations among the signal coefficients are higher than those introduced by the transform.}
Another key issue observed in Fig. \ref{intra} [top] is the specific spatial arrangement of
these relations: the presence of oriented structures in natural
images gives rise to strong anisotropic intraband relations in the different subbands.
Coefficients following these relations are expected to be representative of natural features.
\blue{These mutual information results match recently reported results on autocorrelation of
intraband wavelet coefficients \citep{Goossens09}.}
The results obtained in these experiments will be further used in Section 4 to
design specific kernels that take into account the {\em observed} natural image
relations.

\blue{Summarizing, natural images have singular features in the steerable wavelet domain (Figs. \ref{inter_intra_ori} and \ref{intra}): given a distorted image, enforcing these singular oriented relations among
coefficients in every subband (with the appropriate kernels) will eventually preserve the natural signal relations and remove the noise. Of course,
the bigger the difference between the shape of the intraband relations in signal and noise
the better the results are expected to be.}

\section{Restoring wavelet relations with SVR}

The effect of noise in the wavelet domain is introducing artificial deviations from the original signal
and hiding the natural relations among the coefficients (see an illustrative example in Fig. 3).
\green{In the more general case, the degraded observation, $\mathbf{i_d}$, can be written as the result
of the addition of a certain realization of noise, $\mathbf{n}$, to the original signal, $\mathbf{i}$:}
\begin{equation}
   \green{\mathbf{i_d}=\mathbf{i}+\mathbf{n}}
   \label{adicion}
\end{equation}
\green{ Note that this (convenient) way to state the problem does not necessarily mean that the physical
degradation has to be additive. In fact, the nature of the degradation should ideally be expressed
through a probabilistic noise model that may depend on the original signal, $p(\mathbf{n}|\mathbf{i})$.
The other desirable piece of information is a probabilistic model of the signal, $p(\mathbf{i})$.
However, in most practical situations, the complete probabilistic description of the problem,
i.e. having $p(\mathbf{i})$ and $p(\mathbf{n}|\mathbf{i})$, is not available in analytical form.}

\green{In order to avoid this lack of information, we propose to use the regularization ability of SVRs.
 In this section, first we review the capabilities of the SVR for signal approximation.
Afterwards, general constraints to the SVR parameter space are given for the particular problem
of natural image denoising. Finally, we present an automatic procedure to choose the appropriate SVR parameters
(from the above restricted space) to be used for any combination of image and noise.}

\begin{figure}[t!]
   \begin{center}
      \begin{tabular}{cc}
         \hspace{-0.5cm}\includegraphics[width=0.33\textwidth]{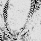} &
         \hspace{-0.2cm}\includegraphics[width=0.33\textwidth]{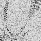}
      \end{tabular}
         \caption{\blue{Effect of noise on the wavelet coefficients. Patch of a subband of a wavelet representation of the original image Barbara (left) and its
         noisy version (right). Darker values indicate higher amplitudes.}}
         \label{figura1}
   \end{center}
\end{figure}

\subsection{Capabilities of SVR for signal estimation}

\green{ Throughout this work,
a wavelet transform, matrix $T$, is applied to the observed image, leading to a set of
(noisy) coefficients, ${\bf y} = T  \cdot {\bf i_d}$. The original set of wavelet coefficients, $\mathbf{x} = T  \cdot {\bf i}$, has
to be estimated from the distorted observation, $\mathbf{y}$.}
\blue{Due to the observed strong intraband relations, we will use the SVR
in the wavelet domain in patches inside each subband. Subbands are decomposed
into non-overlapping $16\times 16$ patches, leading to sets of $N=256$ samples. Now,
given input-output pairs $\{p_i,y_i\}_{i=1}^N$,
whe\-re $p_i$ are the wavelet indices and $y_i$ are the corresponding noisy wavelet coefficients in a patch,
we train the \emph{adaptive} SVR \citep{Camps01NIPS,Navia01,Gomez05} to approximate the signal.}

Let $\boldsymbol{\phi}$ be a non-linear mapping to a higher dimensional
feature space, then the adaptive SVR computes the weights ${\bf w}$ to obtain the estimation, $\hat x_{i}$ =
$\boldsymbol{\phi}^\top(p_i)\mathbf{w}$, by minimizing the following regularized functional:
\begin{eqnarray}
\|{\bf w}\|^2 + \sum_i C_i \ \xi_i
\label{eq:lap}
\end{eqnarray}
subject to $|y_{i}-\boldsymbol{\phi}^\top(p_i)\mathbf{w}| \leq \varepsilon_i + \xi_i$, $\forall i=1,\ldots, N$,
where $\xi_i$ are the magnitude of the deviations of the estimated signal from the observed noisy data
outside the (sample-dependent) insensitivity zones $\varepsilon_i$.
Sample-dependent penalization parameters, $C_i$, tune the trade-off between fitting the model to
the observed noisy data (minimizing the deviations) and
keeping model weights $\|{\bf w}\|$ small (enforcing flatness in the
feature space).

This adaptive SVR differs from the standard formulation \citep{Smola04}, in
two aspects: (1) the loss function given by ($\varepsilon_i,C_i$) is sample-dependent, which
is convenient in wavelet domains where signal and noise variances strongly depend on the subband, and (2)
the usual bias term in SVM formulations has been intentionally dropped to account for the fact that the expected value of wavelet coefficients is zero. The appropriate design of $C_i$ and $\varepsilon_i$ profiles is analyzed in Section \ref{SVMparams}.

Explicitly working with the non-linearity $\boldsymbol{\phi}$ is no longer necessary
since the whole formulation can be expressed in the form of
dot products of the mapping functions called {\em kernels}, \blue{$K(p_i,p_j)$ =
$ \boldsymbol{\phi}(p_i)^\top\boldsymbol{\phi}(p_j)$}.
In this case, the estimation is given by $\mathbf{\hat x}$ = $K \cdot \boldsymbol{\alpha}$, where $\boldsymbol{\alpha}$ is the dual representation of weights $\mathbf{w}$ \citep{Smola04}.
The kernel matrix can be seen as a similarity matrix between samples (or coefficients), and
should reflect the relations between them. Many kernel functions have
been proposed in the literature \citep{Smola04}.
In the image denoising case in wavelet domains, we focus on the basic structure of the
\blue{generalized} Radial Basis Functions (RBF) kernel since the
relationship among the wavelet coefficients corresponding to
spatial neighbors within a subband is local. However, as it will be
analyzed in subsection \ref{SVMparams}, the kernel will be adapted to incorporate
the anisotropic signal relations studied in subsection \ref{steerablerelations}, see Fig. \ref{intra}.

\subsection{General constraints on SVR parameter space in image denoising}
\label{SVMparams}

\green{As stated above, SVR \green{signal approximation} will depend on the penalization parameters, $C_i$, the insensitivities,  $\varepsilon_i$, and the kernel $K$.
In the following, we restrict the range of possible values of these parameters, $\theta = (C_i,\varepsilon_i,K) $, in the particular case of image denoising in wavelet domains:}
\begin{description}
\item[Penalization factor.]
\green{In general, the penalization factor of SVRs should be related to the standard deviation of the signal \citep{Cherkassky04}.
In the denoising problem considered here, the signal variance substantially differs in each wavelet scale. According to this,
it is strictly necessary to set a different penalization factor {\em per} scale, $C_i=C \,k_i$, where $k_i$ is a scale-dependent profile. This profile $k_i$ was obtained by averaging the standard deviation of wavelet coefficients over 100 images from the database used in Section \ref{features}}. This profile was multiplied by a factor, $C$, varied in the range [$10$, $10^4$], \blue{which did not show a strong impact on the results provided a sufficiently large value.} \green{This is consistent with the suggestions reported in \citep{Chalimourda04} in a more general context}. \blue{Note that, for instance, in the examples of the next section (Fig.~\ref{figura1}), indistinguishable results are obtained for a large enough $C$. In our experiments, we found that a reasonable prescription for the global factor on the penalization profile is $C\approx 10^3$.}

\item[Adaptive insensitivity zone.] \green{In general, the insensitivity has to be related to the standard deviation of the noise~\citep{Kwok03}. In transformed domains, the effect of the transform has to be taken into account in order to estimate the corresponding standard deviation. In redundant wavelet representations, this standard deviation is coefficient dependent. Thus it is strictly necessary to introduce a subband-dependent $\varepsilon_i$ profile \citep{Camps01NIPS,Gomez05}. The transformed standard deviations can be estimated either
(1) empirically from noise samples, or (2) computed from the noise covariance matrix if it is known.
In the empirical case, noise samples can be experimentally obtained by applying the noise source to
a large enough set of images, and writing the noise as in Eq. \ref{adicion}.
In our experiments, we used the natural image database used in Section \ref{features}, and we obtained fairly stable
results for the profile by considering 100 images.
In the case that the noise covariance is known, the corresponding matrix in the selected wavelet domain can be obtained
from the noise covariance matrix in the spatial domain, $\Sigma_n$, and the transform $T$ \citep{Stark94}.}
Therefore, the insensitivity profile can be computed as:
\begin{equation}\label{tubito}
\varepsilon_{i} = \tau \ diag(T \cdot \Sigma_n \cdot T^\top)_i^{1/2}
\end{equation}
In the case of \blue{white} noise, $\Sigma_n= \sigma_n^2 \cdot I$, and thus Eq. \eqref{tubito} reduces to:
\begin{equation}\label{tubito2}
\varepsilon_{i} = \tau \ \sigma_n \ diag(T \cdot T^\top)_i^{1/2},
\end{equation}
where $\sigma_n^2$ is the noise variance in the spatial domain, and $\tau$ is a scaling factor \green{to be adapted for each particular image and noise combination}.
The scaling factor, $\tau$, should be in the range [$0.5$, $3$] according to the known relationship between the $\varepsilon$-insensitivity zone
and the noise standard deviation~\citep{Kwok03}. Note that \eqref{tubito} may cope
with colored noise. Considering the off-diagonal elements of the covariance matrix (neglected in \eqref{tubito} and \eqref{tubito2}) would give rise to coupling $\varepsilon$-insensitivities among samples. This issue has been already considered and solved in the context of image coding by using an additional
non-linear transform and a constant $\varepsilon$
in the transformed domain \citep{Camps07}. However, in this paper, we restrict ourselves
to the approximated diagonal case.

\item[Including signal relations in the kernel.] \blue{
In the kernel methods literature, the use of {\em prior} knowledge about the problem can be encoded through bagged, cluster, or probabilistic kernels~\citep{Jebara04,Weston05}. In our case, we propose to take into account image coefficient relations by analyzing a large (representative) database and taking the (oriented) mutual information among samples as core distance measure.} \green{However, using these empirical measures to set the kernels is not straightforward since the kernels have to fulfill Mercer's Theorem \citep{Mercer1905}. According to this,
we propose to use generalized Gaussian kernels. In particular, we fitted anisotropic
Laplacian kernels to the MI measures to consider the intraband oriented relations within each
subband:}
\begin{eqnarray}
  K_\alpha(p_i,p_j) &=&  \exp \big(-((p_i-p_j)^\top G(\alpha)^\top \Sigma^{-1} G(\alpha)(p_i-p_j))^{1/2}\big),
\end{eqnarray}
\green{where $\Sigma$ =
$\left(
\begin{array}{cc}
\sigma_1 & 0 \\
0 & \sigma_2 \\
\end{array}
\right)$, $\sigma_1$ and $\sigma_2$ are the widths of the kernels, $p_i\in {\mathbb R}^2$ denotes the spatial position of coefficient $y_i$ within a subband, and $G(\alpha)$ is the 2D rotation matrix with rotation angle, $\alpha$, corresponding to the orientation of each subband (see Fig.~\ref{fig:gaussians}).
Note that these set of
oriented kernels describe the signal relationships that emerge from experiments in Section \ref{features} (cf. Fig. \ref{intra}[top]).}

\begin{figure}[t!]
\begin{center}
\begin{tabular}{ccccccccc}
\epsfig{figure=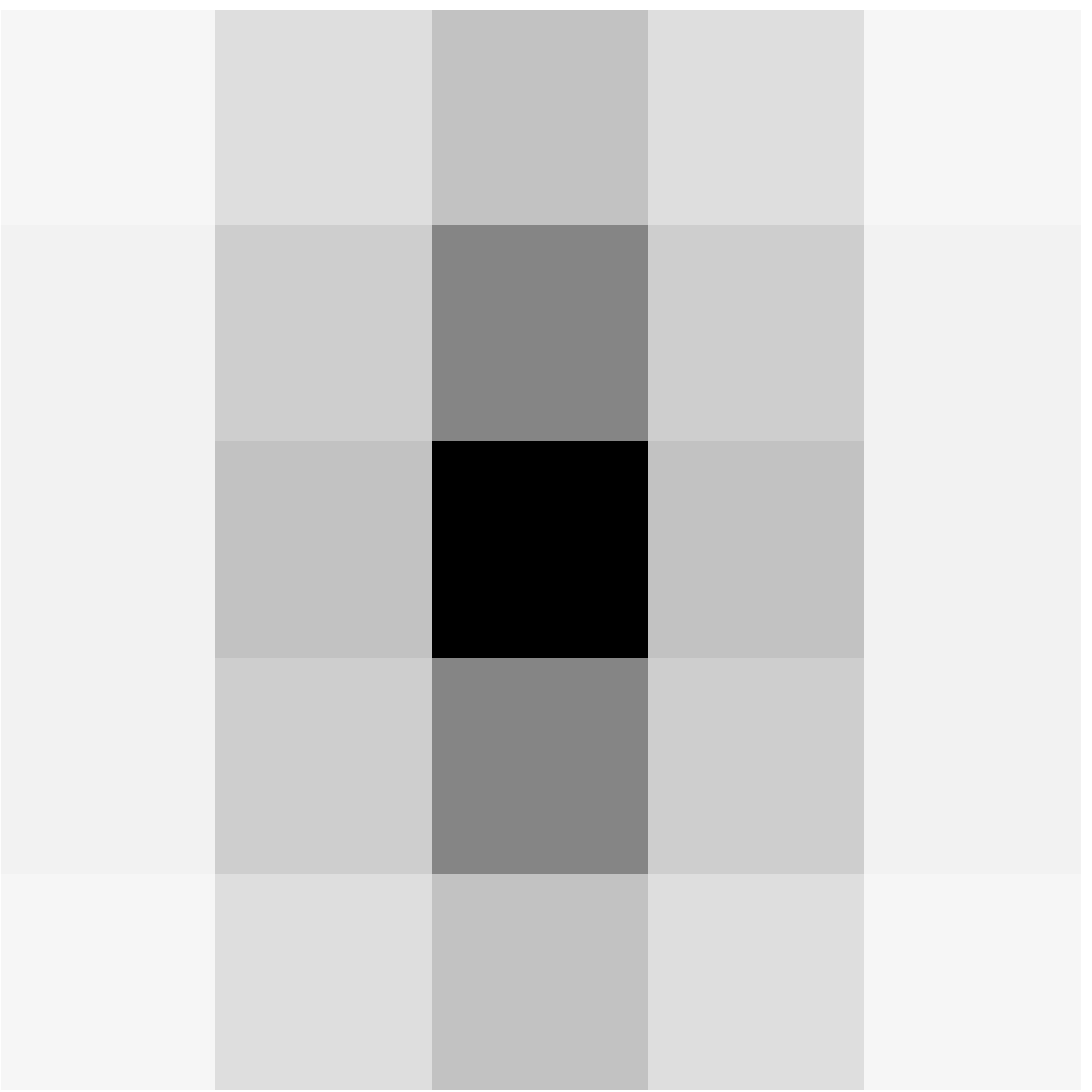,width=1.5cm}	 &
\epsfig{figure=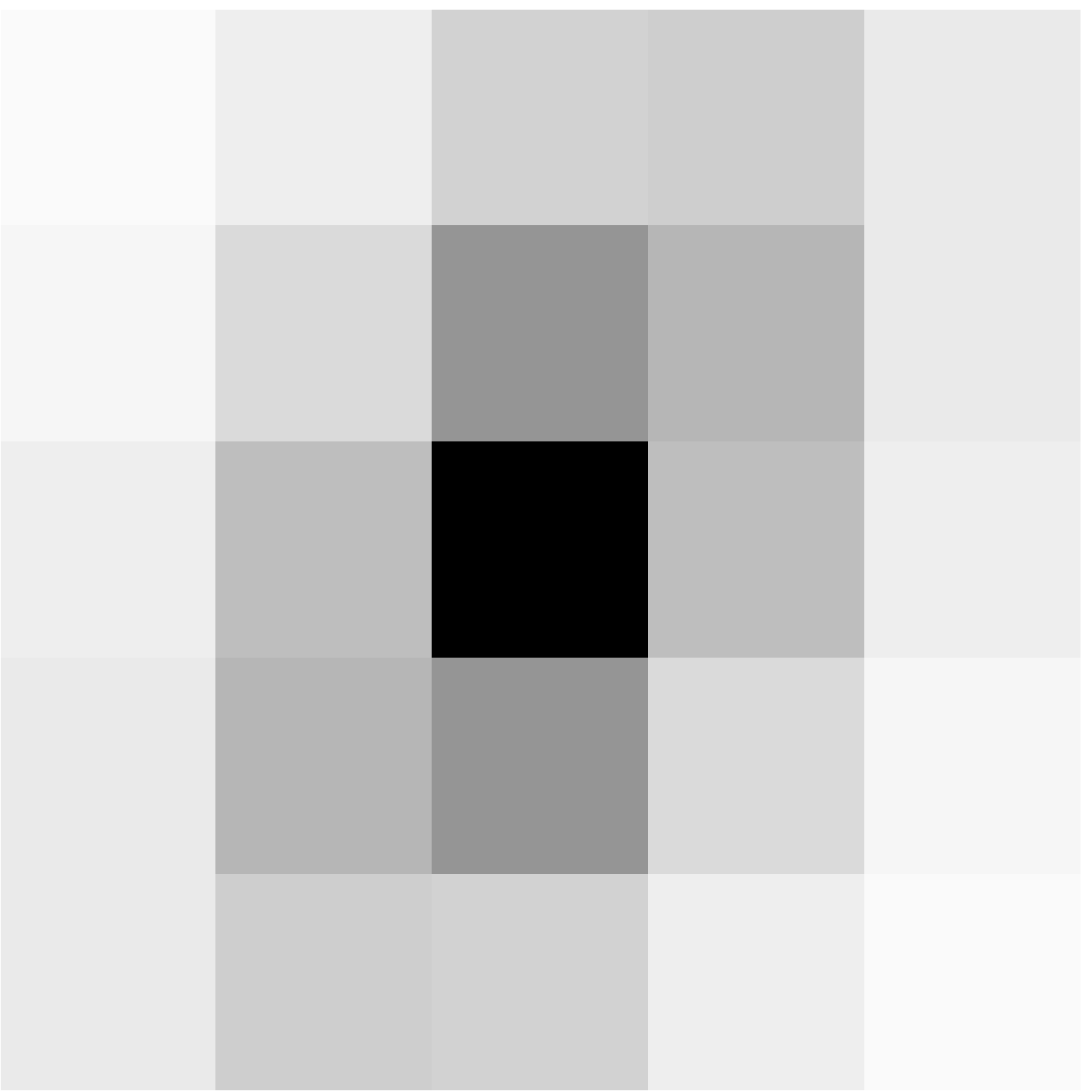,width=1.5cm}	 &
\epsfig{figure=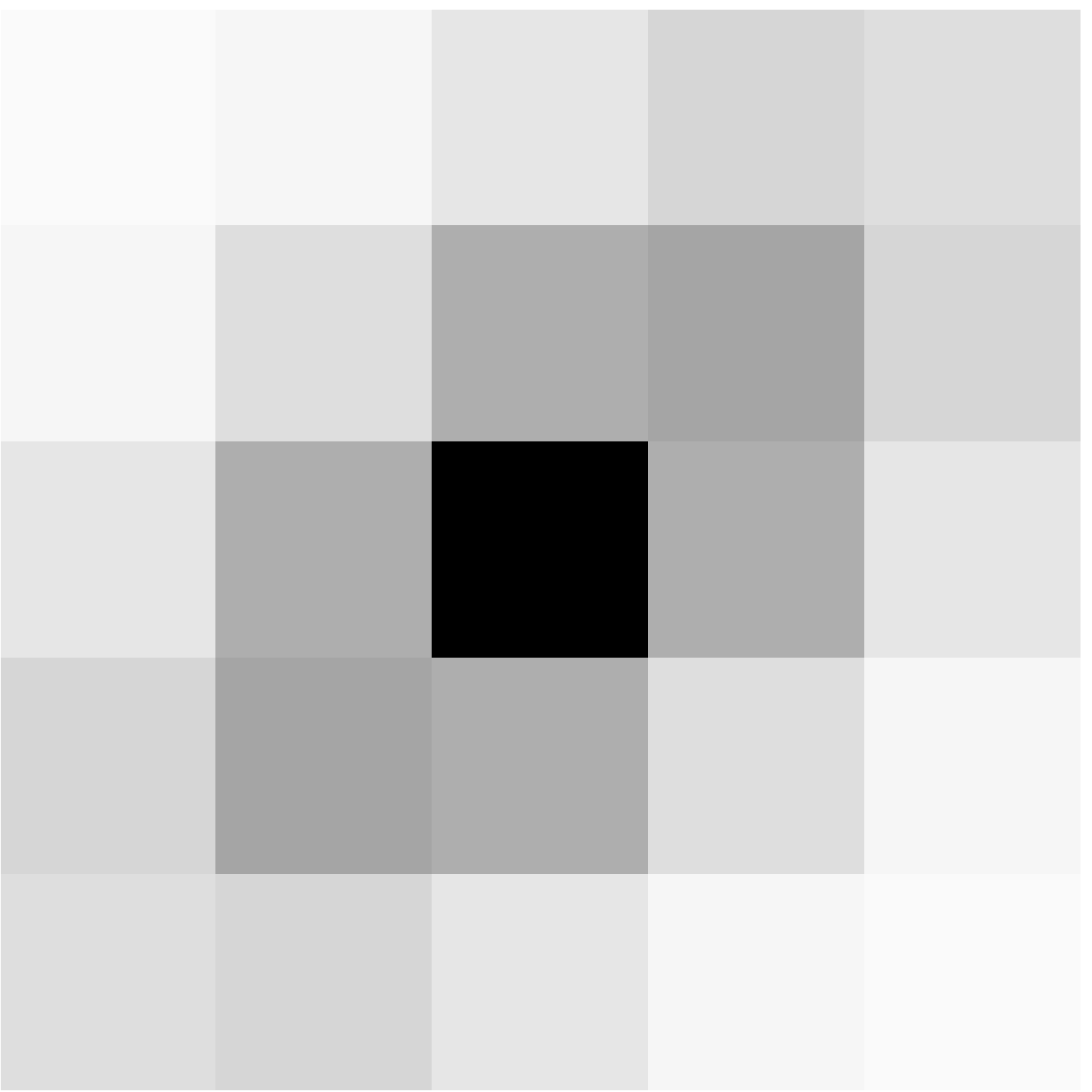,width=1.5cm}	 &
\epsfig{figure=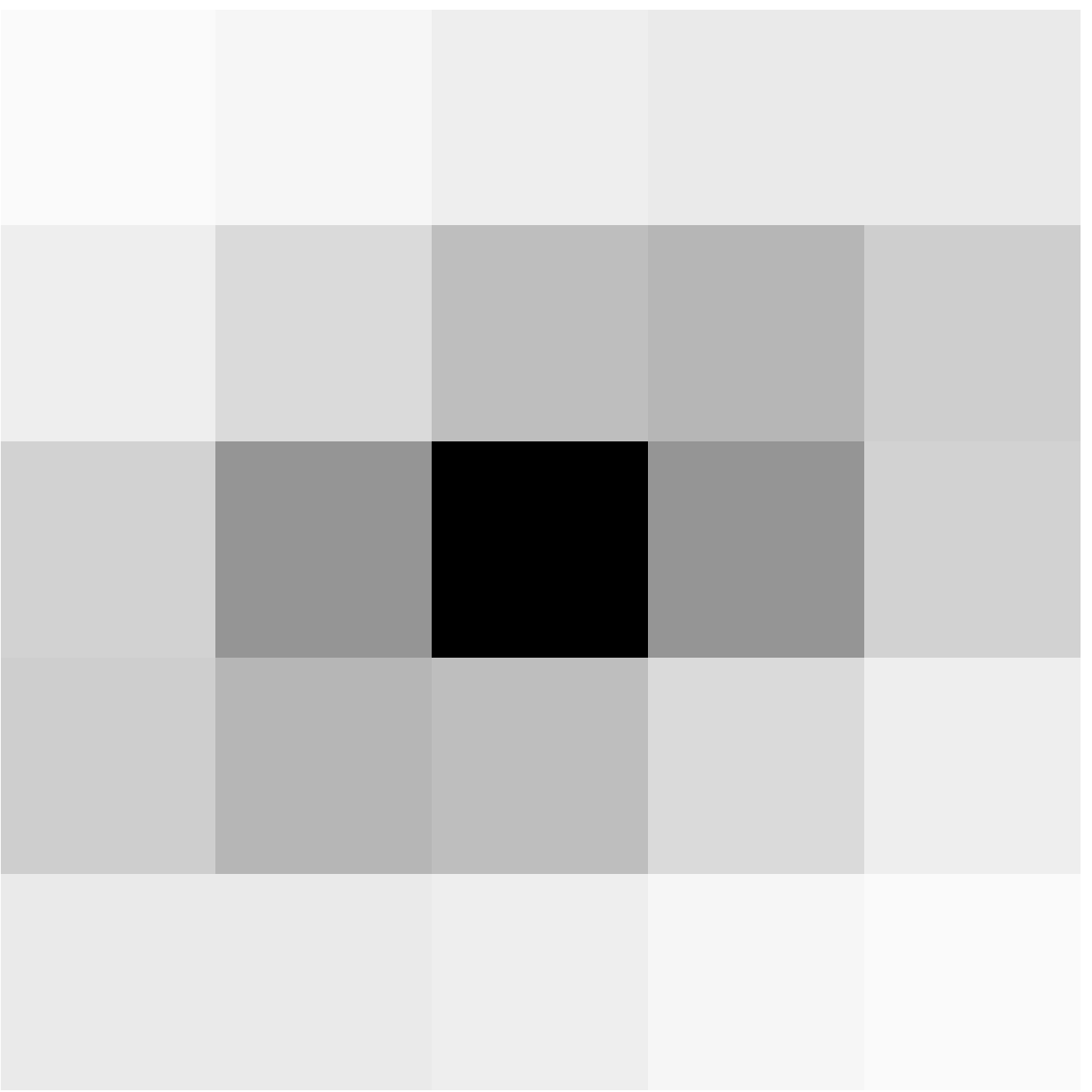,width=1.5cm}	 &
\epsfig{figure=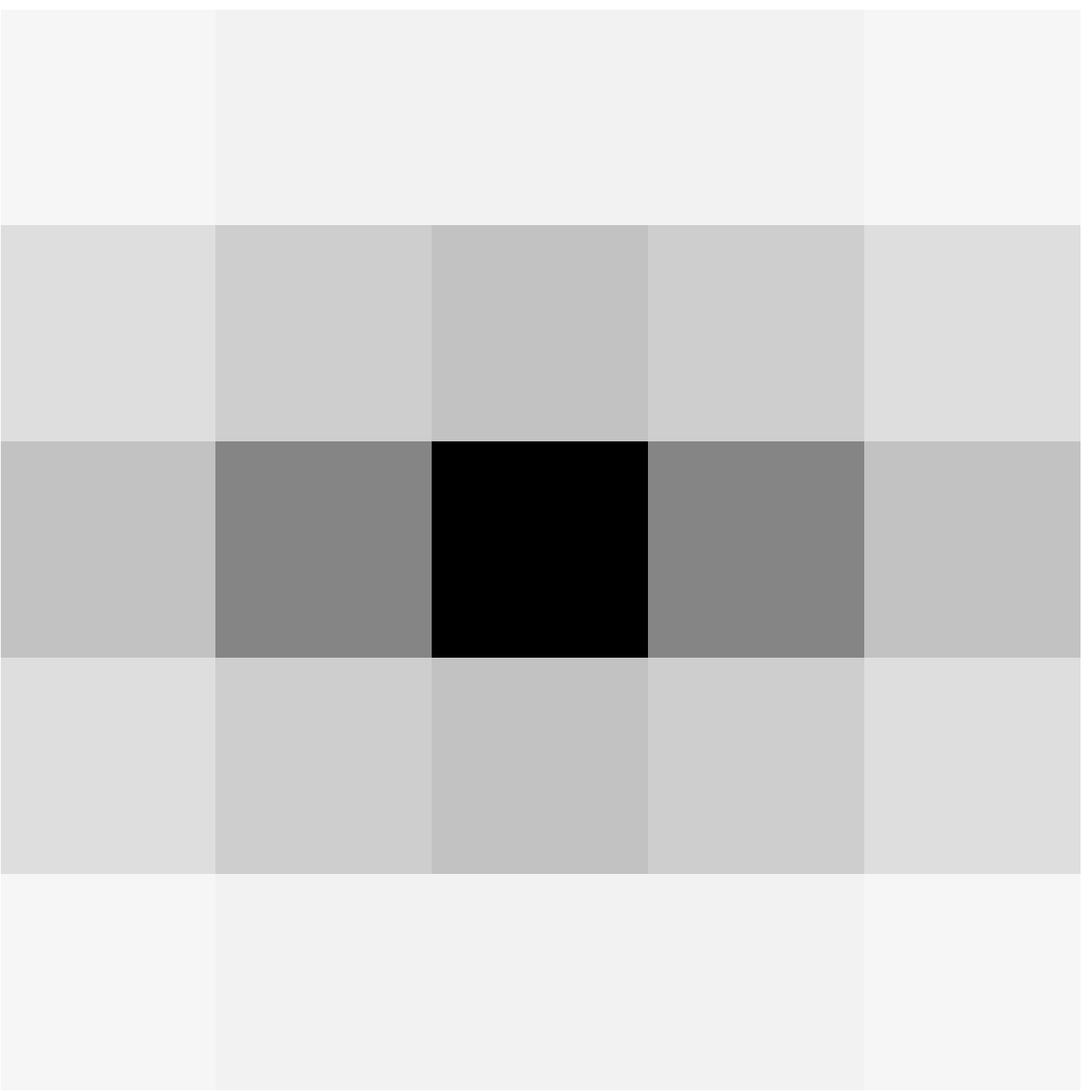,width=1.5cm}	 &
\epsfig{figure=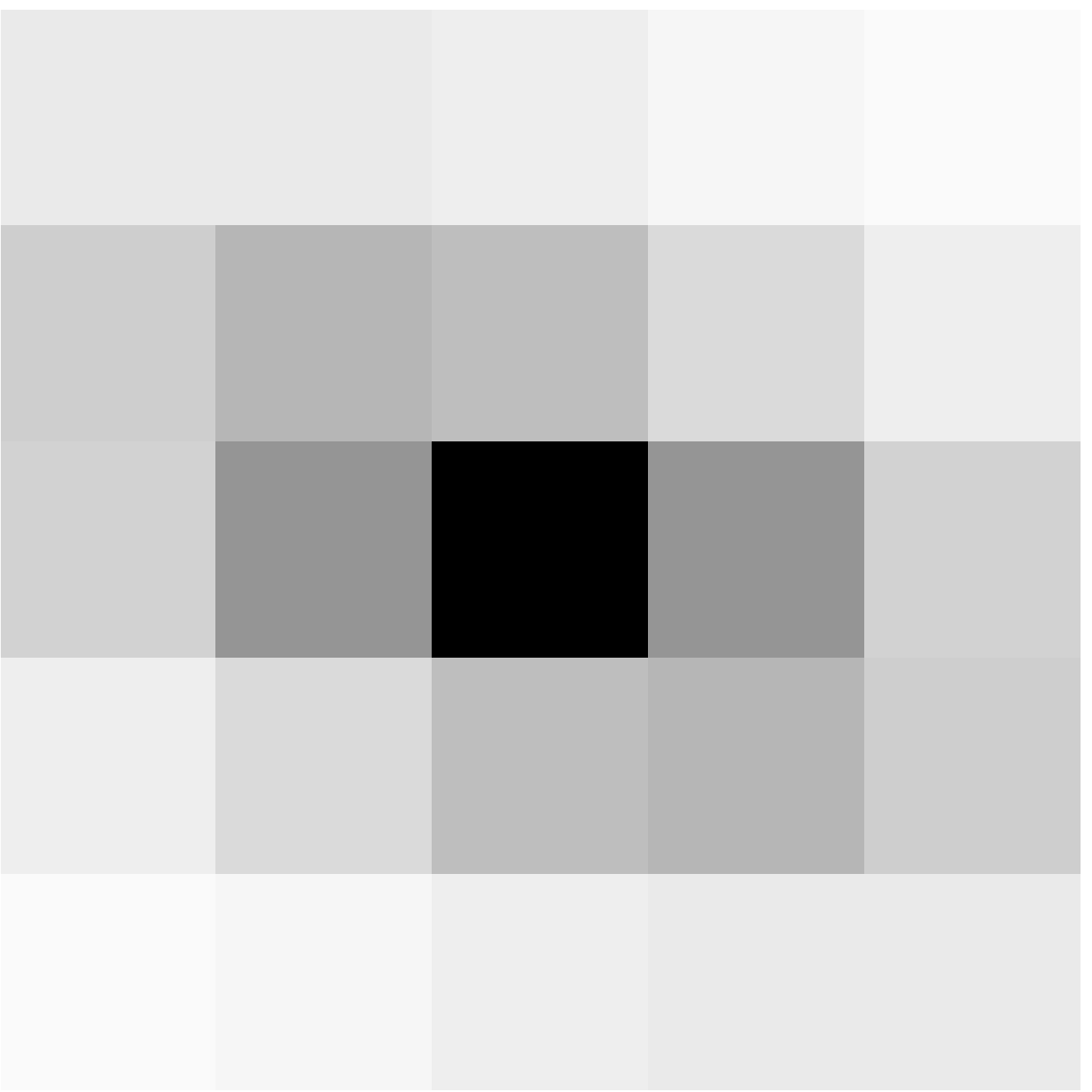,width=1.5cm}	 &
\epsfig{figure=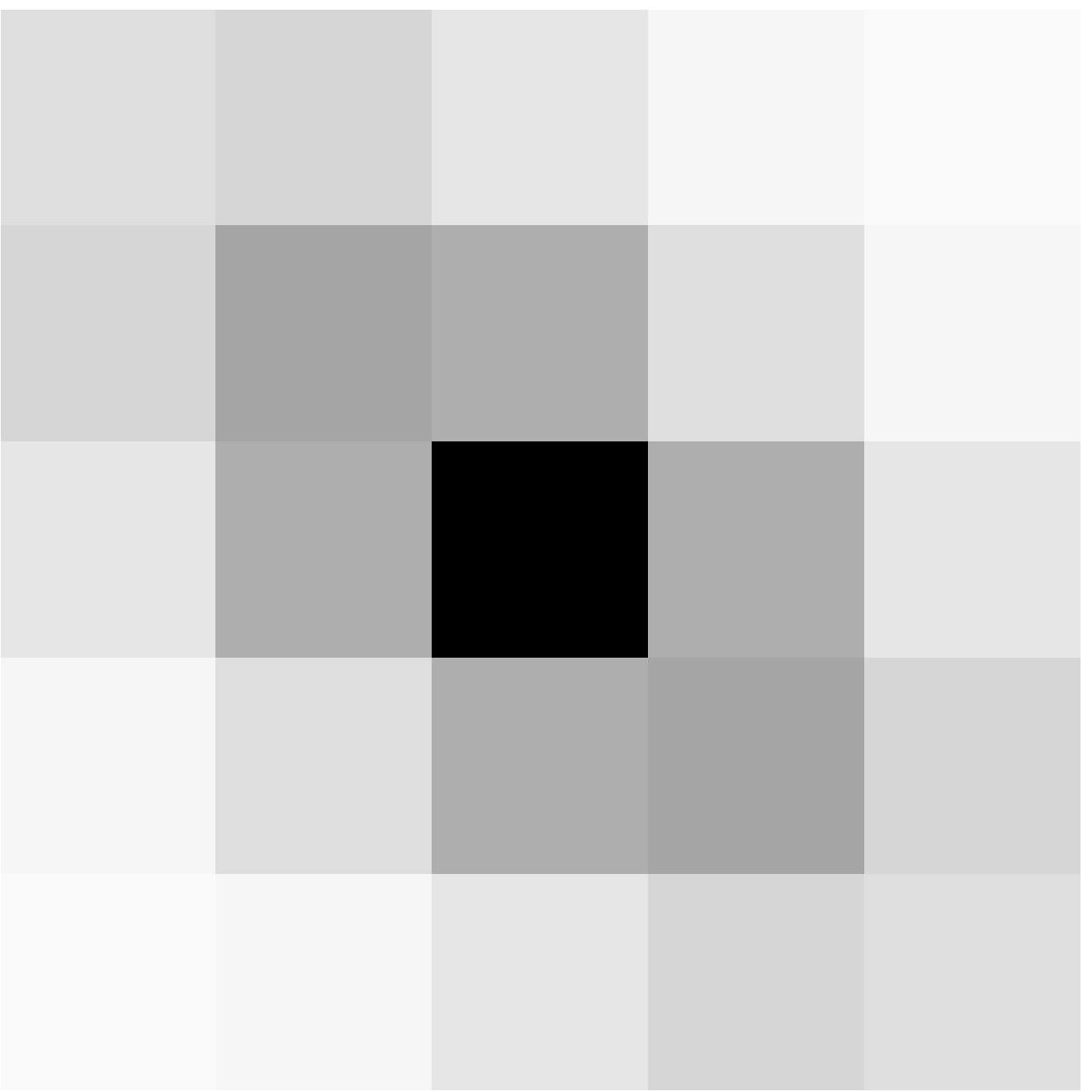,width=1.5cm}	 &
\epsfig{figure=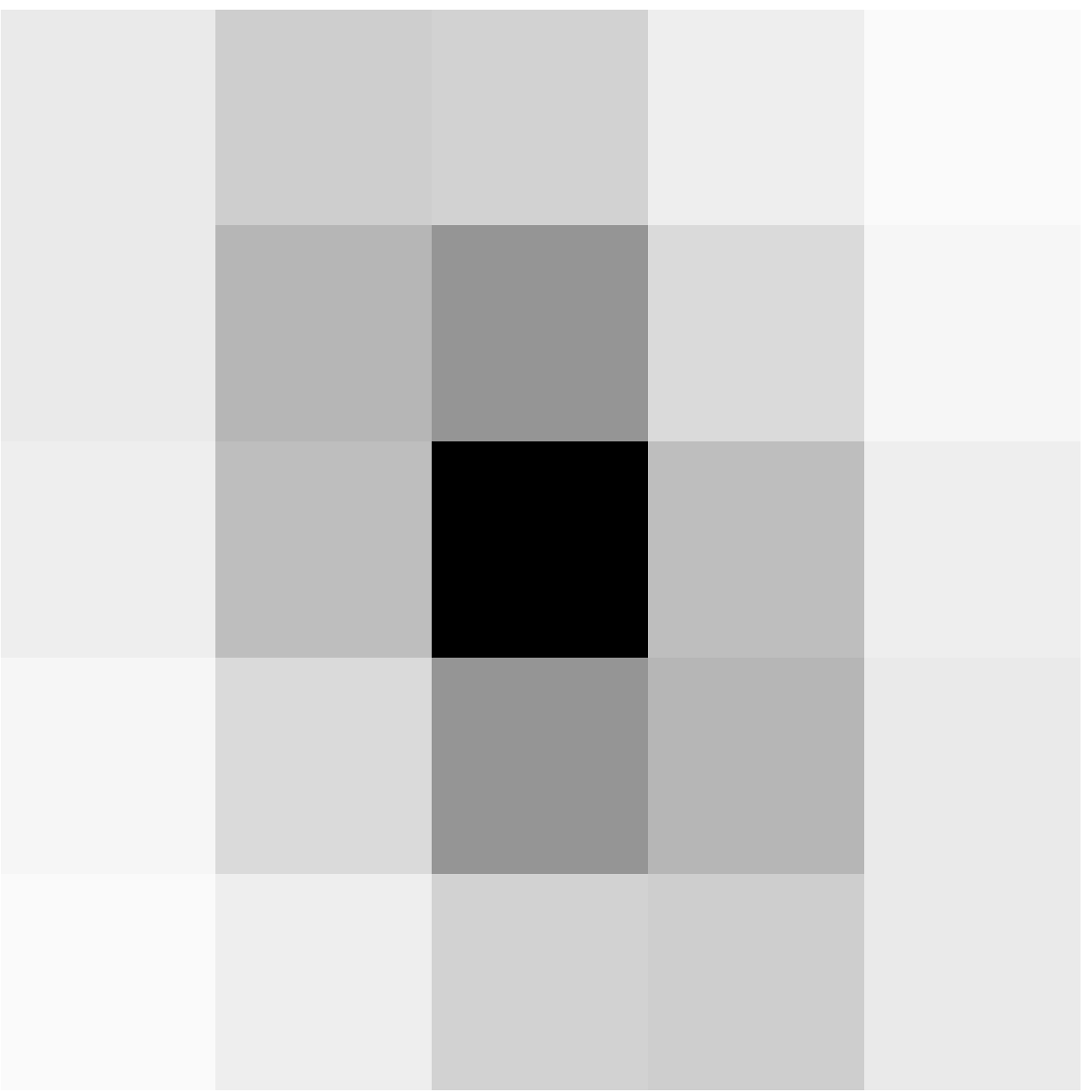,width=1.5cm}	 & \\
$\alpha = 0$ &
$\alpha =\frac{\pi}{8}$&
$\alpha =\frac{2\pi}{8}$&
$\alpha =\frac{3\pi}{8}$&
$\alpha =\frac{4\pi}{8}$&
$\alpha =\frac{5\pi}{8}$&
$\alpha =\frac{6\pi}{8}$&
$\alpha =\frac{7\pi}{8}$&
\end{tabular}
\end{center}
\caption{Anisotropic kernel functions used in the support vector regression method for the eight considered orientation subbands.}
\label{fig:gaussians}
\end{figure}

We obtained proper values for the widths $\sigma_1$ and $\sigma_2$
by fitting the above kernel to the MI measures among coefficients described in Section \ref{features} ($\sigma_1 = 2 \sigma_2$, and $\sigma_1 = 4.8$ in spatial coefficient units).
The kernel was further normalized in the standard way.
\blue{Since this width comes from direct measures from images, it describes a fundamental property
of natural images so it can be kept fairly constant.
}

\end{description}

\green{The conclusion of this section is that in the case of image denoising in wavelet domains,
an appropriate analysis of the signal variance, the noise variance, and the relations
among the wavelet coefficients of the signal can be used to strongly reduce the dimensionality of the
SVR parameter space. After this analysis, the only SVR parameter that remains fixed is the global scaling, $\tau$, to be applied to the insensitivity profile.}

\subsection{Procedure for automatic SVR selection}
\label{procedure3}

\green{In the more general case, applying SVRs with a given set of parameters, $\mathbf{\theta}$, to
a noisy image leads to a certain image estimate, $\hat{\mathbf{i}}_\theta = T^{-1} \cdot \hat{\mathbf{x}}_\theta$.
From this image estimate, and the convenient additive notation for the noise (Eq. \eqref{adicion}), a
noise estimate can be obtained: $\hat{\mathbf{n}}_\theta = \mathbf{i_d} - \hat{\mathbf{i}}_\theta$.
In this section we propose a procedure to select the SVR parameters, $\theta$, that better approximates the noise free image, using the available information.}

\green{In the more general situation the only available information is the noisy image.
However, as stated above, denoising methods usually assume that additional probabilistic information
on the signal and noise is available: $p(\mathbf{i})$ and $p(\mathbf{n}|\mathbf{i})$. Note that
this knowledge is equivalent to the knowledge of the joint signal and noise distribution since
$p(\mathbf{i},\mathbf{n}) = p(\mathbf{n}|\mathbf{i}) \, p(\mathbf{i})$.}

\green{Let us momentarily assume that this information is available to propose the general procedure
to set the SVR parameters. Afterwards, we will relax the requirements by considering an approximation
that can be easily applied in practical situations.}

\green{In order to enforce solutions that closely follow the (assumed to be known) statistics of
signal and noise, we propose to select the SVR that minimizes the $k$-th order Kullback-Leibler
divergence (KLD) \citep{Cover91} between the joint PDF of signal and noise, and the joint
PDF of the estimated signal and the estimated noise:}

\begin{equation}
\green{\theta^\ast = \arg~\min_{\theta}\bigg\{ \, D_{KL} \big[ \, p(\hat{\bf i}_\theta,\hat{\bf n}_\theta) \, \parallel \,  p({\bf i},{\bf n}) \, \big]\bigg\}}
\label{optim1}
\end{equation}

\green{The underlying idea is that the SVR that minimizes the divergence between the above PDFs
is the one that better captures the features of the true signal and better removes the degradation.}

\green{Although in ideal situations the application of this procedure would obtain the best results in statistical
terms, in practical situations the full probabilistic description of the problem is not available.
A number of approximations are done in practical situations. For instance, thermal noise in CCD cameras
is not independent from the input signal since diffusion increase with the irradiance.
However, thermal noise is usually assumed to be independent of the input signal. Additional assumptions as
additivity or certain analytical marginal PDF of the noise are also widely used.}

\green{In our case, we assume independence between signal and noise:
\begin{eqnarray}
p(\mathbf{i},\mathbf{n}) = p(\mathbf{i}) \, p(\mathbf{n})
\end{eqnarray}
However, no analytical model for these PDFs is assumed.
Under this independence assumption, it is easy to see that eq. \ref{optim1} reduces to:}
\begin{equation}
\green{\theta^\ast = \arg~\min_{\theta}\bigg\{ D_{KL} \big[ \, p(\hat{\bf i}_\theta) \, \parallel \,  p({\bf i}) \, \big] + D_{KL} \big[ \, p(\hat{\bf n}_\theta) \, \parallel \,  p({\bf n}) \, \big]\bigg\}}
\label{optim}
\end{equation}
\green{This means that the selected SVR parameters are those that give rise to both signal and noise estimates
probabilistically similar to the true signal and noise respectively. Note that this similarity does not require
analytical models of the PDFs since it can be computed from histograms (or signal and noise samples).}

\green{Of course, the independence assumption does not hold in general, however, as it will be shown in
section \ref{Validation}, this is not a critical fact for a good behavior of the method even in
non-additive cases in which the noise is clearly signal-dependent.
Moreover, the independence assumption simplifies the practical application of the criterion for SVR selection since,
for a limited number of samples, histogram estimations of $p(\mathbf{i})$ and $p(\mathbf{n})$ are far
more reliable than histogram estimations of $p(\mathbf{i},\mathbf{n})$, which implies the duplication
of the dimensionality (in an already high dimensional situation).}

\green{In the examples throughout the paper we restricted ourselves to second order KLD measures due
to the lack of samples, yet capturing the second order structure of signal and noise.
The optimization in Eq. \eqref{optim} was carried out by exhaustive search.}

\vspace{0.4cm}

\subsection{\green{Summary of the proposed denoising method}}

\green{The proposed denoising method can be summarized as follows. First the noisy image is
transformed by a steerable wavelet filter bank. Then, a set of SVRs is applied to the patches of
the subbands of the transform. These SVRs use the profiles for the penalization factor and the insensitivity
computed from signal and noise samples as described in section \ref{SVMparams}.
The SVRs use the kernel based on MI that captures signal relations in the wavelet domain as described in section \ref{SVMparams}.
While the scaling of the penalization profile and the kernels are kept fixed as indicated in section \ref{SVMparams},
the scaling of the insensitivity profile is automatically selected following the procedure described
in section 3.3.}

\section{\green{Behavior of the proposed method}}

\green{In this section, we show an illustrative example of how the SVR parameters affect
the estimated solution. Moreover we validate the proposed automatic procedure for SVR selection
considering examples with different noise sources including non-additive and signal dependent cases.}

\subsection{Impact of SVR parameters in image denoising}

\begin{figure}[t!]
    \begin{center}
       \includegraphics[width=8.2cm]{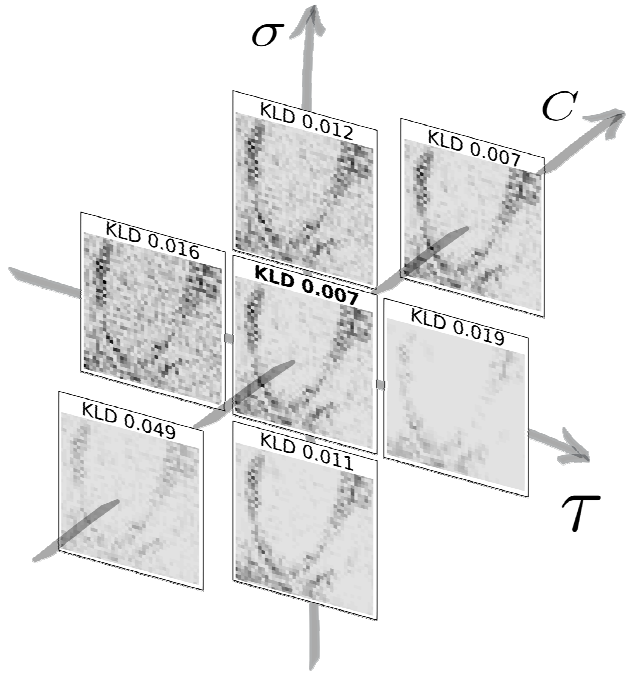}\\
    \end{center}
\vspace{-0.5cm}
\caption{Effect of SVR parameters on the noisy wavelet patch of Fig.~\ref{figura1}. The values of the KL-divergence criterion between the
         estimated and the actual PDFs of noise and signal are given in each case (see text in Section 3.3).}\label{figura2}
\vspace{-0.4cm}
\end{figure}

As stated above, the regularization behavior of the SVR depends on $\theta=(C_i,\varepsilon_i,K)$.
Here we show the qualitative effect of the global penalization scaling $C$, the global insensitivity scaling $\tau$, and the kernel width $\sigma$ assuming a generalized RBF kernel.
Figure~\ref{figura2} shows the qualitative effect of SVR estimation as a
function of these parameters. Compare the results with the original and noisy subbands shown in Fig.~\ref{figura1}.

Increasing the kernel width, $\sigma$ (vertical direction), introduces
too strong relations among coefficients in such a way that spurious energy appears in the
reconstruction.
Increasing the insensitivity, $\tau$ (horizontal direction),
a sparser solution is obtained, leading to information loss and thus relevant
features of the signal are discarded.
On the contrary, a too small insensitivity value gives rise to overfitting, and hence noise is not removed.
Small values of the $C$ parameter gives rise to over-regularized estimations. Large enough values of $C$
give rise to similar behavior (see comments in Section \ref{SVMparams}).

\blue{Of course, interactions among these parameters occur, and have been studied in other contexts
elsewhere \citep{Chalimourda04,Cherkassky03,Cherkassky04}.
In the image denoising case, the deviation from an \emph{appropriate} solution in combined directions
of the parameters gives rise to different solutions that combine the negative effect of the
departure in each direction.}

The above example suggests that \emph{appropriate} SVRs can certainly
recover the underlying structure of the original signal from the noisy
observation, which is the rationale of the proposed method.

\subsection{Validation of the automatic procedure for SVR selection}
\label{Validation}
In this section, we validate the previous SVR selection procedure in two different ways.
Firstly, note that KLD values in the example of Fig.~\ref{figura1}
\emph{qualitatively} illustrate the usefulness of the proposed procedure: the minimum divergence
solution (central subband patch) gives also a reasonable trade-off between
smoothness and detail preservation of the original subband patch.

\green{Secondly, we \emph{quantitatively} show that the SVR that enforces the similarity
between the estimated and actual signal and noise joint PDFs (in KLD terms) is not far from the SVR that maximizes the structural similarity between the estimated and the original image.
In order to do so, we compare the KLD measures for different SVRs, with the corresponding
distortion measured with the Structural SIMilarity (SSIM) index \citep{Wang04}. The SSIM index is a widely used similarity measure, which is better related to human quality assessment than Euclidean measures, such as MSE or PSNR. Note that while KLD values are available in real situations (provided the noise histogram and a generic natural images histogram are known), distortion measures are not available since the original image is unknown. Consequently, the SSIM results next presented are for mere comparison purposes.}

\begin{figure}[p!]
\begin{center}
\vspace{-1cm}
    \begin{tabular}{c@{}c}
     \multicolumn{2}{c}{Gaussian Noise} \\
     \multicolumn{2}{c}{\includegraphics[width=0.48\textwidth]{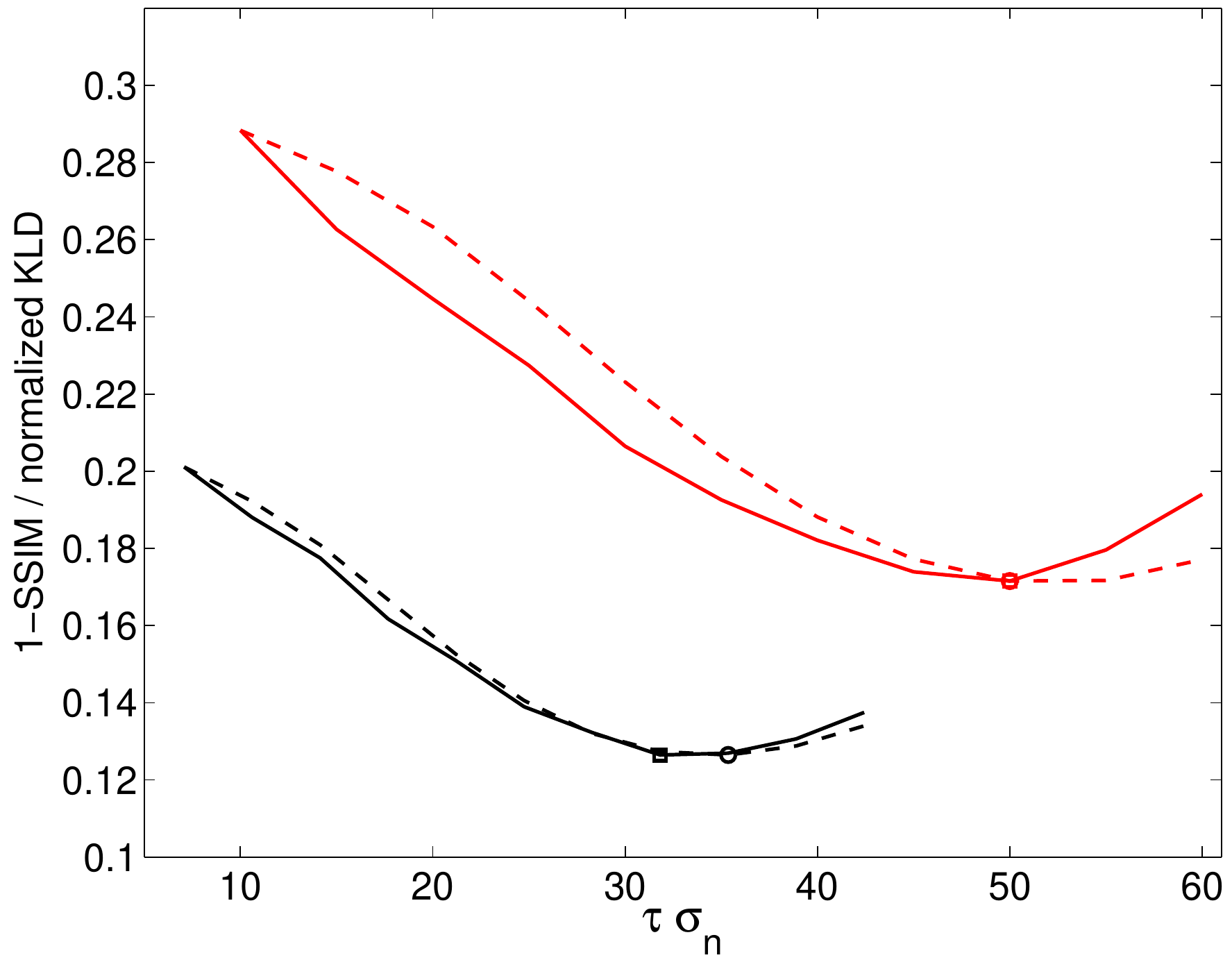}} \\
     JPEG & JPEG2000\\
     \includegraphics[width=0.48\textwidth]{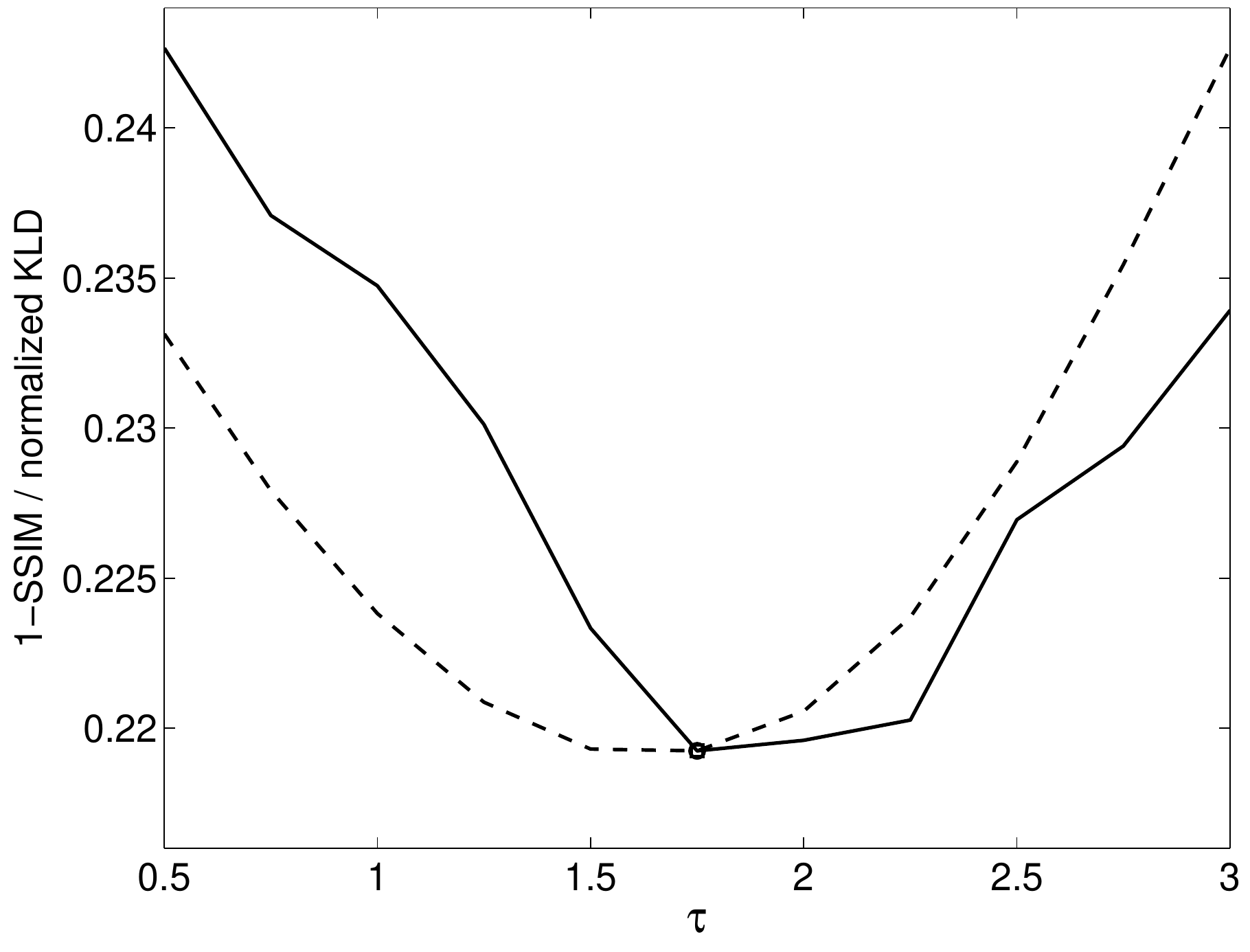} \hspace{0.02cm} & \hspace{0.02cm}
     \includegraphics[width=0.48\textwidth]{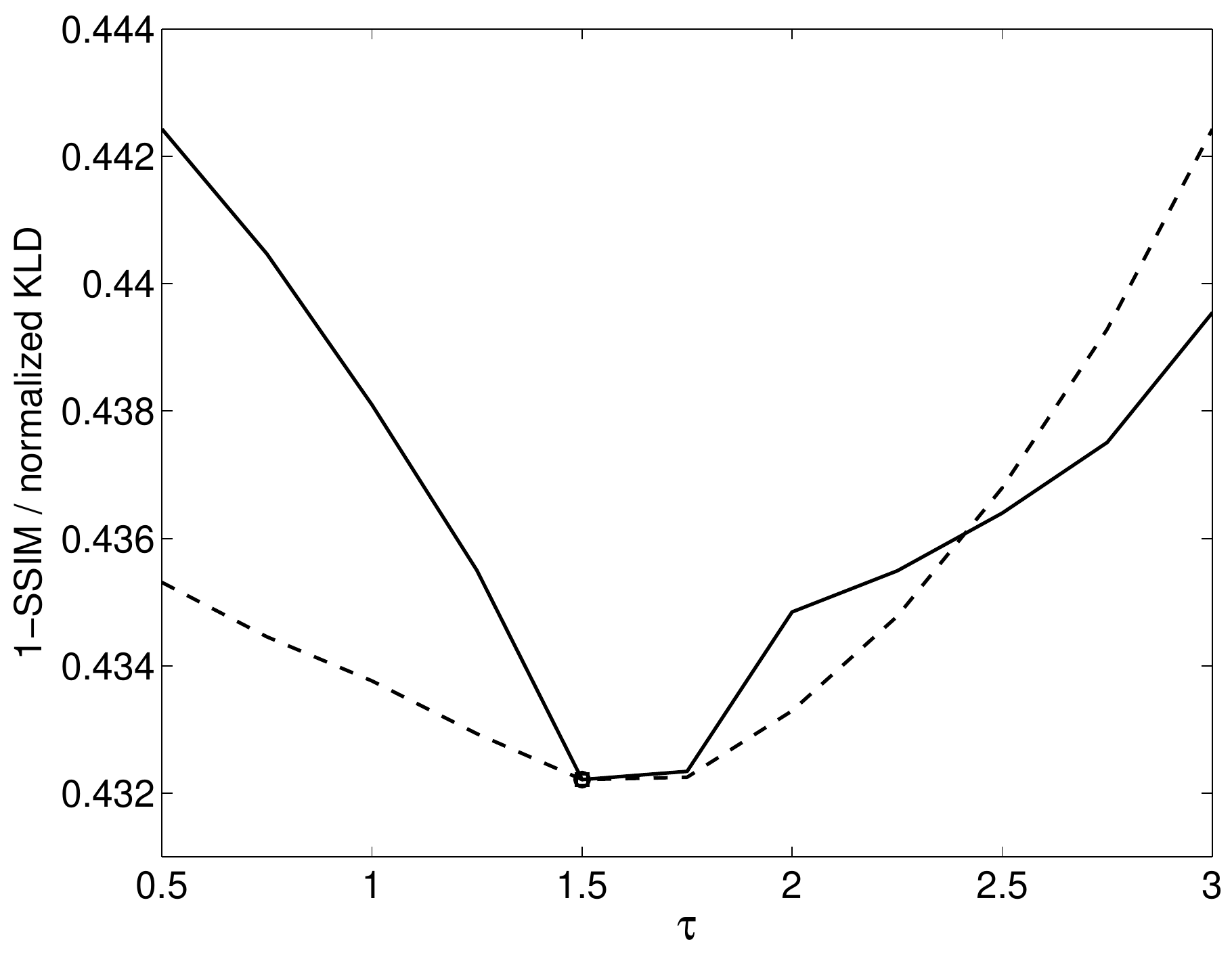} \\
     Vertical Striping & IRIS noise\\
     \includegraphics[width=0.48\textwidth]{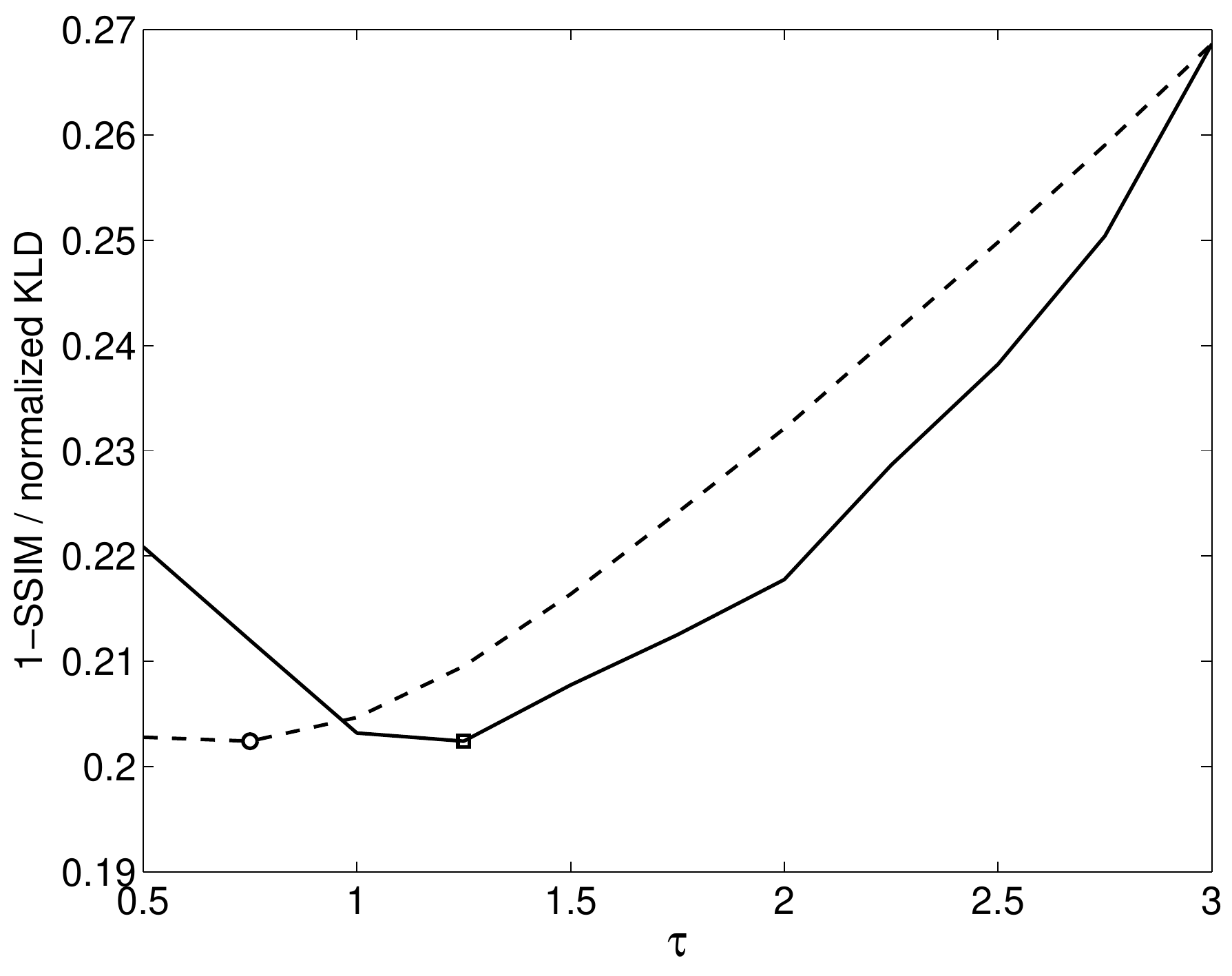} \hspace{0.02cm} & \hspace{0.02cm}
     \includegraphics[width=0.48\textwidth]{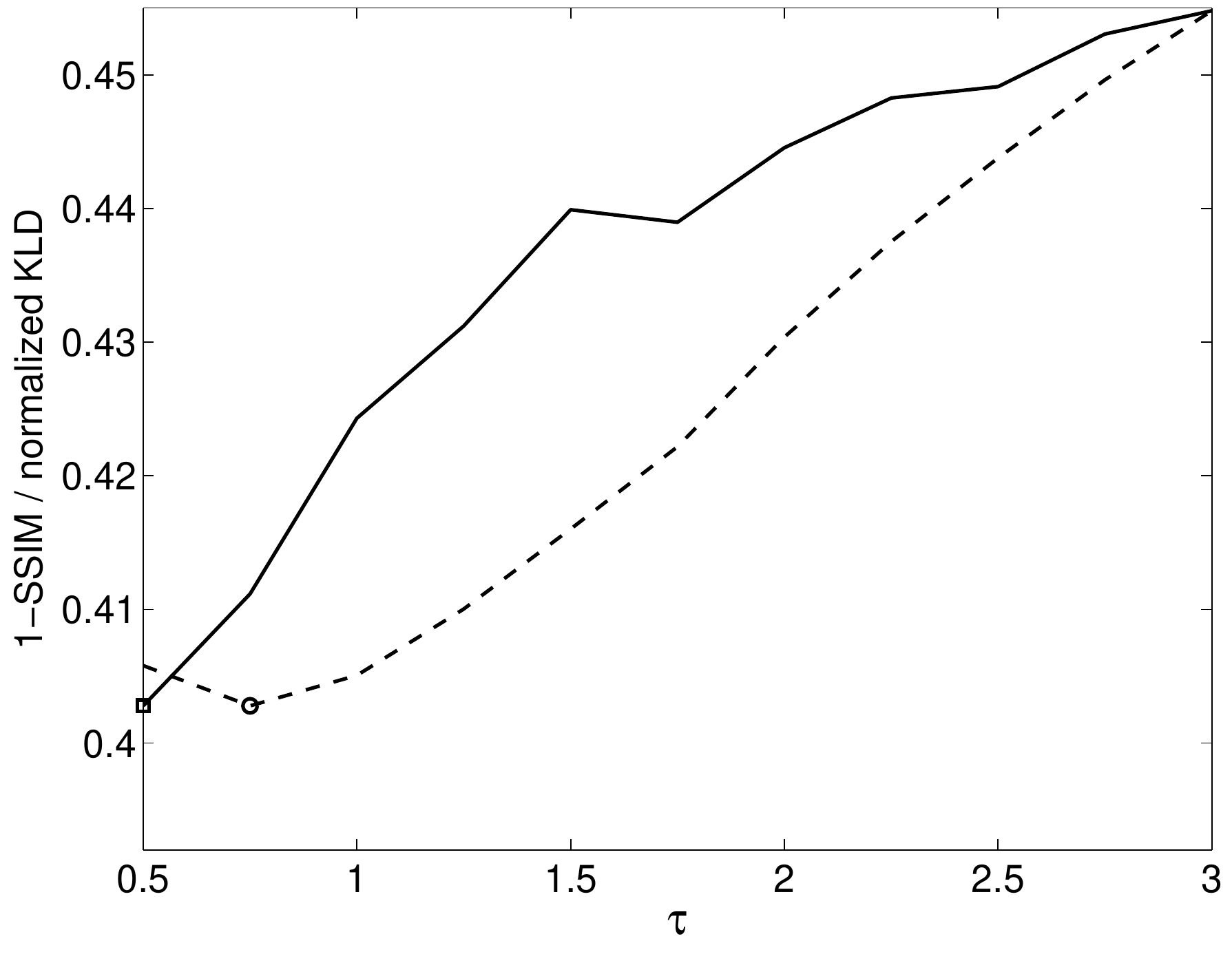}
     \\
     \end{tabular}
\end{center}
       \caption{\blue{Validation of the proposed KLD criterion to adjust SVR parameter $\varepsilon$ (or equivalently $\tau$, see text).
         In every distortion case, solid lines represent the KLD criterion and dashed lines represent
         the distortion (1-SSIM). For proper visualization,
         KLD curves were normalized to fall in the same range as the distortion. In the Gaussian noise case,
         two different noise variances are considered: $\sigma^2_n=200$ (black lines) and  $\sigma^2_n=400$ (red lines).
         As can be seen, the minima of the KL distance (squares) are always in the same region as the minima
         of the distortion (circles), thus giving rise to similar SSIM performance.}}
         \label{superficieKLD}
\end{figure}

In this experiment, the SVM parameter space is reduced to the scaling factor on the insensitivity profile
as recommended in Section \ref{SVMparams}.
Accordingly, Fig.~\ref{superficieKLD} shows the KLD and distortion (1-SSIM) results as a function
of $\tau$ (see Eq. \eqref{tubito2}).
Curves are shown for different kinds of (Gaussian and non-Gaussian) noise sources corrupting a
particular image (details on the noise sources are given in section \ref{experimentos}).
\newline
For the Gaussian noise case, two different variances are shown.
It is worth noting that (1) the KLD criterion (solid) closely follows the actual distortion curve
(dashed), and (2) the minima for low and high noise regime curves are very similar. These facts suggest
that, in the Gaussian noise case, the proposed criterion is quite robust and provides consistent results:
the higher the noise (red curves) the higher the $\varepsilon$ zone minima. Besides, the linear
relation between $\varepsilon$ and the noise standard deviation, reported in \citep{Kwok03}, is confirmed here: as expected, the
scaling factor keeps fairly constant, $\tau\approx 2.5$, for both $\sigma^2_n=200$ and
$\sigma^2_n=400$. Obviously, higher noise levels imply more distorted estimations.
For other (non-Gaussian) noise sources, similar results are obtained.
For the JPEG and JPEG2000 quantization noise sources, the KLD criterion also matches SSIM performance.
For the case of more complex noise sources, such as vertical striping (VS) and Infra Red Imaging System (IRIS) noise,
the criterion gives close-to-optimal solutions in SSIM terms.
\green{Note that, remarkably, the KLD criterion is better suited to the error minimization when the signal and noise independence assumption holds (Gaussian case).
Therefore there is room to further improve the SVR selection criterion.
The above results suggest that the proposed SVR selection procedure can be considered
as a convenient approximation to distortion minimization (which is not possible in real situations).}

\section{Denoising experiments and discussion}
\label{experimentos}

In this section, we evaluate the performance of the proposed method in different
scenarios for image denoising. Our algorithm is compared to several wavelet-based
denoising methods using standard
256$\times$256 images (`Barbara', `Boats', `Lena') with different levels and sources
of degradation.
\blue{In the following, we first give details on implementation issues of the considered
algorithms. Then, we analyze their performance for several kinds of noise sources:
\begin{itemize}
\item Experiment 1. Additive Gaussian noise of different variances ($\sigma_n^2$ = $\{200,400\}$).
\item Experiment 2. Coding noise: JPEG and JPEG2000 at different quantization coarseness.
\item Experiment 3. Acquisition noise: vertical striping and Infra Red Imaging System (IRIS) noise.
\end{itemize}
Note that the noise PDF is in general unknown, except for the \green{academic} case of Gaussian noise,
but the histogram can be computed from samples in all cases.
\newline
All results are compared numerically by using the standard (yet not perceptually meaningful) RMSE, and the perceptually meaningful SSIM index~\citep{Wang04}. Moreover, representative examples are shown in every case for visual inspection.
For proper visualization, all the results are equalized in the same way by truncating the values outside the $[0,255]$ range.
}

\subsection{\blue{Implementation details}}\label{implementation}
The algorithms that do not use information about the inter-coefficient relations \citep{Donoho95,simoncelli99b,Figueiredo01}
are straightforward to implement and have few parameters to tune. All these methods use orthogonal
wavelet representations. In our particular implementation, we used $4$-scale QMF wavelets from
MatlabPyrTools\footnote{\url{http://www.cns.nyu.edu/~eero/software.php}}. In every case, we followed authors' prescriptions to choose these parameters for the best performance:
\begin{itemize}
\item {\em Hard Thresholding (HT)}. The key parameter is the threshold
             value $\lambda$. We use the noise variance to set the threshold,
             $\lambda = 3\sigma_n$, as suggested in \citep{Donoho95}.
\item {\em Soft Thresholding (ST)}. In our implementation, the threshold in
             each subband is derived from the standard deviation of the noise, $\sigma_n$,  using optimized values
             to minimize the mean square error (MSE) in a set of $100$ natural images.
             Threshold values were optimized for the $\sigma_n^2$ in the range [$0$,$1600$].
\item {\em Bayesian Laplacian (BL)}.
             In this case, the parameters of the Laplacian distribution ($s$ and $p$ in \citep{simoncelli99b})
             for the marginal PDFs in each subband are estimated by maximum likelihood (ML),
             as suggested by the author.
\item {\em Bayesian Gaussian (BG)}. The threshold value was set according to the function of
             noise variance provided in \citep{Figueiredo01}.
\end{itemize}
On the other hand, in the case of the Gaussian Scale Mixture (GSM) \citep{Portilla03}, which does consider inter-coefficient relations,
we used the implementation provided by the authors\footnote{\url{http://decsai.ugr.es/~javier/denoise/}}.
We have used (1) the same representation as in the proposed
method ($4$-scale, $8$-orientation steerable pyramid), and (2) we also provided the
average noise power spectral density (PSD) to achieve the best possible performance of the GSM method.

Details of the proposed SVR method are included in previous Section \ref{SVMparams}. A Matlab implementation
of the algorithm is available online\footnote{\url{http://www.uv.es/vista/vistavalencia/denoising_SVR/}}.
Since the $C_i$ and $\varepsilon_i$ profiles are computed off-line, the computational cost of the
proposed method is mainly constrained by the SVR training. \green{In our current implementation, we used
the IRWLS algorithm in Matlab \citep{PerezCruz00} in order to drop the bias term and incorporate the insensitivity and penalization profiles easily. These modifications are not trivial in faster implementations \citep{Huang04,Kecman04}.} As a result, our Matlab implementation
takes about $30$ seconds\footnote{Computations were carried out in a 2.8GHz processor with 4GB RAM.} for each image/noise estimation for a set of SVR parameters.
Three strategies can be carried out for speeding up the optimization: (1) using faster
SVR implementations \citep{Platt99,Chang01,Tsang05}, (2) alternative procedures to exhaustive search on convex error surfaces \citep{Torczon97,Lewis02,Vishwanathan06}, and (3) restricting the dimension of the parameter space (as done in Section \ref{SVMparams}).

\subsection{Experiment 1. Additive Gaussian noise}

Table \ref{tablagaussian} shows the distortion results for the \green{three considered images and the two noise variances, $\sigma_n^2=200$ and $\sigma_n^2=400$.}.
\blue{The best SSIM values in each case are highlighted.
In every case, we also provide the SVR$^{opt}$ result, which is the best result the proposed method
can get in SSIM terms. This is useful to assess the performance of the proposed divergence-based
criterion and to give an upper bound of method's performance}.
Results show that our algorithm performs better than the methods that neglect signal
relations (HT, ST, BG and BL), and obtains similar (yet slightly lower)
numerical results than the one which incorporates them (GSM).
It is not surprising that the GSM method achieves the best performance in this case, since
it is analytically suited to deal with Gaussian noise.
The SVR performance is consistent through all images and noise variances, thus suggesting that the
guiding criterion is robust. Finally, it must be noted that,
in the most difficult case of $\sigma_n^2=400$, GSM and SVR offer more
similar results, and clearly outperform the rest of the methods.

\begin{table}[t!]
\caption{\green{Results for the Gaussian noise: distortions for different images and methods are given at $\sigma_n^2 = 200$ (top) and $\sigma_n^2 = 400$ (bottom).}}
\footnotesize
\begin{center}
\begin{tabular}{l||cc||cc||cc}
\hline
\hline
          & \multicolumn{2}{c||}{`Barbara'}  &  \multicolumn{2}{c||}{`Boats'}    &  \multicolumn{2}{c}{`Lena'} \\
   Method             &  SSIM         &       RMSE     &  SSIM         &       RMSE     &  SSIM         &       RMSE     \\
\hline
\hline
HT              & 0.77   &   16.48  & 0.76  &   13.62  & 0.73  &   18.97  \\
ST              & 0.78   &   14.37  & 0.79  &   10.26  & 0.74  &   12.59  \\
BG              & 0.80    &   14.14  & 0.79  &   11.70   & 0.76  &   12.75  \\
BL              & 0.81   &   12.95  & 0.83  &   8.30    & 0.78  &   11.66  \\
GSM             & {\bf 0.90}  &   8.94   & {\bf 0.87}  &   8.94   & {\bf 0.83}  &   13.61  \\
\hline
SVR             & 0.87   &   10.11  & 0.84  &   10.16  & 0.81  &   12.54  \\
SVR$^{opt}$     & 0.87   &   10.11  & 0.85  &   10.36  & 0.82  &   12.30   \\
\hline
\hline
          & \multicolumn{2}{c||}{`Barbara'}  &  \multicolumn{2}{c||}{`Boats'}    &  \multicolumn{2}{c}{`Lena'} \\
   Method             &  SSIM         &       RMSE     &  SSIM         &       RMSE     &  SSIM         &       RMSE     \\
\hline
\hline
HT		&   0.67  &   24.52  &   0.68  &   20.15  &   0.67  &   20.22 \\
ST  		&   0.69  &   19.04  &   0.71  &   16.16  &   0.66  &   19.72 \\
BG		&   0.70   &   20.40   &   0.70   &   19.17  &   0.67  &   19.26 \\
BL 		&   0.73  &   16.52  &   0.77  &   10.26  &   0.67  &   18.45 \\
GSM             &   {\bf 0.86}  &   11.02  &   0.80   &   17.40   &   {\bf 0.79}  &   15.95 \\
\hline
SVR  		&   0.83  &   13.13  &   {\bf 0.81}  &   10.73  &   0.78  &   14.50  \\
SVR$^{opt}$ 	&   0.83  &   13.13  &   0.81  &   10.73  &   0.78  &   14.06 \\
\hline
\hline
\end{tabular}
\end{center}
\label{tablagaussian}
\end{table}

\begin{figure}[p!]
\begin{center}
\vspace{-1cm}
    \begin{tabular}{c@{}c}
      Noisy Image \hspace{0.2cm} (0.46) & HT \hspace{0.2cm} (0.67)\\
     \includegraphics[width=4.9cm,height=4.9cm]{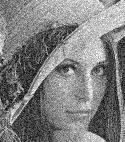}
      \hspace{0.02cm} & \hspace{0.02cm}
      \includegraphics[width=4.9cm,height=4.9cm]{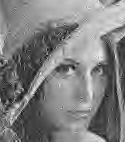} \\
      ST \hspace{0.2cm}(0.66) & BG \hspace{0.2cm}(0.67)\\
      \includegraphics[width=4.9cm,height=4.9cm]{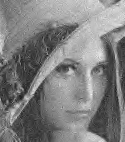} \hspace{0.02cm} & \hspace{0.02cm}
      \includegraphics[width=4.9cm,height=4.9cm]{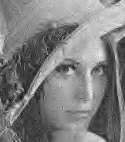} \\
     BL \hspace{0.2cm}(0.67) & GSM \hspace{0.2cm}(\textbf{0.79})\\
      \includegraphics[width=4.9cm,height=4.9cm]{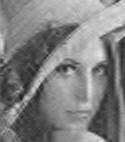} \hspace{0.02cm} & \hspace{0.02cm}
      \includegraphics[width=4.9cm,height=4.9cm]{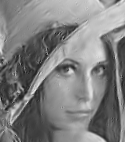}
      \\
     SVR \hspace{0.2cm}(0.78) & SVR$^{opt} \hspace{0.2cm}(0.78)$\\
      \includegraphics[width=4.9cm,height=4.9cm]{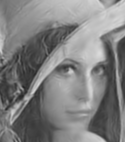}
         \hspace{0.02cm} & \hspace{0.02cm}
      \includegraphics[width=4.9cm,height=4.9cm]{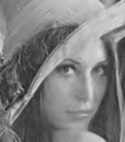} \\
    \end{tabular}
\end{center}
\vspace{-0.5cm}
\caption{Visual results for the `Lena' image corrupted with Gaussian noise, $\sigma_n^2=400$. SSIM values are given in parentheses.}\label{fig:res}
\end{figure}

Figure~\ref{fig:res} shows representative visual results in the challenging situation
of $\sigma_n^2=400$.
It can be noticed that thresholding methods (HT, ST)
and Bayesian generalizations not including signal relations in the model
(BG, BL) show poor performance, producing images either grained
or corrupted by too salient wavelet functions.
Even though SVR yields slightly lower numerical scores than GSM,
global visual performance is comparable.

\subsection{\blue{Experiment 2. Coding noise: JPEG and JPEG2000}}

\blue{In this section, we focus on restoring grayscale images
after JPEG or JPEG2000 compression, which induces non-Gaussian noise: it produces heavy tailed
marginal error PDFs in the spatial domain with non-negligible relations among the pixels
(see comments in subsection \ref{analisis_residuos}).
Quantization noise is an illustrative example of how the proposed method can cope with
non-Gaussian, colored and signal-dependent noise.
In order to obtain the necessary samples to build the noise histograms, we
used $100$ images from the database described in Section 2 encoded by JPEG and JPEG2000. In the first case, the Matlab implementation of
the JPEG algorithm with quality factors $Q=9$ (small distortion) and $Q=7$ (large distortion) was used.
In the second case, scalar quantization of the QMF wavelet domain using standard JPEG2000 bit
allocation tables \citep{Taubman01} was used. Different values of quantization coarseness, that will be referred to
as $\Delta_1$ (small distortion) and $\Delta_2$ (large distortion) were applied.}

\begin{figure}[p!]
\begin{center}
\vspace{-1cm}
    \begin{tabular}{c@{}c}
      Noisy Image \hspace{0.2cm} (0.68) & HT \hspace{0.2cm} (0.65)\\
     \includegraphics[width=4.9cm,height=4.9cm]{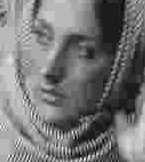}
      \hspace{0.02cm} & \hspace{0.02cm}
      \includegraphics[width=4.9cm,height=4.9cm]{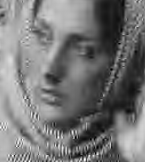} \\
      ST \hspace{0.2cm} (0.68) & BG \hspace{0.2cm} (0.66) \\
      \includegraphics[width=4.9cm,height=4.9cm]{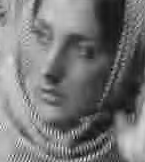} \hspace{0.02cm} & \hspace{0.02cm}
      \includegraphics[width=4.9cm,height=4.9cm]{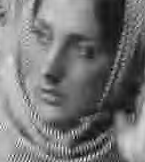} \\
     BL \hspace{0.2cm} (0.64) & GSM \hspace{0.2cm} (\textbf{0.71}) \\
      \includegraphics[width=4.9cm,height=4.9cm]{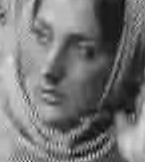} \hspace{0.02cm} & \hspace{0.02cm}
      \includegraphics[width=4.9cm,height=4.9cm]{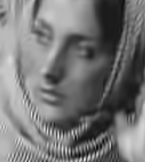}
      \\
     SVR \hspace{0.2cm} (\textbf{0.71}) & SVR$^{opt}$ \hspace{0.2cm} (0.73)\\
      \includegraphics[width=4.9cm,height=4.9cm]{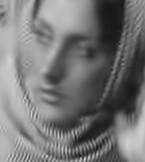}
         \hspace{0.02cm} & \hspace{0.02cm}
      \includegraphics[width=4.9cm,height=4.9cm]{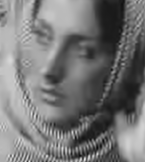} \\
    \end{tabular}
\end{center}
\vspace{-0.5cm}
\caption{Visual results for the `Barbara' image with JPEG quantization noise ($Q=7$). SSIM values are given in parentheses.}\label{fig:quantization}
\end{figure}

\begin{figure}[p!]
\begin{center}
\vspace{-1cm}
    \begin{tabular}{c@{}c}
      Noisy Image \hspace{0.2cm} (0.54) & HT \hspace{0.2cm} (0.54)\\
      \includegraphics[width=4.9cm,height=4.9cm]{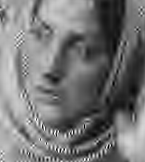}
      \hspace{0.02cm} & \hspace{0.02cm}
      \includegraphics[width=4.9cm,height=4.9cm]{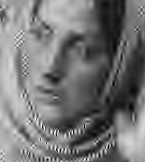} \\
      ST \hspace{0.2cm} (0.55) & BG \hspace{0.2cm} (0.54)\\
      \includegraphics[width=4.9cm,height=4.9cm]{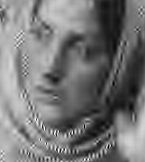} \hspace{0.02cm} & \hspace{0.02cm}
      \includegraphics[width=4.9cm,height=4.9cm]{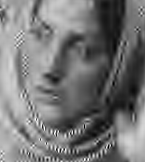} \\
     BL \hspace{0.2cm} (0.54) & GSM \hspace{0.2cm} (0.55)\\
      \includegraphics[width=4.9cm,height=4.9cm]{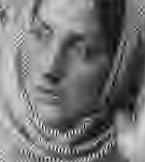} \hspace{0.02cm} & \hspace{0.02cm}
      \includegraphics[width=4.9cm,height=4.9cm]{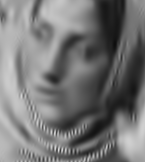}
      \\
     SVR \hspace{0.2cm} (\textbf{0.57}) & SVR$^{opt}$ \hspace{0.2cm} (0.57)\\
      \includegraphics[width=4.9cm,height=4.9cm]{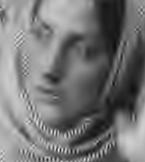}
         \hspace{0.02cm} & \hspace{0.02cm}
      \includegraphics[width=4.9cm,height=4.9cm]{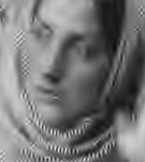} \\
    \end{tabular}
\end{center}
\vspace{-0.5cm}
\caption{Visual results for the `Barbara' image with coarse quantization JPEG2000 noise. SSIM values are given in parentheses.}\label{fig:quantization2000}
\end{figure}

\blue{Table \ref{tablaquantization} shows the quantitative results for all considered
methods for \blue{the three images at different quantization levels}.
It can be noticed that again the SVR method outperforms the thresholding methods (HT, ST)
and those not including signal relations in the model (BG, BL).
SVR yields similar numerical scores than GSM in JPEG (Fig. \ref{fig:quantization}).
However, in JPEG2000 better numerical (Table \ref{tablaquantization} [bottom]) and visual (Fig. \ref{fig:quantization2000}) results are obtained with SVR.
In general, high frequency details are better preserved by our method, while
GSM yields over smoothed solutions, particularly in JPEG2000.}

\begin{table}[t!]
\footnotesize
\begin{center}
\caption{\green{Results for the coding noise: distortions at different quality levels of
JPEG ($Q=\{9,7\}$) and JPEG2000 (coarseness $\Delta_1$ and $\Delta_2$) are given for different images and methods.}}
\begin{tabular}{l||p{0.70cm}p{0.70cm}|p{0.70cm}p{0.70cm}|p{0.70cm}p{0.70cm}||p{0.70cm}p{0.70cm}|p{0.70cm}p{0.70cm}|p{0.70cm}p{0.70cm}}
\hline
\hline
{\sf JPEG}	&  \multicolumn{6}{|c||}{$Q=9$} &  \multicolumn{6}{|c}{$Q=7$} \\
\hline
\hline
 		& \multicolumn{2}{c|}{`Barbara'} & \multicolumn{2}{c|}{`Boats'} & \multicolumn{2}{c||}{`Lena'}	 & \multicolumn{2}{c|}{`Barbara'} & \multicolumn{2}{c}{`Boats'} & \multicolumn{2}{|c}{`Lena'}	 \\
\hline
\hline
Method		& SSIM & RMSE & SSIM & RMSE & SSIM & RMSE 	 & SSIM & RMSE & SSIM & RMSE & SSIM & RMSE  	 \\
\hline
HT 		&  0.70  &   20.05 & 0.75  &   13.07 & 0.70  &   18.40 & 0.65  &   22.11 & 0.71  &   16.34 & 0.65  &   24.99 \\
ST 		&  0.73  &   17.51 & 0.78  &   11.59 & 0.73  &   15.13 & 0.68  &   19.71 & 0.75  &   12.72 & 0.68  &   18.77 \\
BG 		&  0.72  &   18.76 & 0.77  &   12.30 & 0.72  &   16.27 & 0.66  &   21.57 & 0.74  &   13.32 & 0.67  &   21.05 \\
BL 		&  0.71  &   20.37 & 0.77  &   13.43 & 0.73  &   16.52 & 0.64  &   21.67 & 0.74  &   14.70 & 0.69  &   17.65 \\
GSM		&  0.77  &   15.50 & {\bf 0.80}  &   11.15 & {\bf 0.75}  &   13.66 & {\bf 0.71}  &   18.56   &    {\bf 0.77}  &   12.18 & {\bf 0.71}  &   17.45    \\
\hline
SVR		&  {\bf 0.78}  &   14.89  & 0.78  &   12.13 & 0.74  &   13.22  & {\bf 0.71}  &   18.42 & 0.76  &   12.84 & {\bf 0.71}  &   15.68  \\
SVR$^{opt}$	&   0.78       &   14.89        & 0.80  &   11.35 & 0.75  &   13.97        &   0.73      &   18.28       & 0.76  &   12.89 &   0.71  &   15.72 \\
\hline
\hline
{\sf JPEG2000}	&  \multicolumn{6}{|c||}{$\Delta_2$} &  \multicolumn{6}{|c}{$\Delta_1$} \\
\hline
\hline	
 		& \multicolumn{2}{c|}{`Barbara'} & \multicolumn{2}{c|}{`Boats'} & \multicolumn{2}{c||}{`Lena'}	 & \multicolumn{2}{c|}{`Barbara'} & \multicolumn{2}{c}{`Boats'} & \multicolumn{2}{|c}{`Lena'}	 \\\hline
\hline
Method	& SSIM & RMSE & SSIM & RMSE & SSIM & RMSE 	& SSIM & RMSE & SSIM & RMSE & SSIM & RMSE \\
\hline
HT	& 0.54  &   30.81 & 0.55  &   26.23 & 0.51  &   32.66 & 0.67  &   24.82 & 0.59  &   25.18 & 0.56  &   28.25 \\
ST	& 0.55  &   28.83 & 0.55  &   25.15 & 0.51  &   31.24 & 0.68  &   22.52  & 0.60  &   23.69 & 0.56  &   27.47 \\
BG	& 0.54  &   30.37 & 0.55  &   26.08 & 0.51  &   32.45 & 0.67  &   24.16 & 0.59  &   24.92 & 0.56  &   28.10 \\
BL	& 0.54  &   30.30 & 0.55  &   25.87 & 0.51  &   29.05 & 0.67  &   24.35 & 0.59  &   24.79 & 0.56  &   28.12 \\
GSM	&   0.55        &   28.47     &  {\bf 0.57}  &   20.92 & {\bf 0.52}  &  25.84 & 0.68        &   20.54   &    {\bf 0.64}  &   17.94 & 0.58        &   23.64   \\
\hline
SVR     & {\bf 0.57}  &   25.31  & {\bf 0.57}  &   21.88       & {\bf 0.52}  &   29.32        & {\bf 0.71}  &   17.23  & {\bf 0.64}  &   18.27        & {\bf 0.59}  &   21.55  \\
SVR$^{opt}$ &   0.57  &   25.31 &   0.57  &   21.74 &   0.52  &   25.35 &   0.72  &   17.04 &   0.64  &   18.27 &   0.59  &   21.55 \\
\hline
\hline
\end{tabular}
\label{tablaquantization}
\end{center}
\end{table}

\subsection{\blue{Experiment 3. Acquisition noise: vertical striping and IRIS noise}}

\begin{figure}[p!]
\begin{center}
\vspace{-1cm}
    \begin{tabular}{c@{}c}
      Noisy Image \hspace{0.2cm} (0.77) & HT \hspace{0.2cm} (0.69)\\
     \includegraphics[width=4.9cm,height=4.9cm]{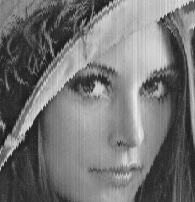}
      \hspace{0.02cm} & \hspace{0.02cm}
      \includegraphics[width=4.9cm,height=4.9cm]{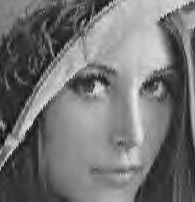} \\
      ST \hspace{0.2cm} (0.75) & BG \hspace{0.2cm} (0.73)\\
      \includegraphics[width=4.9cm,height=4.9cm]{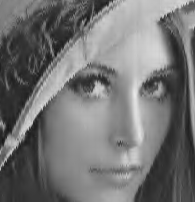} \hspace{0.02cm} & \hspace{0.02cm}
      \includegraphics[width=4.9cm,height=4.9cm]{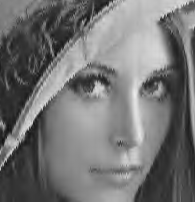} \\
     BL \hspace{0.2cm} (0.77) & GSM \hspace{0.2cm} (0.75)\\
      \includegraphics[width=4.9cm,height=4.9cm]{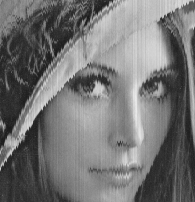} \hspace{0.02cm} & \hspace{0.02cm}
      \includegraphics[width=4.9cm,height=4.9cm]{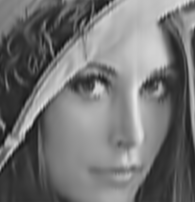}
      \\
     SVR \hspace{0.2cm} (\textbf{0.79}) & SVR$^{opt}$ \hspace{0.2cm} (0.80)\\
      \includegraphics[width=4.9cm,height=4.9cm]{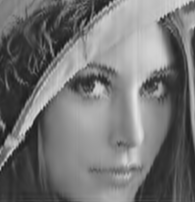}
         \hspace{0.02cm} & \hspace{0.02cm}
      \includegraphics[width=4.9cm,height=4.9cm]{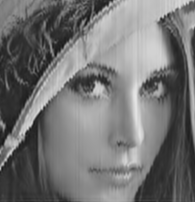} \\
    \end{tabular}
\end{center}
\vspace{-0.5cm}
\caption{Visual results for the `Lena' image with vertical striping noise. SSIM values are given in parentheses.}\label{fig:vs}
\end{figure}

\begin{figure}[p!]
\begin{center}
\vspace{-1cm}
    \begin{tabular}{c@{}c}
      Noisy Image \hspace{0.2cm} (0.59) & HT \hspace{0.2cm} (0.58)\\
      \includegraphics[width=4.9cm,height=4.9cm]{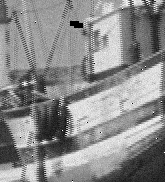}
      \hspace{0.02cm} & \hspace{0.02cm}
      \includegraphics[width=4.9cm,height=4.9cm]{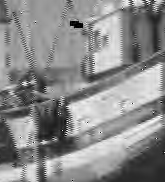} \\
      ST \hspace{0.2cm} (0.64) & BG \hspace{0.2cm} (0.62)\\
      \includegraphics[width=4.9cm,height=4.9cm]{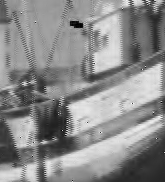} \hspace{0.02cm} & \hspace{0.02cm}
      \includegraphics[width=4.9cm,height=4.9cm]{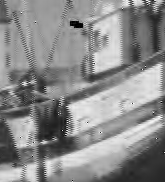} \\
     BL \hspace{0.2cm} (0.60) & GSM \hspace{0.2cm} (0.64)\\
      \includegraphics[width=4.9cm,height=4.9cm]{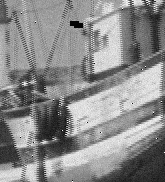} \hspace{0.02cm} & \hspace{0.02cm}
      \includegraphics[width=4.9cm,height=4.9cm]{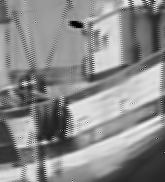}
      \\
     SVR \hspace{0.2cm} (\textbf{0.67}) & SVR$^{opt}$ \hspace{0.2cm} (0.70)\\
      \includegraphics[width=4.9cm,height=4.9cm]{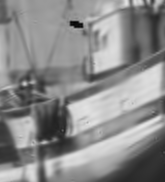}
         \hspace{0.02cm} & \hspace{0.02cm}
      \includegraphics[width=4.9cm,height=4.9cm]{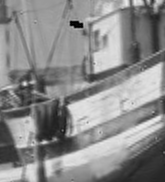} \\
    \end{tabular}
\end{center}
\vspace{-0.5cm}
\caption{Visual results for the `Boats' image with IRIS noise. SSIM values are given in parentheses.}\label{fig:tecnobit}
\end{figure}

\blue{Real imaging systems introduce complex forms of noise depending
on the acquisition process, so
assuming a particular PDF for all cases is far from being realistic. For instance, variation
of the intensity between neighboring elements of
the CCD typically leads to vertical striping noise
in pushbroom sensors \citep{Mouroulis00,Barducci01}. Other
typical acquisition noise source is observed in infrared imaging cameras,
which is a complex mixture of different noise sources. In this section,
we pay attention to these two particular non-Gaussian realistic
 acquisition noises through controlled experiments:
\begin{enumerate}
\item {\em Vertical striping noise.} We simulated this noise \green{by modifying 4\% of the image columns selected randomly. The luminance of the selected columns was modified by a random factor following a uniform distribution between 0.8 and 1.} Spatial coherence was forced by attaching groups of contiguous $5$ to $10$ strips.
\item {\em InfraRed Imaging System (IRIS) noise.} Inspired in the observed characteristics
    of a representative number of acquired images by a commercial IR camera, the noise was modeled by
a combination of four noise sources: low-variance Gaussian noise ($\sigma_n^2 \approx 50$),
`salt-and-pepper' noise (with a percentage of corrupted pixels about $0.05\%$),
some spatially coherent missing pixels (black patches), and interlaced lines all over the image.
\end{enumerate}
In both cases, we computed the contrast noise PDF, $p({\bf n})$, from $100$ noisy images. In the next
subsection \ref{analisis_residuos}, the non-Gaussian nature of these acquisition noise PDFs is shown.
\newline
Table \ref{acquisition} shows the obtained numerical results for all images and both acquisition noise sources. In
both complex scenarios, the proposed SVR-based method outperforms GSM and the rest of methods numerically. A
noticeable gain in SSIM is observed, which is confirmed when looking at the restored images in
Figs. \ref{fig:vs} and \ref{fig:tecnobit}. It is worth noting that
in the vertical striping noise (Fig. \ref{fig:vs}), SVR yields a sharper (and more realistic) reconstruction while
GSM produces an over-blurred solution. In the case of the IRIS noise, only SVR removes the
interlacing noise contribution, producing better visual results. Including the average PSD information in GSM, as we do in the experiments,
improves its performance. However, it is not enough to remove the interlacing artifact due to the particular nature of IRIS noise.
IRIS noise is difficult because the PSD and variance of each particular realization
of the noise may substantially differ from the (estimated) averages. On the contrary, the proposed SVR method uses an adaptive cost function
learned from the noisy image. Here, nevertheless, the upper
bound of performance is not met, suggesting that there is still room for improving the selection criterion proposed, \green{possibly considering the joint density}.
}

\begin{table}[t!]
\caption{\green{Acquisition noise: vertical striping (top) and IRIS noise (bottom). Distortions for different images and methods.} }
\footnotesize
\begin{center}
\begin{tabular}{l||cc||cc||cc}
\hline
\hline
Method          & \multicolumn{2}{c||}{`Barbara'}  &  \multicolumn{2}{c||}{`Boats'}    &  \multicolumn{2}{c}{`Lena'} \\
                &  SSIM         &       RMSE     &  SSIM         &       RMSE     &  SSIM         &       RMSE     \\
\hline
\hline
HT              &  0.73  &   17.43  &   0.73  &   15.99  &  0.69  &   18.07  \\
ST              &  0.77  &   15.71  &   0.78  &   14.04  &  0.75  &   14.08  \\
BG              &  0.76  &   16.01  &   0.76  &   14.75  &  0.73  &   15.14  \\
BL              &  0.77  &   16.56  &   0.81  &   14.96  &  0.77  &   14.64  \\
GSM             &  0.79  &   14.83  &   0.79  &   14.36  &  0.75  &   14.45  \\
\hline
SVR             &  {\bf 0.80}   &   15.66 &   {\bf 0.80}  &   13.47  &  {\bf 0.79}  &   13.18 \\
SVR$^{opt}$     &  0.80   &   15.45 &   0.82  &   14.25  &  0.80  &   13.31 \\
\hline
\hline
HT              &  0.50  &   30.80  &  0.58  &   28.70 & 0.56  &   28.81   \\
ST              &  0.55 &   27.02  &  0.64  &   23.48 & 0.60  &   24.40    \\
BG              &  0.54 &   28.40  &  0.62  &   25.44 & 0.59  &   26.20    \\
BL              &  0.50  &   28.74  &  0.60  &   21.77 & 0.55  &   24.08   \\
GSM             &  0.53 &   30.51  &  0.64  &   25.92 & 0.61  &   30.99    \\
\hline
SVR             &  {\bf 0.59} &   31.07 &  {\bf 0.67}  &   21.44 & {\bf 0.66}  &   31.44 \\
SVR$^{opt}$     &  0.60  &   30.71 &  0.70   &   24.56 & 0.66  &   32.05 \\
\hline
\hline
\end{tabular}
\end{center}
\label{acquisition}
\end{table}

\subsection{\blue{Analysis of the residuals}}
\label{analisis_residuos}

\blue{Further qualitative insight in the obtained solutions can be achieved by
comparing the estimated and actual PDFs of signal and noise with the different methods
and noise sources. Since we are restricting ourselves to second order KLD criterion,
this comparison reduces to assess the difference between 2D histograms (in the spatial domain).}

\blue{It is widely known that the PDF of pairs of neighbor pixels in natural images is an oriented
ellipsoid reflecting the strong correlation among luminance values in the spatial domain \citep{Clarke85}. The corresponding restored images (even for the worse performing algorithms)
also display such strong local correlation. Therefore, no relevant conclusion is gained by
direct inspection of these histograms (results not shown).
On the contrary, the 2D histograms of the noise are more suitable for direct inspection because
(1) actual noise histograms are quite different for the different noise sources, and (2)
the estimated histograms strongly depend on the denoising method.}

\begin{landscape}
\begin{figure}[t]
\begin{center}
\setlength{\tabcolsep}{0.8pt}
    \begin{tabular}{p{1cm}p{2.6cm}p{2.6cm}p{2.6cm}p{2.6cm}p{2.6cm}p{2.6cm}p{2.6cm}}
      & \hspace{-0.25cm}Spatial Noise & \hspace{1cm}HT & \hspace{1cm}ST & \hspace{1cm}BG & \hspace{1cm}BL & \hspace{1cm}GSM & \hspace{1cm}SVR \\ 
      \hspace{-0.25cm}\begin{sideways}{$\sigma^2=400$}\end{sideways} &
      \hspace{-0.8cm}\includegraphics[width=2.75cm]{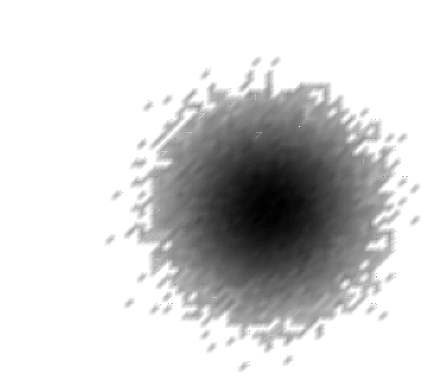} &
      \hspace{-0.5cm}\includegraphics[width=2.75cm]{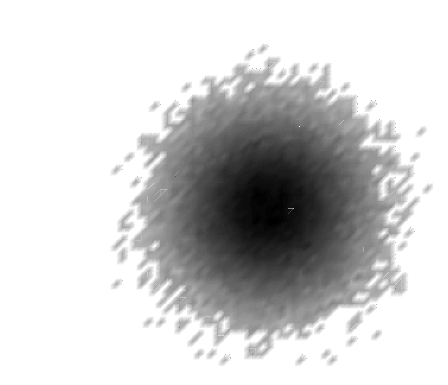} &
      \hspace{-0.5cm}\includegraphics[width=2.75cm]{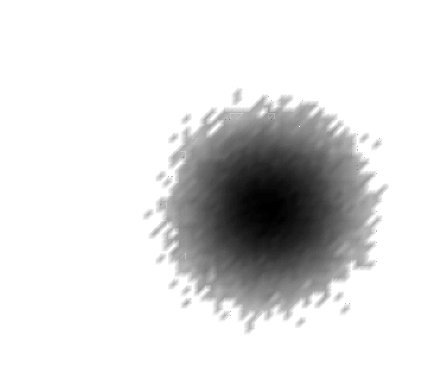} &
      \hspace{-0.5cm}\includegraphics[width=2.75cm]{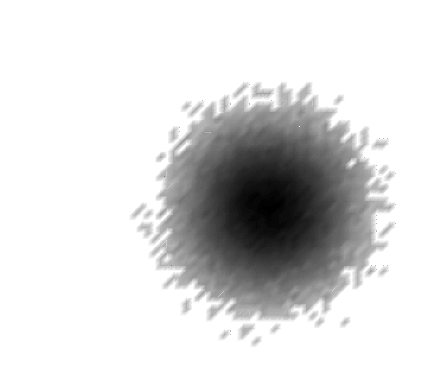} &
      \hspace{-0.5cm}\includegraphics[width=2.75cm]{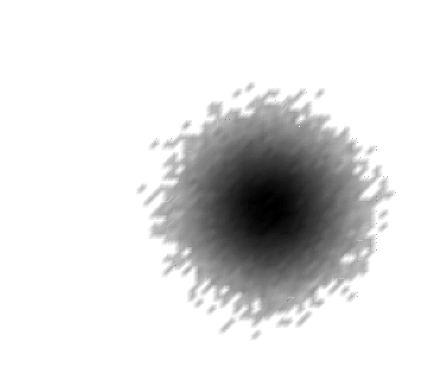} &
      \hspace{-0.5cm}\includegraphics[width=2.75cm]{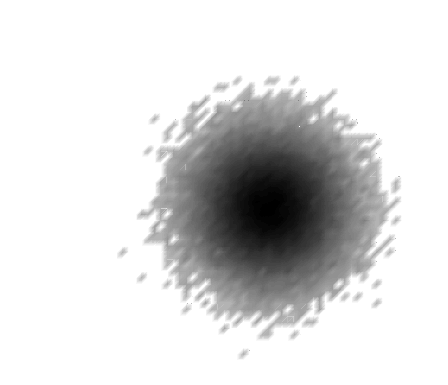} &
      \hspace{-0.5cm}\includegraphics[width=2.75cm]{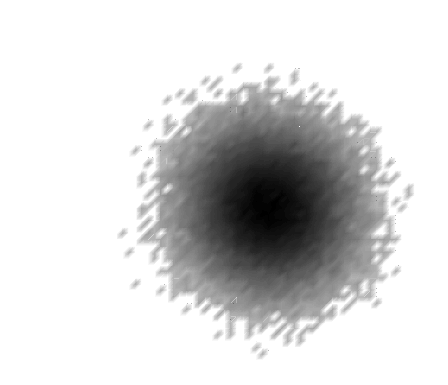} \\ 
      \hspace{-0.25cm}\begin{sideways}\hspace{0.6cm}{JPEG}\end{sideways}  &
      \hspace{-0.8cm}\includegraphics[width=2.75cm]{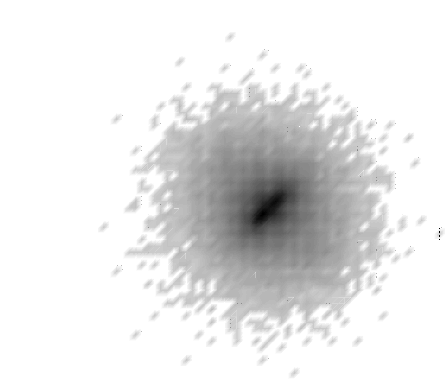} &
      \hspace{-0.5cm}\includegraphics[width=2.75cm]{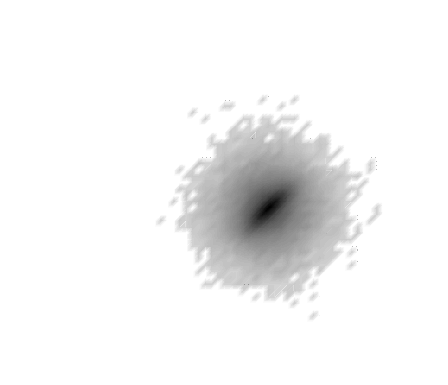} &
      \hspace{-0.5cm}\includegraphics[width=2.75cm]{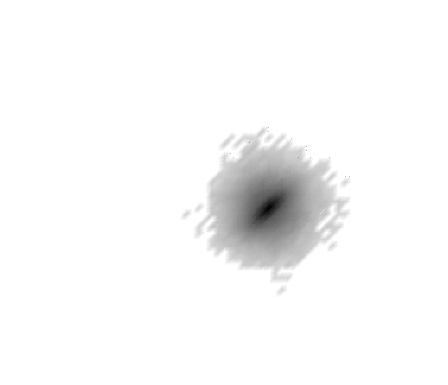} &
      \hspace{-0.5cm}\includegraphics[width=2.75cm]{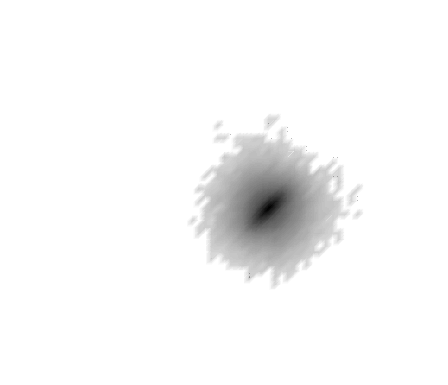} &
      \hspace{-0.5cm}\includegraphics[width=2.75cm]{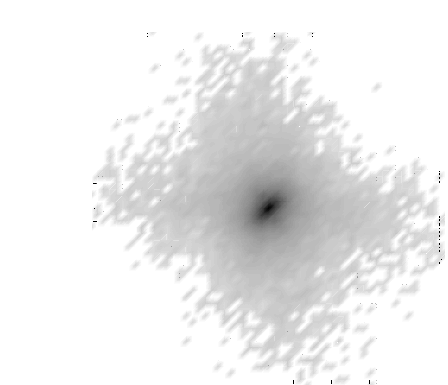} &
      \hspace{-0.5cm}\includegraphics[width=2.75cm]{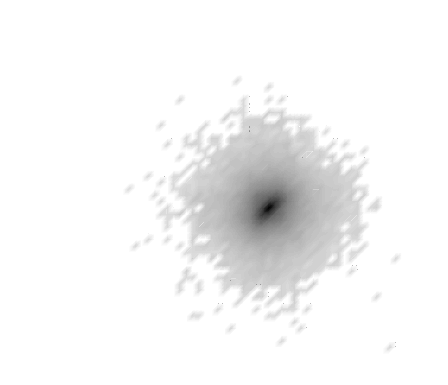} &
      \hspace{-0.5cm}\includegraphics[width=2.75cm]{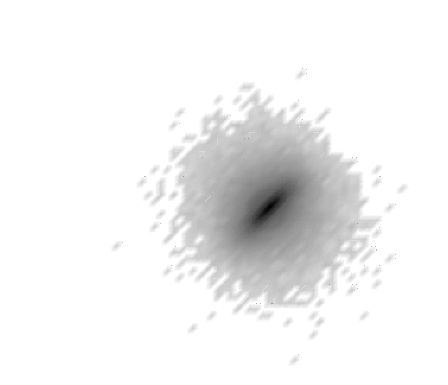} \\ 
      \hspace{-0.25cm}\begin{sideways}{JPEG2000}\end{sideways}  &
      \hspace{-0.8cm}\includegraphics[width=2.75cm]{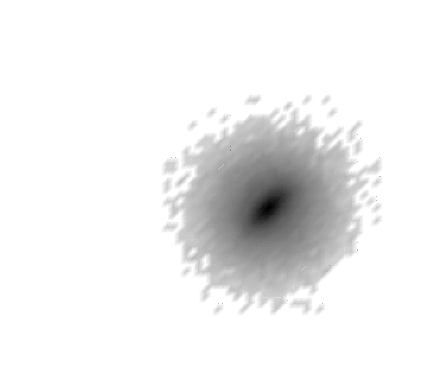} &
      \hspace{-0.5cm}\includegraphics[width=2.75cm]{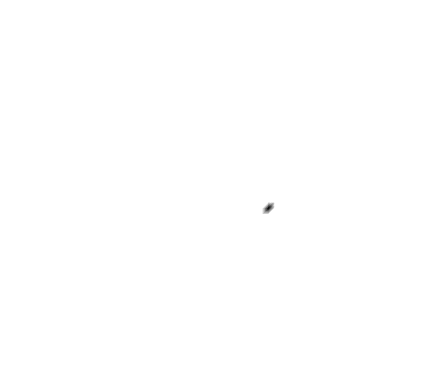} &
      \hspace{-0.5cm}\includegraphics[width=2.75cm]{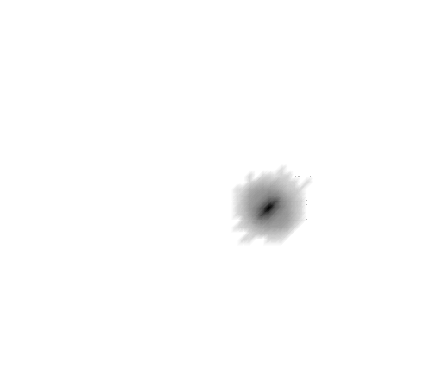} &
      \hspace{-0.5cm}\includegraphics[width=2.75cm]{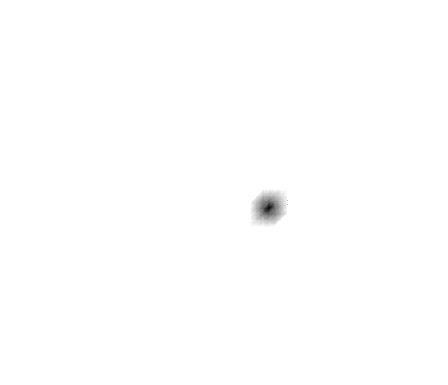} &
      \hspace{-0.5cm}\includegraphics[width=2.75cm]{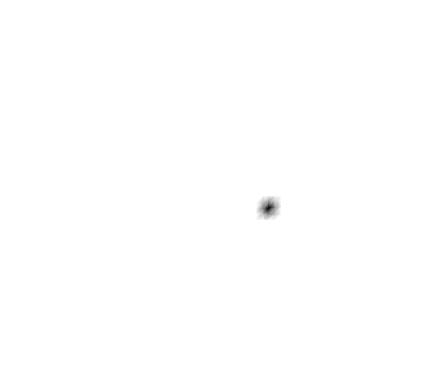} &
      \hspace{-0.5cm}\includegraphics[width=2.75cm]{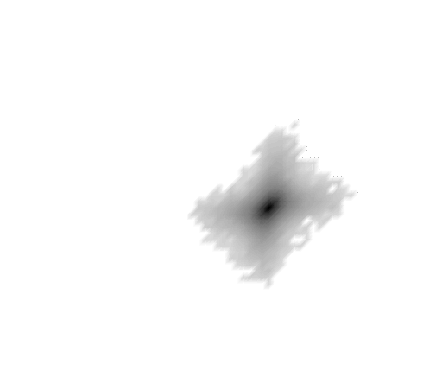} &
      \hspace{-0.5cm}\includegraphics[width=2.75cm]{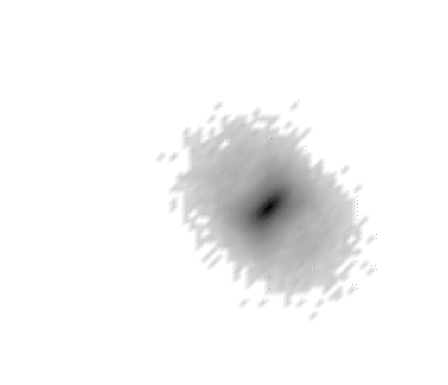}\\ 
      \hspace{-0.25cm}\begin{sideways}\hspace{0.7cm}{VS}\end{sideways}  &
      \hspace{-0.8cm}\includegraphics[width=2.75cm]{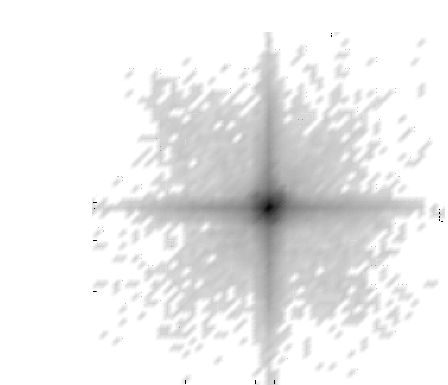} &
      \hspace{-0.5cm}\includegraphics[width=2.75cm]{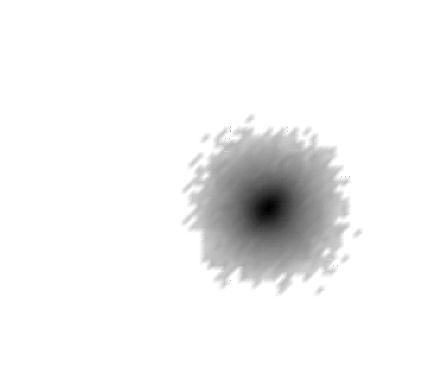} &
      \hspace{-0.5cm}\includegraphics[width=2.75cm]{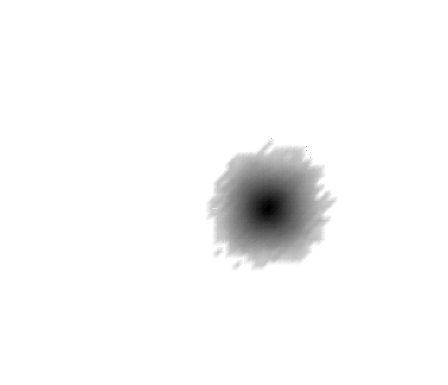} &
      \hspace{-0.5cm}\includegraphics[width=2.75cm]{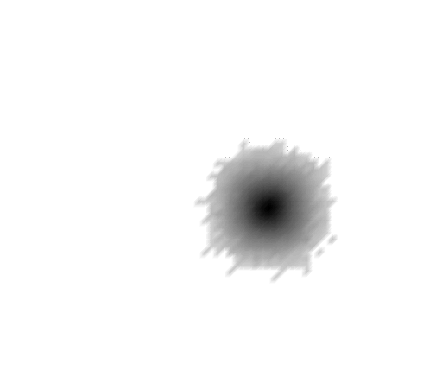} &
      \hspace{-0.5cm}\includegraphics[width=2.75cm]{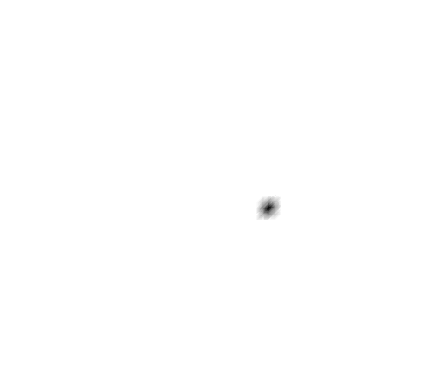} &
      \hspace{-0.5cm}\includegraphics[width=2.75cm]{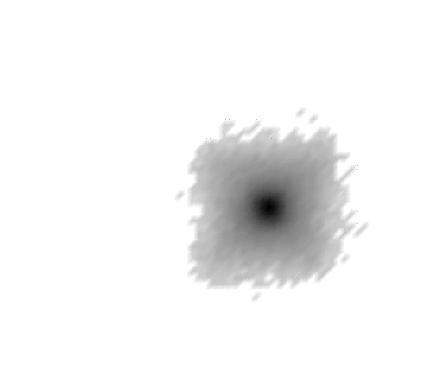} &
      \hspace{-0.5cm}\includegraphics[width=2.75cm]{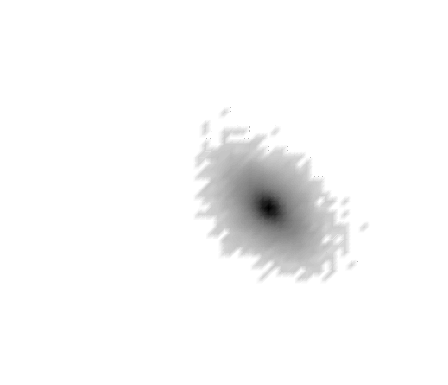} \\
      \hspace{-0.25cm}\begin{sideways}\hspace{0.6cm}{IRIS}\end{sideways}  &
      \hspace{-0.8cm}\includegraphics[width=2.75cm]{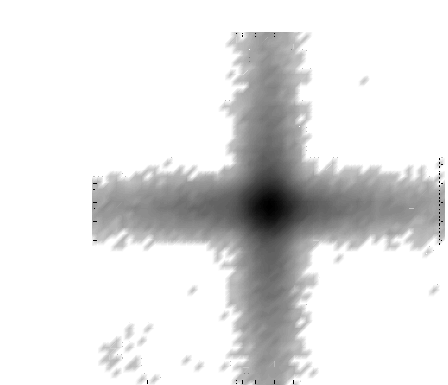} &
      \hspace{-0.5cm}\includegraphics[width=2.75cm]{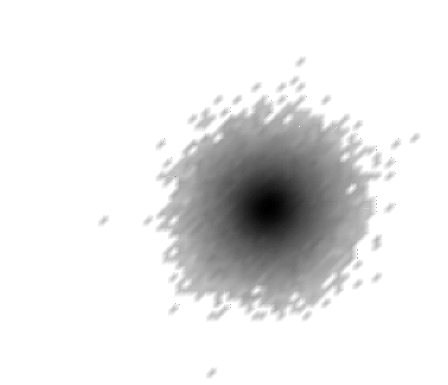} &
      \hspace{-0.5cm}\includegraphics[width=2.75cm]{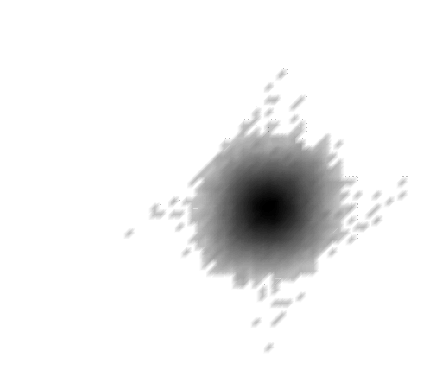} &
      \hspace{-0.5cm}\includegraphics[width=2.75cm]{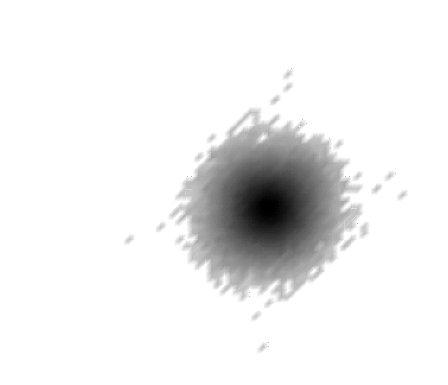} &
      \hspace{-0.5cm}\includegraphics[width=2.75cm]{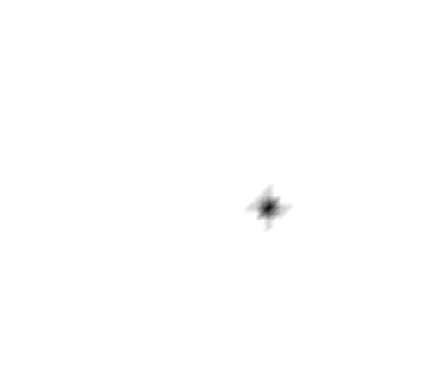} &
      \hspace{-0.5cm}\includegraphics[width=2.75cm]{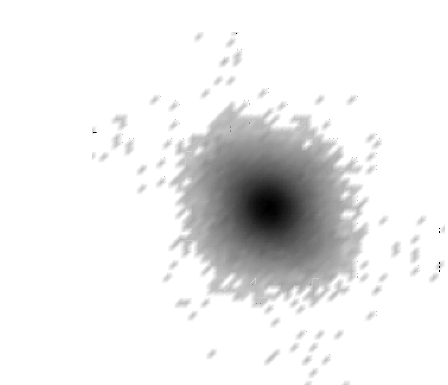} &
      \hspace{-0.5cm}\includegraphics[width=2.75cm]{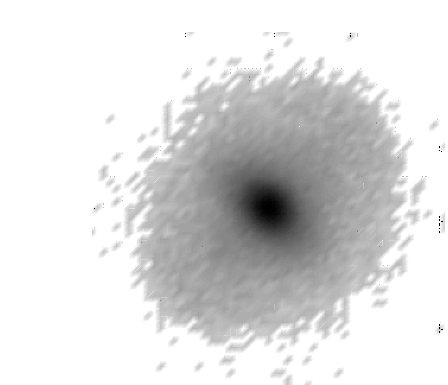} \\
    \end{tabular}
\end{center}
\caption{2D histograms of the residuals in the spatial domain for all
methods and noise sources (darker regions indicate higher probability). \green{In all cases, we considered a pixel and its right-hand neighbor (one pixel shift).} All the histogram values have been exponentiated to $0.25$ for better visualization.}
\label{residuos}
\end{figure}
\end{landscape}

\blue{Figure~\ref{residuos} represents the
distribution of the actual and estimated noise PDFs by all the considered methods in
the spatial domain. It can be noticed that, for the Gaussian noise, all methods reproduce quite
well the shape and extent of the PDF, as expected for the parametric models, which
use a proper Gaussian noise model. Note that the SVR method also succeeds in approximating
the energy of the noise even without using the Gaussian assumption explicitly.}

\blue{For non-Gaussian noise sources, the behavior of the methods markedly differ.
For instance, the quantization noise induced by JPEG/JPEG2000 follows a non-Gaussian,
oriented joint distribution (the central dark area is actually an oriented ellipsoid),
indicating correlation among noise samples. In the case of
JPEG, this central ellipsoid is better reproduced by hard thresholding and the proposed SVR method.
The other methods slightly underestimate the variance of the noise.
For the case of JPEG2000, methods not considering signal relations
dramatically underestimate the noise variance. In the case of more complex noise sources, such as
vertical striping or IRIS, none of the methods reproduce the low probability structure
(light gray regions). However, the central peak is poorly reproduced by marginal
methods, either overestimating (HT, ST, BG) or underestimating (BL) the width.
On the contrary, GSM and SVR give more reasonable width estimation. To conclude, methods
assuming an (inadequate) Gaussian noise model do not match, in general, the noise
distribution, so they should be reformulated for each particular noise source,
which may be complicated or even impossible. GSM constitutes an exception to this
statement, since results suggest that the quality of the signal model compensates the unsuitability of
the noise model. On the contrary, this is not necessary
for the proposed method, which only needs examples of
noisy images to {\em learn} from.
}

\section{Conclusions}

In this work, we proposed an alternative non-parametric way to take into account the relations
among natural image wavelet coefficients for denoising: we used
SVR in the wavelet domain to enforce these relations in the estimated signal.
The specific signal relations, which proved to be
more relevant in intraband coefficients, are encoded in an anisotropic
kernel based on mutual information computed from a
representative image database. \blue{An adaptive SVR with different cost function {\em per} subband was developed: the
subband-dependent $\varepsilon_i$ and $C_i$ are modeled by analyzing the particular
signal and noise variances in a representative image database.
By following general recommendations for the design of the kernel, $\varepsilon_i$ and $C_i$, and adapting them to the particular image denoising problem, we restricted the class of appropriate SVRs.
A KLD-based criterion was proposed to automatically select
the SVR that best recovers the relevant wavelet coefficient relations of the true
signal. The criterion was quite consistent but there is still room for improvement,
specially in the case of complex noise sources.}

\blue{Results show that the performance of the proposed non-parametric method is (1)
better than conventional wavelet methods
that assume coefficient independence, (2) similar to state-of-the-art methods
that do explicitly include these relations when the noise source is Gaussian, and (3)
numerically and visually better results are obtained when more complex realistic
noise sources are considered. Therefore, the proposed SVR approach can be seen as a more flexible (model-free)
alternative to the explicit description of coefficient relations. The important
thing here is that no reformulation is needed for dealing with any other kinds of noise.
Moreover, these results are an additional indication that relation between
local frequency coefficients is a salient natural image feature that should
not be neglected in denoising applications.}

\blue{Future work is tied to the incorporation of new information in the kernels:
here we focused on the consideration of signal relations in the kernel, but
the particular structure of the noise could be eventually incorporated.
Note that the denoising procedure is quite general and admits any kind
of non-parametric regression machine, such as Gaussian Processes. }

\section*{Acknowledgments}

This work was partially supported by projects CICYT-FEDER TEC2009-13696,
AYA2008-05965-C04-03, and CSD2007-00018. Valero Laparra acknowledges the
support of the Ph.D grant BES-2007-16125. The authors thank the reviewers for thorough and constructive comments on the
submitted manuscript.

\bibliographystyle{natbib}

\end{document}